\documentclass[journal]{IEEEtran}
\usepackage{cite}
\usepackage{times}
\usepackage{epsfig}
\usepackage{graphicx}
\usepackage{amsmath}
\usepackage{amssymb}
\usepackage{multirow}
\usepackage{float}
\usepackage{colortbl,booktabs}
\usepackage{soul}

\usepackage[pagebackref=true,breaklinks=true,letterpaper=true,colorlinks,bookmarks=false]{hyperref}

\newcommand{\figref}[1]{Figure~\ref{fig:#1}}
\newcommand{\tabref}[1]{Table~\ref{tab:#1}}

\newcommand{\secref}[1]{Section~\ref{sec:#1}}

\newcommand{\ve}[1]{\mbox{{\bf #1}}} 

\begin{document}
%
\title{Multi-View Photometric Stereo: A Robust Solution and Benchmark Dataset for Spatially Varying Isotropic Materials}
%
%
%
%

\author{Min~Li,
	Zhenglong~Zhou,
	Zhe~Wu,
	Boxin~Shi,
	Changyu~Diao,
	and~Ping~Tan

    \thanks{ M. Li is with the College of Computer Science and Technology and  C. Diao is with the Cultural Heritage Institute and School of Art and Archeology, Zhejiang University, China.}
   \thanks {B. Shi is with National Engineering Laboratory for Video Technology, Department of Computer Science and Technology, Peking University, Beijing, China.}
    \thanks{ P. Tan is with the School of Computing Science, Simon Fraser University, Canada.}
   
}
  \maketitle

\begin{abstract}
We present a method to capture both 3D shape and spatially varying reflectance with a multi-view photometric stereo (MVPS) technique that works for general isotropic materials. Our algorithm is suitable for perspective cameras and nearby point light sources.
Our data capture setup is simple, which consists of only a digital camera, some LED lights, and an optional automatic turntable.
From a single viewpoint, we use a set of photometric stereo images to identify surface points with the same distance to the camera.
We collect this information from multiple viewpoints and combine it with structure-from-motion to obtain a precise reconstruction of the complete 3D shape.
The spatially varying isotropic bidirectional reflectance distribution function (BRDF) is captured by simultaneously inferring a set of basis BRDFs and their mixing weights at each surface point. 
In experiments, we demonstrate our algorithm with two different setups: a studio setup for highest precision and a desktop setup for best usability.
According to our experiments, under the studio setting, the captured shapes are accurate to 0.5 millimeters and the captured reflectance has a relative root-mean-square error (RMSE) of 9\%. We also quantitatively evaluate state-of-the-art MVPS on a newly collected benchmark dataset, which is publicly available for inspiring future research. 
\end{abstract} 


\maketitle


%
\IEEEpeerreviewmaketitle

\vspace{1cm}

\IEEEraisesectionheading{\section{Introduction}\label{sec:introduction}}

\IEEEPARstart{C}{lassical} photometric stereo algorithms \cite{woodham1979photometric} estimate a per-pixel normal map from a set of images taken by a fixed camera under different lighting conditions. While they can reconstruct high-frequency geometric details, they often cause low-frequency shape distortions in a coarse scale\cite{Nehab2005}.
In comparison, multi-view photometric stereo\cite{Hernandez2008} (MVPS) algorithms integrate the results of photometric stereo from multiple different viewpoints. This approach can correct the low-frequency shape distortion by multi-view geometry constraints and also inherits the advantages of capturing fine details from original photometric stereo algorithms.

On the other hand, conventional photometric stereo methods typically make the following assumptions to facilitate normal estimation:
\begin{itemize}
	\item[(i)] The camera is orthogonal;
	\item[(ii)] The surface material is Lambertian with no cast shadows;
	\item[(iii)] Lighting is directional (e.g., distant point light sources).
\end{itemize}

However, the real world often presents much more complicated settings with perspective cameras, nearby light sources, and more complicated materials. It is thus extremely difficult to simultaneously recover the unknown material and object shape, even under known lighting conditions.
To address this challenge, sophisticated hardware such as light stages\cite{Ghosh2009}, coaxial lights\cite{Holroyd2010}, and near-field light stages \cite{kang2019learning} have been designed. 
Though these methods achieve highly accurate results, the setups are expensive and complicated. We design a method with a simple low-cost setup so that it can be adopted more widely. Our setup only contains a digital camera and some LED lights. We design sophisticated calibration and reconstruction algorithms to address these difficulties. Compared with the advanced setups, e.g., the coaxial lights in \cite{Holroyd2010}, our method achieves lower but still useful accuracy ($0.5$ millimeters vs. $50$ microns).
We hope our lightweight solution can enable casual users  to perform high-quality appearance capture in the future.

The conference version of this paper \cite{Zhou2013} relaxes the Lambertian material assumption of MVPS to deal with general spatially varying isotropic materials. Specifically, we exploit reflectance symmetries such as isotropy and reciprocity to deal with those general materials. 
According to \cite{Alldrin2007}, isotropy allows us to identify `iso-depth contours', i.e., pixels corresponding to surface points of equal distances to the image plane, from photometric stereo images. 
We collect iso-depth contours from multiple viewpoints to reconstruct the complete 3D shape. Specifically, we first apply structure-from-motion \cite{Hartley2003} to reconstruct a sparse set of 3D points. We then propagate the depths of these 3D points along iso-depth contours in each viewpoint. Each propagation generates additional 3D points, whose depths can be further propagated in a different viewpoint. A surprisingly small number of 3D points (about two hundred) can be propagated to reconstruct the complete 3D shape (about two hundred thousand points). Once the shape is fixed, we use the same set of input images to infer the spatially varying materials. We model reflectance by the Bidirectional Reflectance Distribution Function (BRDF) and assume the BRDF at each surface point is a linear combination of a few basis isotropic BRDFs, each of which is a 3D discrete table to handle general materials. The basis BRDFs and mixing weights at each point are iteratively estimated by the ACLS method \cite{Lawrence2006}. 


This journal extension further relaxes the assumptions of orthographic camera and distant lighting (i.e., directional lighting). This extension allows us to build a compact desktop scanner of a microwave oven size for appearance capture, where the object is only $400$ mm away from the camera and LED lights. 
To handle perspective cameras, we divide the image plane using a 2D grid, where each grid cell acts as the image of an orthogonal camera, so that the original iso-depth contours can be evaluated safely. 
On the other hand, we also introduce a calibration method with a simple white board to calibrate the 3D position of each LED light and its radiance towards different directions. This sophisticated lighting model allows us to eliminate the undesired non-uniform lighting of nearby LED lights to simplify the photometric stereo problem.

At last, to evaluate multi-view photometric stereo algorithms, we build the `DiLiGenT-MV' dataset, which is a multi-view extension of the `DiLiGenT' dataset in \cite{shi2019benchmark} for benchmarking photometric stereo algorithms. This new dataset contains images of 5 objects of complex BRDFs. The images are taken from 20 viewpoints and in each viewpoint, 96 calibrated point light sources are used. The `ground truth' shape is available for quantitative evaluation. This `DiLiGenT-MV' can be used to evaluate multi-view stereo methods (e.g., \cite{Galliani2015,schops2017multi}) under complex materials for lighting, be used to evaluate conventional single-view photometric stereo algorithms (e.g., \cite{Alldrin2008,shi2014bi}) by treating each viewpoint independently. We quantitatively evaluate recent multi-view photometric stereo algorithms to further understand their pros and cons so as to encourage further research on unsolved issues. 

Our main contributions are threefold:
\begin{itemize}
	\item[$\bullet$] We propose a multi-view photometric stereo technique to work with general spatially varying isotropic materials, which allows faithful appearance capture (i.e., shape + BRDF capture).

	\item[$\bullet$]We relax the assumption of perspective camera and distant lighting to build a simple desktop capture setup, which enables casual users to perform high quality appearance capture.

	\item[$\bullet$]We present the `DiLiGenT-MV' dataset with objects of complex materials and `ground truth' shapes for benchmarking multi-view photometric stereo methods. 
\end{itemize}

\section{Related Work}
\subsection{Image-based Modeling}

These methods reconstruct a 3D shape
and a `texture map' to model objects from images.
The methods in \cite{Lhuillier2005,Furukawa2010} are two recent representative methods. 
Texture color at each surface point is decided according to its image projections. However, a texture map is often insufficient to represent general non-Lambertian materials.

\subsection{Shape Scanning and Reflectance Fitting}

To obtain precise 3D shape, laser scanners and structured-light patterns were used in \cite{Levoy2000,Rusinkiewicz2002}, and \cite{Zhang2004}. Based on a precise 3D reconstruction, parametric reflectance functions can be fitted at each surface point according to the image observations, as in \cite{Sato1997} and \cite{Lensch2003}. These methods require precise registration between images and 3D shapes. Since different sensors are used for shape and reflectance capture, this registration is difficult and often causes artifacts in misaligned regions. Some methods \cite{Nehab2005,Aliaga2008} combine reflectance recovered from photometric stereo and shape recovered from structured-light, where registration is relatively simple. However, they need to capture images under both structured-light and varying directional light at \emph{each} viewpoint, which is tedious and requires a more complicated setup than ours.

\subsection{Photometric Appearance Capture}

Our method belongs to photometric approaches that capture both shape and reflectance from the same set of images. Most of previous methods, e.g., \cite{Lim2005,Hernandez2008,Goldman2005}, assumed specific
parametric BRDF models such as Lambert's or Ward's model
\cite{Ward1992}. The performance of these methods degrades when the real objects have different reflectance from the assumed model.

Some other methods employed a sophisticated hardware setup to achieve high-quality results. Ma et al. \cite{Ma2007} and Ghosh et.al. \cite{Ghosh2009} used a light stage where the intensity of each LED on the stage was precisely controlled. Holroyd et al. \cite{Holroyd2010} required specialized coaxial lights. This requirement of expensive and complicated hardware limits their wide application. Recently, a few algorithms \cite{Alldrin2008,Holroyd2008} were proposed for appearance capture by exploiting various reflectance symmetries that are valid for a broader class of objects. However, the method in \cite{Holroyd2008} required up to
a thousand input images at \emph{each} viewpoint and
\cite{Alldrin2008} relied on fragile optimization. Tan et al. \cite{Tan2011} and Chandraker et al. \cite{Chandraker2011} both recovered iso-contours of depth and gradient magnitude for isotropic surfaces. Additional user interactions or boundary conditions are required to recover the 3D shape.
A recent work \cite{Park2013} developed an uncalibrated mult-iview photometric stereo method to reconstruct meshes with fine geometric details by estimating a displacement map in the 2D texture domain. However, this work does not deal with surface reflectance.
Along the direction of uncalibrated methods, Lu et al. \cite{Lu2013} recovered surface normal for isotropic surfaces. 
But as evaluated in \cite{shi2019benchmark}, this method requires many more input images per viewpoint and produces a  larger error. 

There is only limited works to deal with perspective cameras and near lighting effects until now. Under these settings, the photometric stereo problem becomes nonlinear even with the basic Lambertian model. Extra information is often incorporated to solve this challenge. Tigo et al.\cite{higo2009} employed a sparse depth map to simplify the computation and Xie et al.\cite{xie2015photometric} overcame the difficulty by utilizing a mesh deformation technique. There are also  methods to deal with the near-light effects of point lights, e.g., \cite{iwahori1990,higo2009,papadhimitri2013,papadhimitri2014}. However, these works are all based on Lambert's model and only recover a normal map.
Recently differential photometric stereo methods\cite{Yvain2017,Yvain2018} are proposed to solve the perspective and near-light effects by using nonlinear PDEs, while leads to complex optimization for normal and depth estimation.

The work closest to our method is \cite{Alldrin2008}. Both methods are built upon reflectance symmetry embedded in `isotropic pairs' introduced in \cite{Tan2007}. There are three key differences between our method and \cite{Alldrin2008}. First, we reconstruct a complete 3D shape rather than a single-view normal map. Second, we
combine multi-view geometry and photometric cues to avoid fragile iterative optimization of shape and reflectance. Third, our method works with general tri-variant isotropic BRDFs under the perspective projection while \cite{Alldrin2008} assumed bi-variant BRDFs and the orthogonal projection to simplify the optimization.



Our work is also related to BRDF acquisition methods such as \cite{Dong2010,Ren2011}. These methods are only applicable to near-flat surfaces where the surface normals
are known beforehand. Our method can be considered as a
generalization of these methods to non-planar surfaces. 

\subsection{Datasets for 3D reconstruction}

The first common benchmark, Middlebury dataset, was proposed by Steitz et al.\cite{seitz2006comparison} for evaluating
multi-view stereo on equal grounds, containing only two scenes with Lambertian surfaces. Later, Strecha et al.\cite{strecha2008benchmarking} proposed a new MVS benchmark dataset including 6 outdoor scenes with up to 30 images with higher resolution and `ground truth' shapes captured by a laser scanner. This dataset covers well-textured scenes, though the online benchmark service is not available anymore. To compensate for the lack of diversity in \cite{seitz2006comparison,strecha2008benchmarking}, Jensen et al.\cite{jensen2014large} published a number of real-world objects, which are still limited in the variety of scenes and viewpoints. Knapitsch et al.\cite{knapitsch2017tanks} and Schops et al.\cite{schops2017multi} provided the latest challenging datasets for indoor and outdoor scenes with high-resolution video data and `ground truth' measurements obtained with a laser scanner, focusing on evaluating binocular stereo, multi-view stereo, or structure-from-motion. Yet, their objects share the same limitation as those in earlier datasets, i.e, lacking diversity in both surface reflectance and viewpoints.

Existing multi-view stereo datasets are limited in the variation of lighting conditions and challenging BRDFs, which is the key issue in photometric stereo. Shi et al.\cite{shi2019benchmark} proposed the `DiLiGenT' dataset for single-view photometric stereo. We extend it to the multi-view setup and release the `ground truth' 3D shape as well as camera calibration information to facilitate future research on MVPS.

\section{Shape and reflectance Reconstruction}\label{sec:shape}

\begin{figure*}\centering
	\includegraphics[width= 0.85 \linewidth]{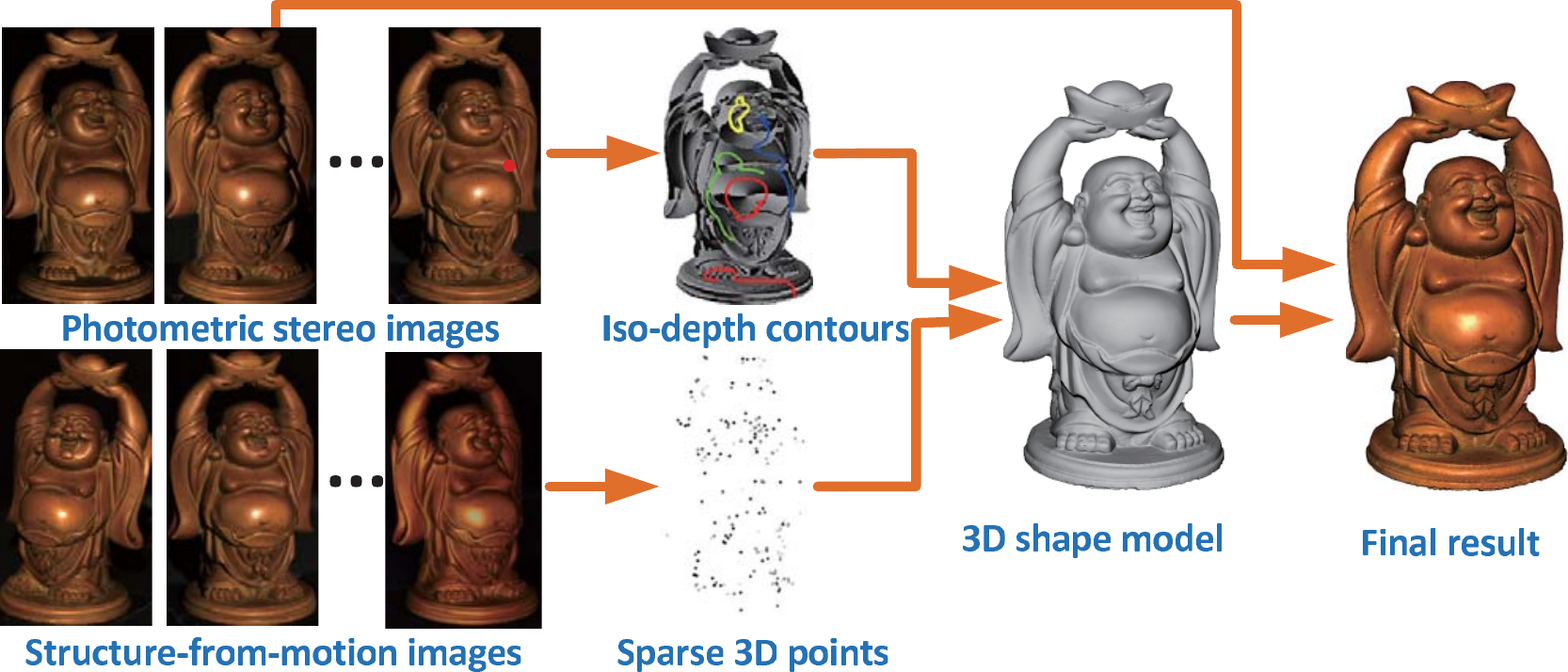}
	\caption{System pipeline. We recover iso-depth contours from photometric stereo images and recover a sparse 3D point cloud by structure-from-motion. In the figure showing iso-depth contours, the gray intensity encodes the estimated azimuth angles, and the colored
	curves are iso-depth contours. We then propagate the depths of these 3D points along the iso-depth contours to recover the complete 3D shape. Once the shape is fixed, we estimate the spatially varying BRDF from the original input images. }\label{fig:pipeline}
\end{figure*}

We provide a block diagram of our system in \figref{pipeline}. 
We capture images from multiple viewpoints, and at each viewpoint, we capture photometric stereo images under different lighting conditions. 
Our algorithm robustly identifies iso-depth contours from these images at each viewpoint.
On the other hand, we apply a structure-from-motion algorithm \cite{Hartley2003} to images from different viewpoints to reconstruct a sparse set of 3D points. 
We then derive a complete 3D shape by propagating the depths of these points along the dense iso-depth contours. 
This initial shape is further refined according to the method described in \cite{Nehab2005}. 
Once the shape is fixed, we estimate a set of basis isotropic BRDFs and their mixing weights at each surface point by the ACLS method \cite{Lawrence2006} to model the surface reflectance. In the following, we first describe the method for orthographic camera and distant lighting in Section\secref{contour} and \secref{propagation}. We then relax the distant lighting and orthogrpahic camera assumptions in Sections \secref{near_light} and \secref{perspective} respectively.

\subsection{Basic Iso-depth Contour Estimation}\label{sec:contour}

	Assuming orthographic camera and directional lighting,
	Alldrin and Kriegman \cite{Alldrin2007} observed that isotropy allows almost trivial estimation of iso-depth contours in the absence of global illumination effects such as shadows and inter-reflections.
	We generalize this algorithm to make it more robust in real data than the na\"{\i}ve approach described in \cite{Alldrin2007}.
	Specifically, we relax the assumption about lighting (i.e., precisely located on a view centered circle as in \cite{Alldrin2007})
	and propose a method to enhance robustness to global illumination effects.


When the light moves on a view-centered circle, the plane spanned by the viewing direction and the surface normal direction of an isotropic\footnote{Note that the original algorithm is based on the bilateral symmetry. Here, we follow \cite{Alldrin2007} to refer it as isotropy because bilateral symmetry is often observed for isotropic surfaces.} surface point can be recovered precisely according to the symmetry of the observed pixel intensity profile. In the camera local coordinate system, where the $z$-axis is aligned with the viewing direction, this plane gives the azimuth angle of the surface normal, which is the angle between the $x$-axis and the projection of normal in the $xy$-plane. 
For easier reference, we refer this direction of a projected surface normal as the azimuth direction in this paper. The details of the azimuth direction computation are in the Section 4.1 of the conference version of this paper \cite{Zhou2013}.

\textbf{Tracing Contours.}
Once an azimuth direction is computed at each pixel, we proceed to generate iso-depth contours. Starting from every pixel, we iteratively trace along the two directions perpendicular to the azimuth direction with a step of $0.1$ pixel. Specifically, suppose the estimated azimuth angle is $\theta$ at a
pixel $\ve{x}$. We trace along the two 2D directions
$$
\ve{d}_{+} = (cos (\theta + \pi/2), sin (\theta + \pi/2) )
$$
and
$$
\ve{d}_{-} = (cos (\theta - \pi/2), sin (\theta - \pi/2) )
$$
to $\ve{x}_{+} = \ve{x}+0.1\ve{d}_{+}$ and $\ve{x}_{-} = \ve{x}+0.1\ve{d}_{-}$. We
then replace $\ve{d}_{+}$ and $\ve{d}_{-}$ according to the azimuth angles of $\ve{x}_{+}$ and $\ve{x}_{-}$ respectively and continue to trace. We stop tracing when the maximum number of iterations is reached (500 in our experiments). Pixels on one traced curve should
have the same distance to the image plane. To avoid tracing across discontinuous surface points, we use the method described in the `NPR camera' \cite{Raskar2004} to identify depth discontinuities. Further, we define a confidence measure for these traced contours as the inverse of the maximum curvature along them. Intuitively, smoother contours with relatively small curvature are more reliable.

\subsection{Multi-View Depth Propagation}\label{sec:propagation}
A standard structure-from-motion algorithm such as
\cite{Lhuillier2005,Snavely2006} can reconstruct a set of sparse 3D points on the object. We capture experiment objects on a turntable with a checkboard pattern to ensure sufficient feature matching for textureless examples. Since structure-from-motion algorithms could be affected by moving highlights, we compute a median image at each
viewpoint by taking the median intensity of each pixel and use these images for feature matching. Reconstructed 3D points are combined with the traced iso-depth contours to recover the complete 3D shape.

\begin{figure}\centering
	\includegraphics[width= 0.8 \linewidth]{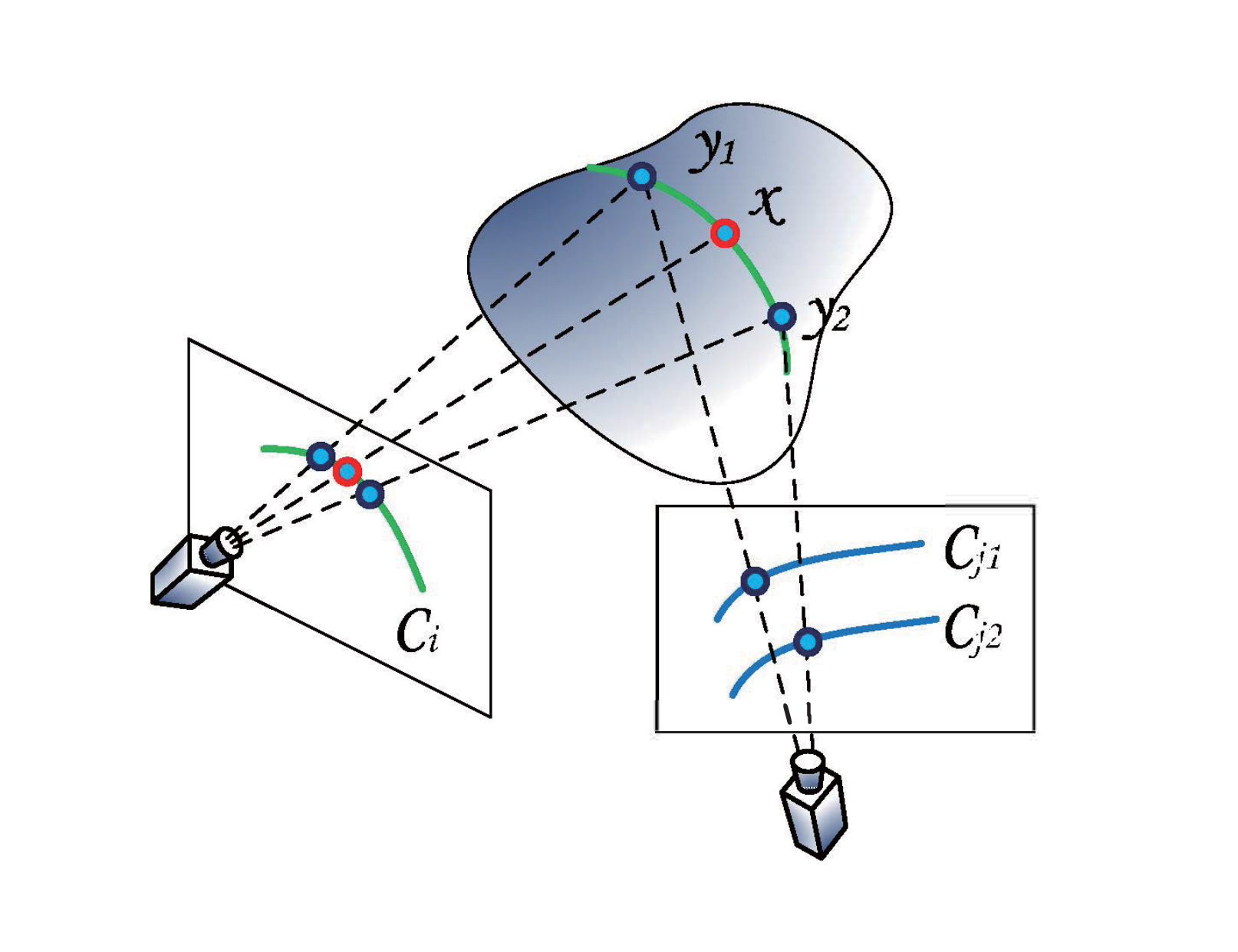}
	\caption{We propagate the depth of $\ve{x}$ to the iso-depth contour
		segment $C_i$ that passes through its projection in the $i$-th view.
		This propagation generates new 3D points, e.g., $\ve{y}_1, \ve{y}_2$,
		whose depths in other images can also be propagated along their
		corresponding iso-depth contours $C_{j1}, C_{j2}$.
	}\label{fig:propagation}
\end{figure}

\textbf{Depth Propagation.} As illustrated in \figref{propagation}, given a reconstructed 3D point $\ve{x}$, we project it to all images where it is visible. Suppose an iso-depth contour $C_i$ goes through its projection in the $i$-th image. We perform a depth propagation to assign the depth of $\ve{x}$ to all pixels on $C_i$ (If the depth of a pixel on $C_i$ is already known, we keep it unchanged).
This propagation generates new 3D points, whose depths can be propagated in other images too.  
We begin with a sparse set of 3D points $P$ reconstructed by structure-from-motion. Depth propagation
with $P$ in all images generates a large set of 3D points $P'$. We then replace $P$ by $P'$ and apply depth propagation iteratively. We keep iterating until $P'$ is empty.
Note when dealing with perspective cameras in \secref{perspective}, we take each sub-divided cell as an individual orthographic camera.

Direct application of the algorithm described above will generate poor results. There are a few important issues which must be addressed for robust 3D reconstruction.

\textbf{Point Sorting.} We sort all points in $P$ according to the confidence of their associated iso-depth contours. Note that if a point is visible in $K$ different views, it is repeated $K$ times in $P$ and each repetition is associated with an iso-depth contour in one view. At each iteration, we only select half of the points in $P$ of high confidence for depth propagation. We then remove those
selected points, and insert $P'$ into the sorted set $P$ for the next iteration.

\textbf{Visibility Check.}\label{sec:consistency} We should not propagate the depth of a 3D point in an image where it is invisible. However, the visibility information is missing for 3D points generated by propagation. So we apply a consistency check when propagating the depth of a 3D point $\ve{x}$ to a contour $C$. We
check pixels on $C$ one by one, starting from the projection of $\ve{x}$ to the two ends of $C$. If a pixel $p$ fails the check, we truncate $C$ at $p$, and only assign the depth of $\ve{x}$ to pixels on the truncated contour. If the updated contour is too short (less
than 5 pixels in our implementation), we do not propagate.

To evaluate consistency at a pixel $p$, we assign it the depth of $\ve{x}$ to determine its 3D position. We then use the surface normal of $\ve{x}$ to select $L$ ($L=7$ in our implementation) most front parallel views where $\ve{x}$ is visible. We assume $p$ is visible in all these $L$ images and check the consistency of the
azimuth angles at its projections. The azimuth angles at
corresponding pixels in two different views uniquely decide a 3D normal direction\footnote{An azimuth angle in one view (with the
camera center) decides a plane where the normal must lie in. Intersecting two such planes determines the 3D normal direction.}.
If different combinations of these $L$ views all lead to consistent 3D normals (the angle between any two normals is within $T$ degrees), we consider $p$ as consistent. Otherwise, we discard one view that leads to the largest number of inconsistent normals and check consistency with the remaining $L-1$ views iteratively. We consider $p$ consistent, if it is consistent over at least 3 views.
Otherwise, it is inconsistent. For each consistent 3D point, we set its normal as the mean of all consistent normals. In our implementation, we begin with $T=3$, and relax it by $1.3$ times whenever $P'$ is empty until $T>15$.

We note the number of consistent views for each 3D point when inserting it to the set $P'$. Points are first sorted by the number of consistent views in descending order. Those with the same number of consistent views are sorted by the confidence of contours.

\subsection{Shape Optimization}
After depth propagation, we have a set of 3D points, each with a normal direction estimated. We apply the Poisson surface reconstruction \cite{Kazhdan2006} to these points to obtain a triangulated surface. This surface is further optimized according to \cite{Nehab2005} by fusing the 3D point positions and their normal directions.

\subsection{Near Light Effects}
\label{sec:near_light}
In practice, we often use LED lights as light sources, which are nearby point lights leading to different illumination directions at different pixels.
The lighting intensity is non-uniform for an LED light too. An LED light has different emission radiance towards different directions, which further falls off along those directions. We address these problems to make the algorithm more practical.

\textbf{Lighting Directions.}
For the setups in our experiments, we calibrate the precise 3D positions of the light sources to compute spatially variant lighting directions at each pixel.
Calibration details are provided in \secref{device_calibration}.
After calibrating light source positions, we take the average depth of an object (computed from the reconstructed sparse 3D points in \secref{propagation}) to estimate an approximate 3D position of each pixel.
The lighting directions at each pixel are then computed according to the 3D positions of that pixel and the light sources.


We interpolate observations under lighting directions lying on a view-centered circle, and compute the azimuth angle from these interpolated observations. More details of this process are in the Section 4.1 of the conference version of this paper \cite{Zhou2013}.

\textbf{Lighting Intensities.}
An LED light often has non-uniform intensity towards different orientations, which needs to be calibrated for high precision reconstruction.
Furthermore, lighting intensities are inversely proportional to the square of its travel distance.
We calibrate a 3D field of lighting intensities for each LED light, which is referred as lighting intensity volume in the following of this paper.
Calibration details are provided in \secref{device_calibration}.
After calibration, we normalize all observations to the same lighting intensity,
i.e., dividing each observed pixel intensity by the lighting intensity at its 3D position. However, we have no knowledge about the accurate 3D position at this stage, so we assume the observations lying on the same plane with an approximated constant depth.

\subsection{Perspective Camera Effects}\label{sec:perspective}
\begin{figure} \centering
	\includegraphics[width= 0.35 \linewidth]{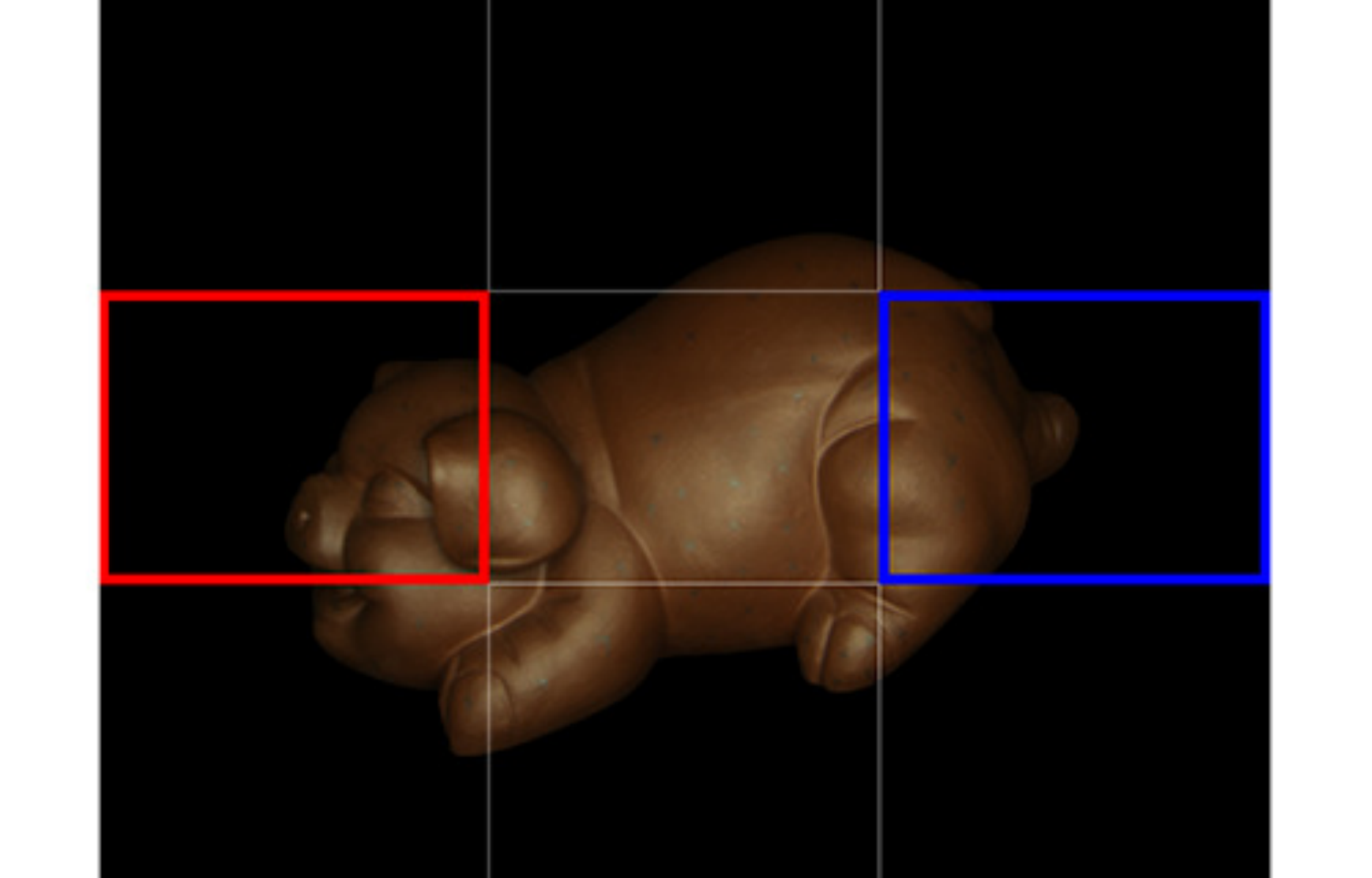}
	\includegraphics[width= 0.60 \linewidth]{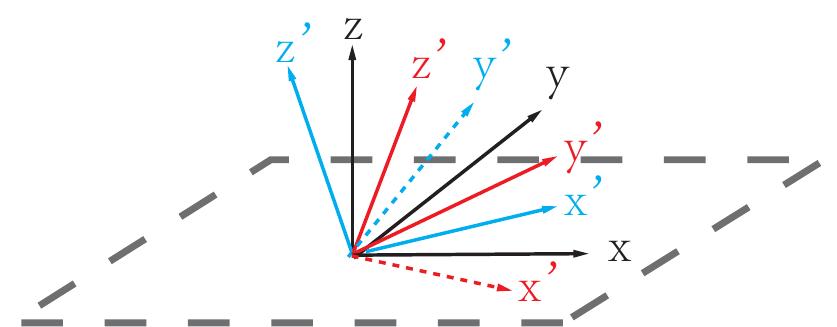}
	\caption{
	Left: an image plane $I$ divided into an array of rectangular cells, each is considered as an orthographic camera.
    Right: black axes are the coordinates of the original perspective camera.
    Red and blue coordinates are those of the two orthographic cameras indicated by the red and blue frames in the left image.
	}
	\label{fig:camera_rotation}
\end{figure}

When the camera is nearby, we need to consider the perspective effects of camera projection.
In other words, every pixel should have a different viewing direction.
To address this problem, we divide the image plane to a 2D array of cells, typically $3\times 3$,
as shown in the left of \figref{camera_rotation}.
Each cell can be considered as an individual camera with much smaller field-of-view (FoV), and this camera can be well approximated as an orthographic camera.

The extrinsic parameters of these sub-divided orthographic cameras can be easily computed by rotating the original perspective camera.
Specifically, as indicated in the right of \figref{camera_rotation}, the principal axis $Z'$ of a sub-divided orthographic camera
is simply the view ray passing through its cell center.
Suppose the principal axis of the original perspective camera is $Z$.
We define a rotation matrix $R$, which is the minimum rotation that rotates $Z$ to $Z'$.
We apply $R$ to the original camera axes $X, Y, Z$ to obtain $X', Y', Z'$ axes of the orthographic camera.
The iso-depth contours in each sub-divided orthographic camera are in different image planes, as the $X'$ and $Y'$ axes are different for different sub-dividing cells.

Note that the observed images of a sub-divided orthographic camera is generated by applying a homography to the original image.
This homography can be easily computed from the intrinsic parameters and the rotation matrix.

%
%

\begin{figure}
	\centering
	\includegraphics[width=0.45\linewidth]{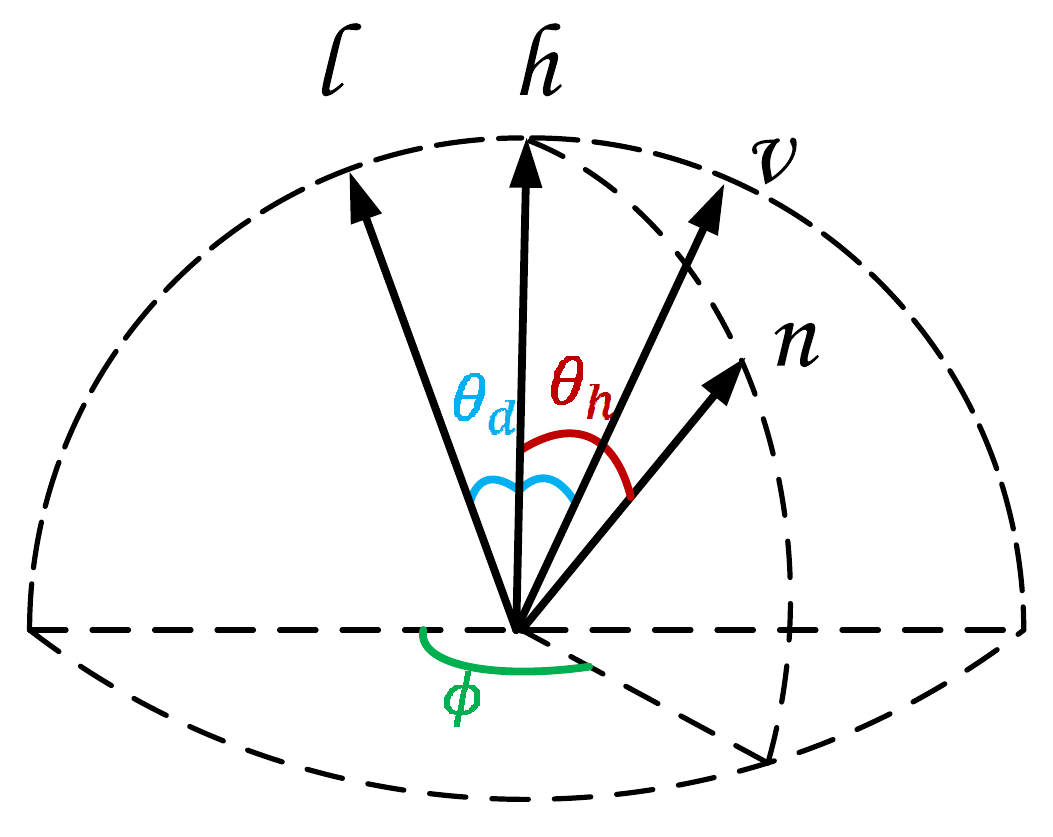}
	\caption{Definition of $\theta_h, \theta_d$ and $\phi$, which an isotropic BRDF is defined as a 3D lookup table.}\label{fig:parameters}
\end{figure}

\subsection{Reflectance Capture}\label{sec:BRDF}

We assume the surface reflectance can be represented by a linear combination of several ($K = 2$) basis isotropic BRDFs. Once the 3D shape is reconstructed, we follow \cite{Lawrence2006} to estimate the basis BRDFs and their mixing weights at each point on the surface. We consider the general tri-variant isotropic BRDF, which is a function of $\theta_h, \theta_d, \phi$ as shown in
\figref{parameters}. We discretize $\theta_h, \theta_d$ and $\phi$ into $90, 2,$ and $5$ bins respectively all in the interval $[0,\pi/2]$. Please refer to \cite{Romeiro2010} for a justification of
choosing this interval. Hence, a BRDF is represented as a $900 \times 1$ vector by concatenating its values at these bins.

We build an $N\times M$ observation matrix $\ve{V}$, and factorize it into a matrix of mixing weights $\ve{W}$ and a matrix of basis BRDFs $\ve{H}$ as
\begin{displaymath}
\ve{V}_{N\times M} = \ve{W}_{N\times K} \ve{H}_{K \times M}.
\end{displaymath}
$M=900$ is the dimension of a BRDF. $N$ is the number of 3D points. Each row of $\ve{V}$ represents the observed BRDF of a surface point. In constructing the matrix $\ve{V}$, we avoid pixels observed from slanted viewing directions (the angle between viewing direction
and surface normal is larger than $40$ degrees in our
implementation), where a small shape reconstruction error can  cause a big change in their projected image positions. $\ve{V}$ contains missing elements because of incomplete observation. We apply the Alternating Constrained Least Squares (ACLS) algorithm
\cite{Lawrence2006} to iteratively compute the rows of $\ve{W}$ and columns of $\ve{H}$.

To further improve reflectance capture accuracy, we first compute $\ve{H}$ from a subset of precisely reconstructed 3D points, whose reconstructed normals from different combinations of azimuth angles are consistent within $1.5$ degrees. We then fix $\ve{H}$ and compute $\ve{W}$ at all surface points.

\section{Multi-view photometric stereo dataset}\label{sec:DATASET}

\begin{figure}[htb]
\includegraphics[height=1.0 \linewidth]{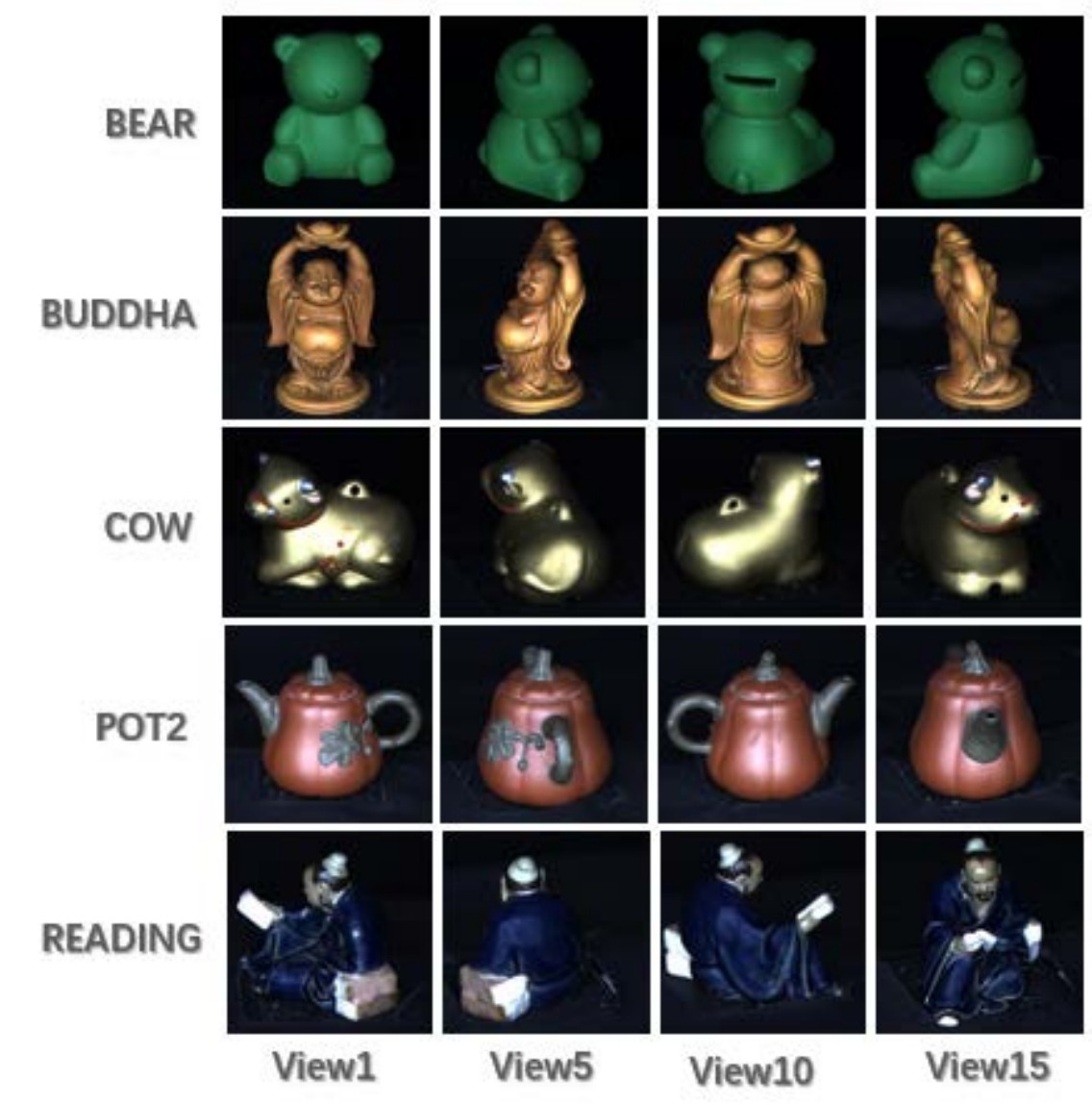}

	\caption{Example images of five objects from different views (4 out of 20) in `DiLiGenT-MV'. From top to bottom: BEAR, BUDDHA, COW, POT2, READING.}
	\label{fig:dataset_name}
\end{figure}

In this section, we introduce the `DiLiGenT-MV' dataset. It includes five objects: BEAR, BUDDHA, COW, POT2, and READING as shown in \figref{dataset_name}.
None of the existing multi-view benchmark datasets are suitable for evaluating our method since they have various limitations: simple reflectance \cite{knapitsch2017tanks,schops2017multi}, a limited number of viewpoints\cite{seitz2006comparison}, or inadequate lighting variations \cite{jensen2014large}.  
The `DiLiGenT' dataset in \cite{shi2019benchmark} captures objects with complex reflectances under large illumination changes, but it only has a single viewpoint for each object. This motivates us to extend `DiLiGenT' to multiple viewpoints. 
Our dataset is captured under the same configuration as `DiLiGenT' except that we provide images illuminated by 96 different lights from 20 different viewpoints. We release the normal maps from all 20 viewpoints as well as the scanned 3D shape. Please refer to \cite{shi2019benchmark} for details of image capture procedure, lighting and camera calibration, scanning and alignment of `ground truth' shape. The data is available for download at: \url{https://sites.google.com/site/photometricstereodata/}.

\begin{figure*}\centering
	\includegraphics[height = 0.25 \linewidth]{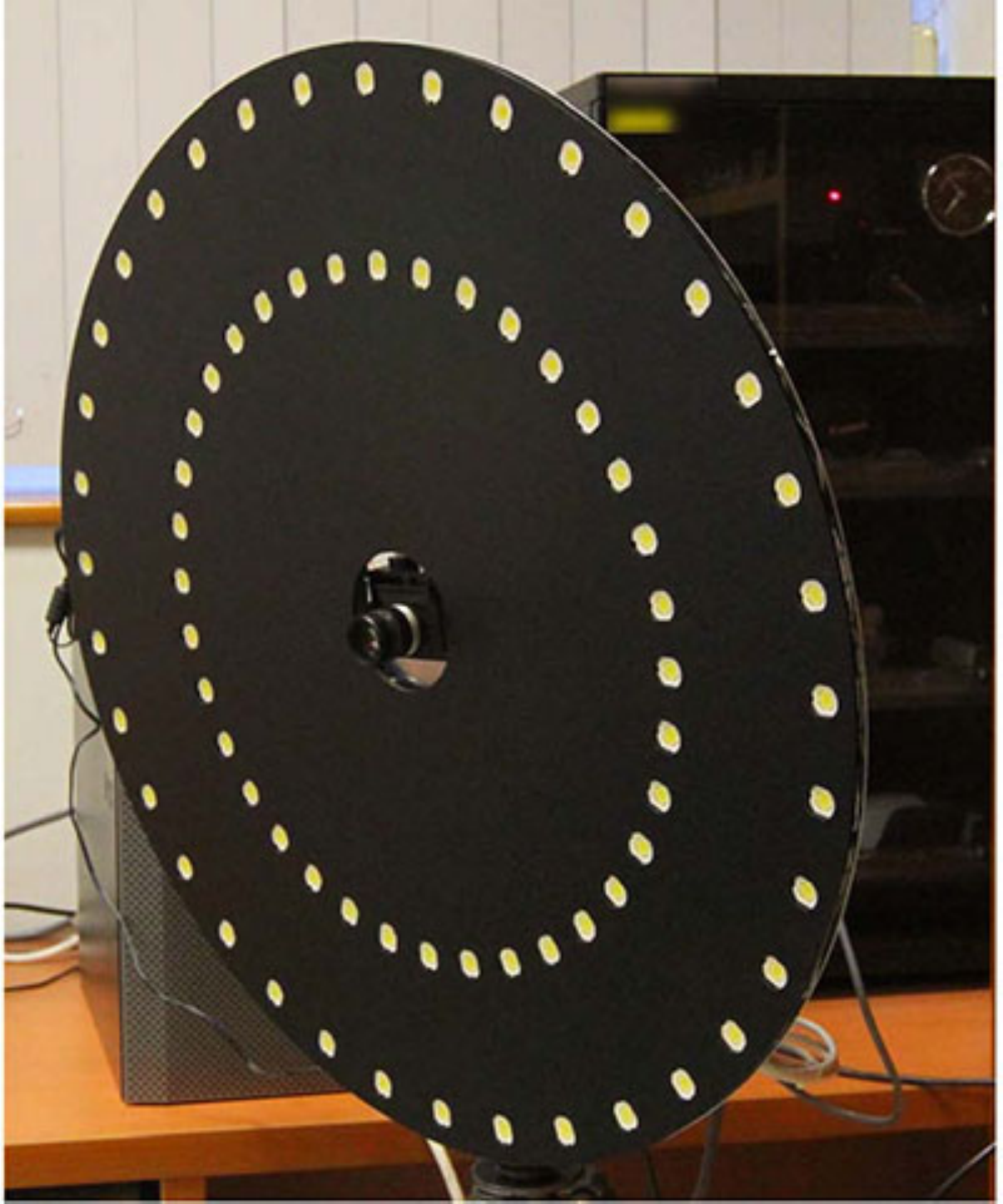}
	\includegraphics[height = 0.25 \linewidth]{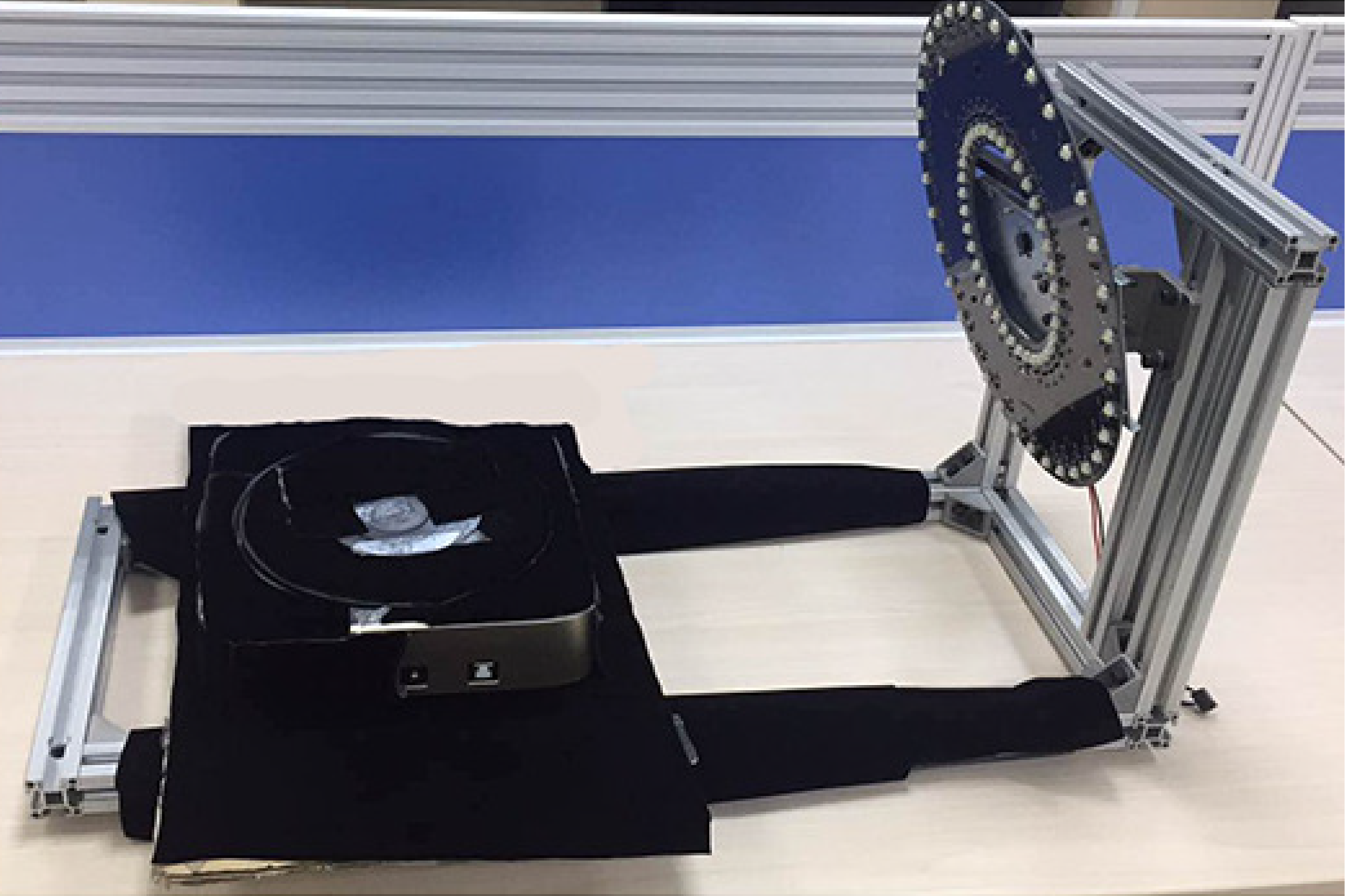}	
	\includegraphics[height = 0.3 \linewidth]{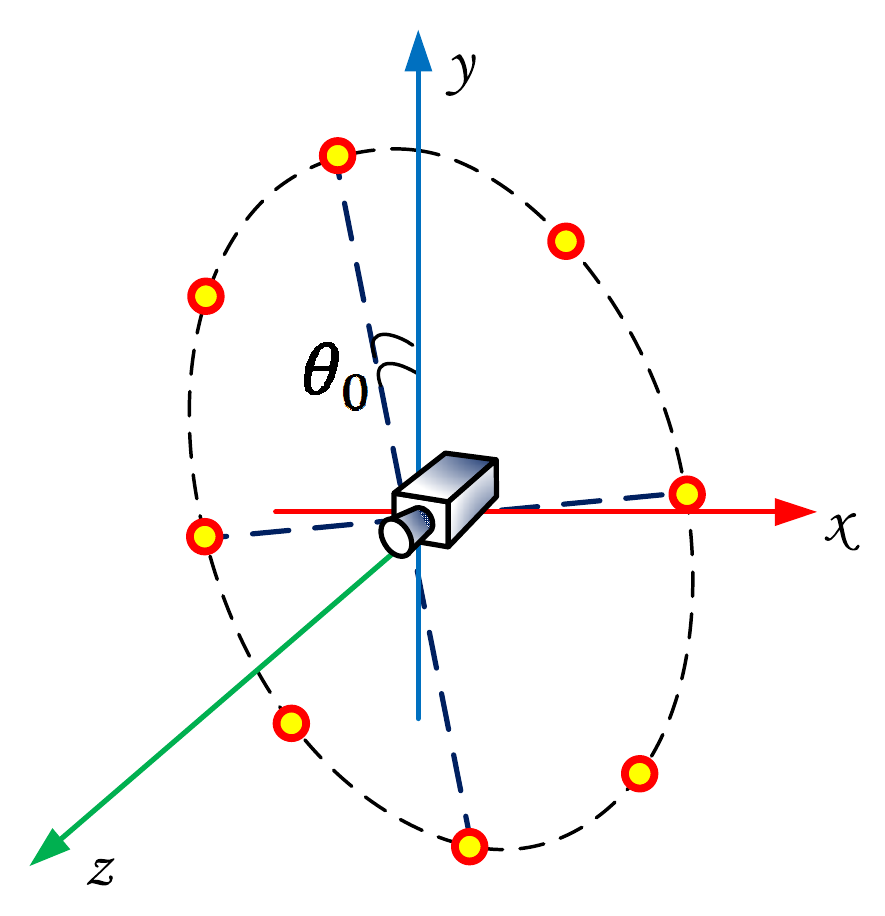}
	\caption{Left: the studio scanner setup. Middle: the desktop scanner setup.
        Our devices consist of a video camera, two circles of LED lights, and an automatic turntable (for the desktop setup).
		The object to camera distance is $1000 \sim 1500$ millimeters and $400$ millimeters for the studio and desktop setup respectively.
        We model the perspective camera projection and near point lights for the desktop setup.
        Right: we can calibrate	one parameter $\theta_0$ to determine 3D position of the LED lights with known radius of the underlying circles.
	}\label{fig:device_calibration}
\end{figure*}

\section{Data Capture Setups}
\label{sec:devices}
We build two different data capture setups: a studio scanner setup and a desktop scanner setup.
The studio and desktop scanner setups use automatically blinking LED lights synchronized with a video camera to speedup data capture.
The studio and desktop scanner setups are shown in the left and middle of \figref{device_calibration} respectively.
In the studio and desktop setups, the testing object is about $1000 \sim 1500$ and $400$ millimeters away from the camera respectively.
Therefore, we have to consider perspective camera effects (as in \secref{perspective}) and near light effects (as in \secref{near_light}), especially for the desktop setup.

The studio setup uses a PointGrey Grasshopper camera, which captures linear images at about $1200\times900$ resolution.
The desktop setup uses a cheaper linear industry camera at $1280\times960$ resolution.
We capture images viewpoint by viewpoint.
After capturing images at one viewpoint, we manually rotate the object to capture the next viewpoint.
The desktop scanner setup further uses an automatic turntable to automate this rotation, making the whole data capturing process automatic.

\begin{figure} [htb] \centering
	\includegraphics[width= 0.5 \linewidth]{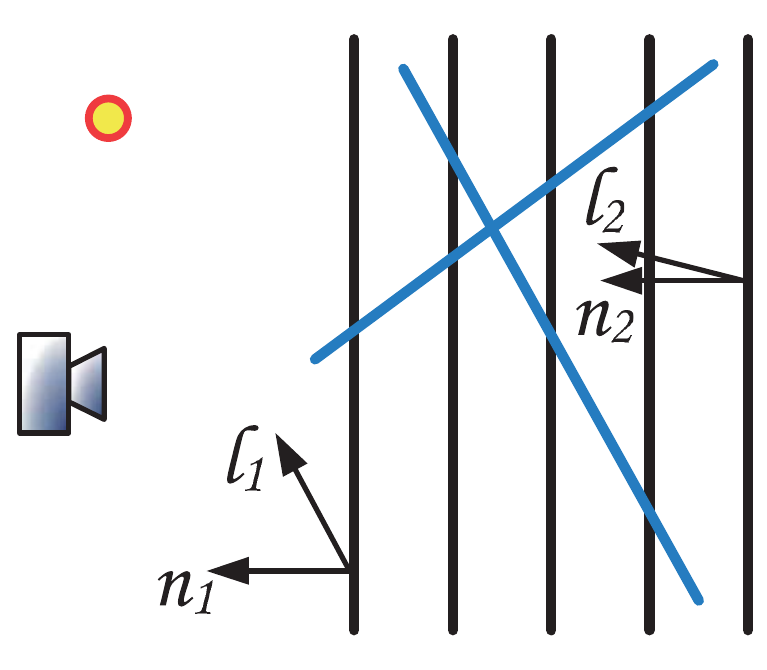}
	\caption{Top view of the calibration setup. We capture a diffuse board at several known positions (black lines) to calibrate camera vignetting and lighting intensity. Some additional boards (blue lines) are used to calibrate the angle $\theta_0$.}
\label{fig:light_calibration}
\end{figure}

\begin{figure} \centering
	\includegraphics[height = 0.30 \linewidth]{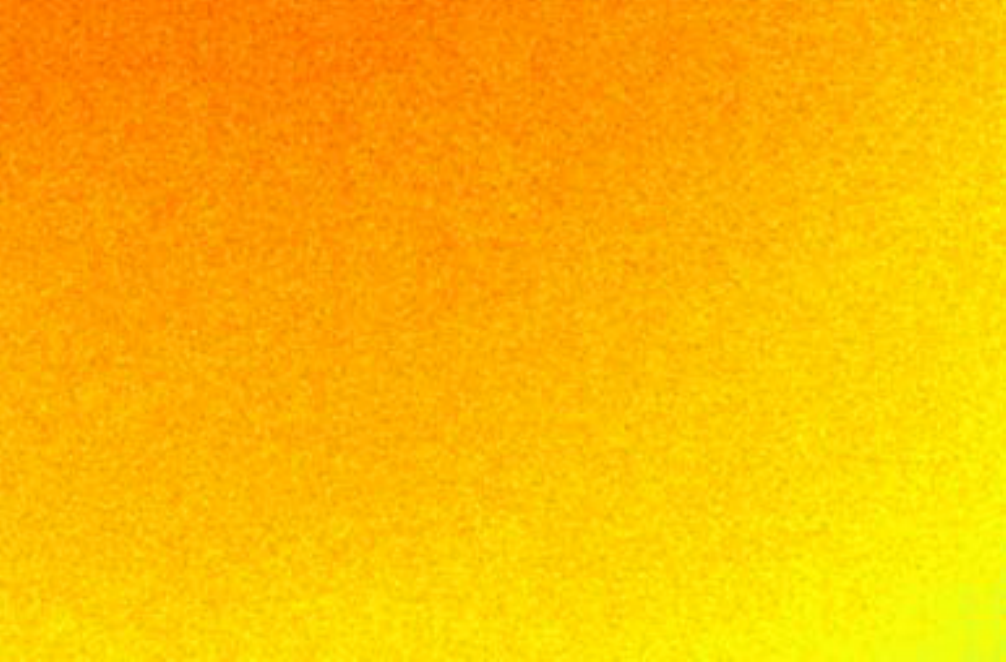}
	\includegraphics[height = 0.30 \linewidth]{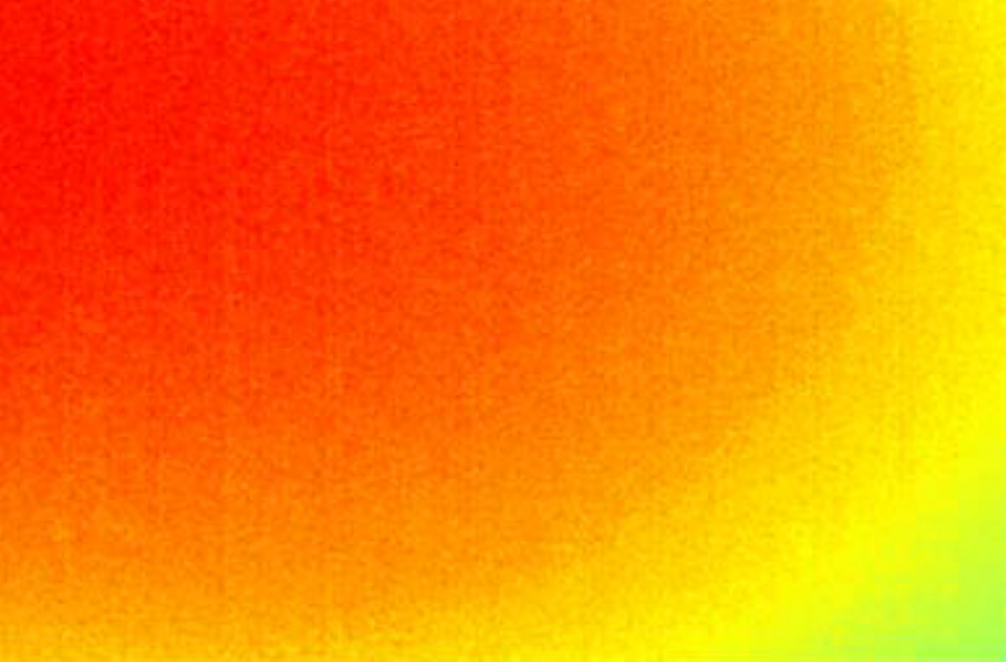}
	\includegraphics[height = 0.30 \linewidth]{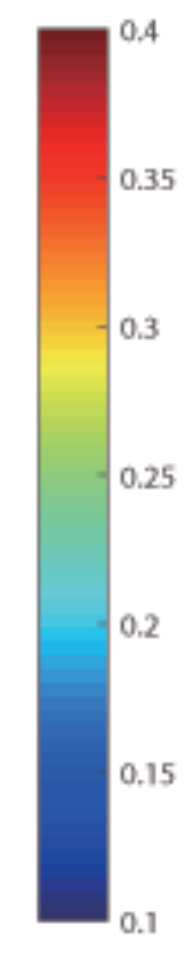}
	\caption{Color coded lighting intensity captured by our method.
			Left: lighting intensity at a plane which is about $1000$ millimeters away from the light.
			Right: lighting intensity of a plane $500$ millimeters away.}
	\label{fig:light intensity_distribution}
\end{figure}

\subsection{Capture Setups}
\label{sec:setups}

\textbf{A Studio Scanner Setup.}
As shown in the left of \figref{device_calibration}, $72$ LEDs are uniformly distributed on two concentric circles of diameter $400$ and $600$ millimeters respectively.
A video camera is mounted at the center of these circles, facing the direction perpendicular to the board\footnote{The camera was mounted manually.
It might not exactly sit on the circle center. Its direction might also be slightly off. We ignored these two factors as they introduce negligible errors according to our simulations.}.
The camera is synchronized with the LED lights such that at each video frame, there is only one light on.
At each viewpoint, we captured 30 images with different lighting directions in 12 seconds (at 2.5fps).
(Please refer to \secref{results_quant} for a justification of the number of images per viewpoint.)

\textbf{A Desktop Scanner Setup.}
We also build a desktop scanner with a ring of LED lights as shown in the middle of \figref{device_calibration}.
The design is almost the same as that of the studio setup, with a smaller footprint.
The diameters of the two LED circles are $150$ and $300$ millimeters respectively.
The object is usually $400$ millimeters away from the camera, such that effects of perspective cameras and point light sources must be considered.
After capturing images at one viewpoint, the turntable will rotate 10 degrees to capture the next viewpoint.

\subsection{Calibration}
\label{sec:device_calibration}
We assume orthographic camera model for the studio setup to simplify the computation,
and apply the sub-division scheme in \secref{perspective} to model perspective camera effects only for the desktop scanner setup.

\textbf{Lighting Directions.}
For the studio and desktop scanner setups, since the LEDs are uniformly distributed on circles with known radius, we only need to calibrate one parameter $\theta_0$ to determine their precise 3D positions.
Here, $\theta_0$ is the reference angle of the first LED light as shown in \figref{device_calibration}. To calibrate the angle $\theta_0$, we capture a diffuse board at some slanted positions (indicated as blue lines in \figref{light_calibration}) and compute the azimuth angle of the board's normal direction.
The computed azimuth angle should be $\theta_0 +\alpha$, where $\alpha$ is the true azimuth angle.
The angle $\alpha$ can be computed separately, by computing the 3D position of the board from a checkerboard calibration pattern.
Hence, we can obtain $\theta_0$ by subtracting $\alpha$ from the initial estimated azimuth angle.
When computing azimuth angles, we performed a Delaunay triangulation based interpolation method as introduced in the conference version \cite{Zhou2013}.
In the following, we describe the details of calibrating intensities of light sources for the studio and desktop scanners.

\textbf{Lighting Intensities.}
To calibrate lighting intensities, we capture a diffuse board roughly parallel to the image plane at multiple depths as shown in \figref{light_calibration},
where the black lines indicate the board positions in a bird view.
The nearest board and the farthest board enclose the lighting intensity volume.
A checkerboard calibration pattern is printed at the four corners of the board, such that its 3D position can be computed.
Assume the board is Lambertian with unit albedo (we used the X-Rite white balance board in our experiments).
At each point, the observed pixel intensity should be $I = \ve{n}^{\top}\ve{l} V$,
were $\ve{l}, \ve{n}$ are the local lighting and normal directions, and $V$ is the light intensity.
Hence, we can capture $V$ at each point on the board as $I/\ve{n}^{\top}\ve{l}$.
We linearly interpolate these captured values to obtain the result in a continuous 3D volume.

We empirically find that, when the distance between the object and the light is around $1000$ millimeters (e.g., the studio setup in \secref{Results}),
we only need to compute the mean lighting intensity of each LED light for all pixels.
This can be seen from the uniform intensity distribution from the left of \figref{light intensity_distribution}.
When the distance is around $400$ millimeters (e.g., the desktop scanner setup in \secref{Results}),
we have to record a different lighting intensity for each pixel for an LED.
This is evident from the right of \figref{light intensity_distribution}.

\begin{table*}
		\caption{Shape reconstruction errors of the objects shown in \figref{all_results} (in millimeters).}  \label{tab:errors}
	\vspace{-0.4cm}
	\begin{center}
		\begin{tabular}{c|ccccccc}
			\hline \hline
			Objects  & \emph{Buddha-S} &  \emph{Teapot2-S}  & \emph{Teapot3-D} & \emph{Gourd-D}  & \emph{Cat2-D} &  \emph{Pig-D}\\
			\hline
			diameter of the object    & 140  & 140 & 130  & 70  & 180 & 140  \\
			\hline
			number of viewpoints   & 35  & 35  & 36 & 36 & 36   &  36   \\
			\hline
			median error      & 0.24 & 0.28 & 0.25 & 0.19 & 0.42 & 0.39  \\
            \hline
			mean error  & 0.53 & 0.49 & 0.49 & 0.36 & 0.76 &  0.57  \\
			\hline \hline
		\end{tabular}
	\end{center}
\vspace{-0.3cm}
\end{table*}

\begin{figure*}
	\centering
	\begin{tabular}{@{\hspace{1mm}}c@{\hspace{1mm}}c@{\hspace{1mm}}c@{\hspace{1mm}}c@{\hspace{1mm}}c}
		\includegraphics[height= 0.32 \linewidth]{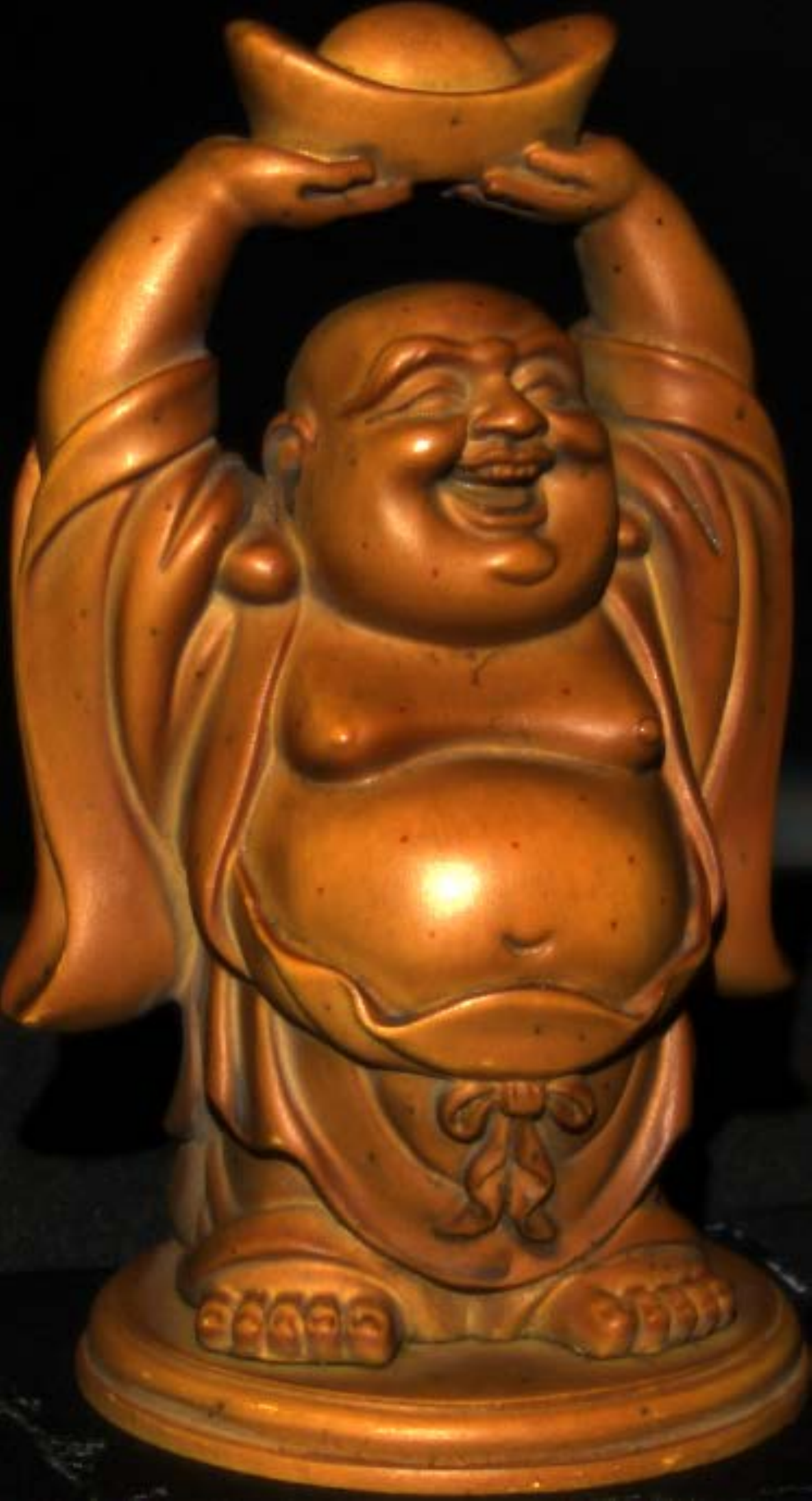}
		& \includegraphics[height= 0.32 \linewidth]{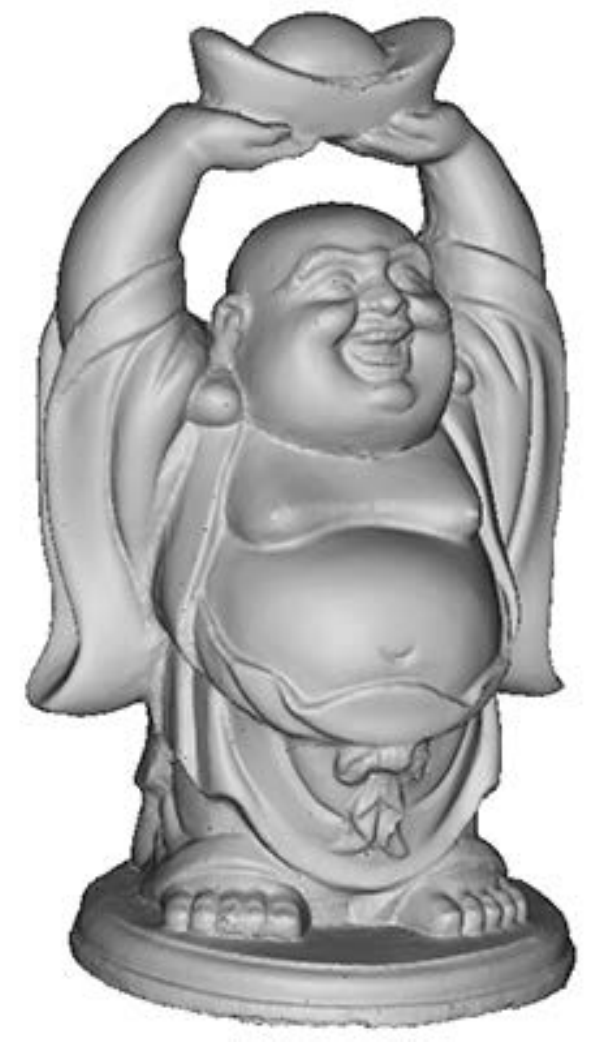}
		& \includegraphics[height= 0.32 \linewidth]{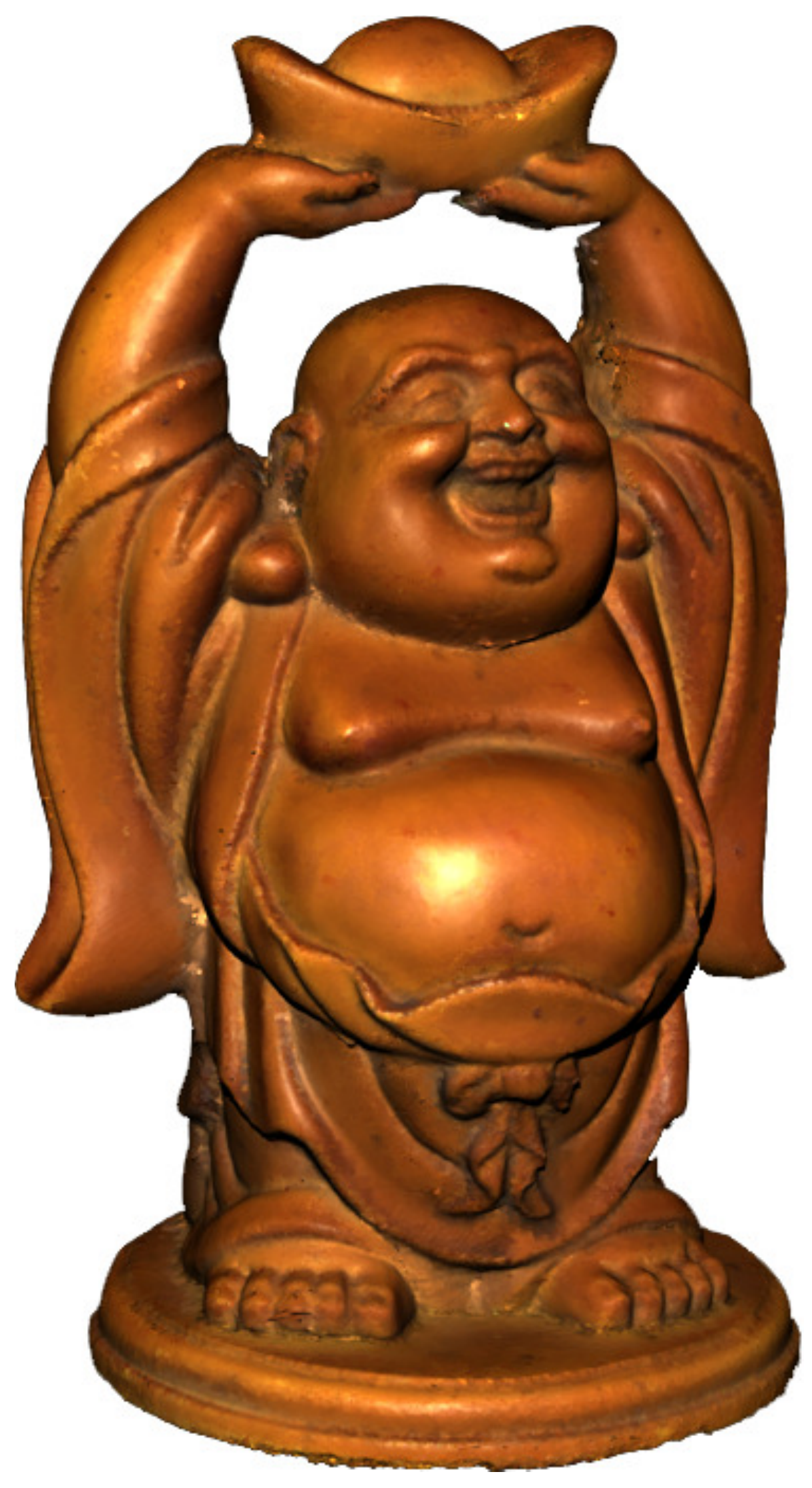}
		& \includegraphics[height= 0.31 \linewidth]{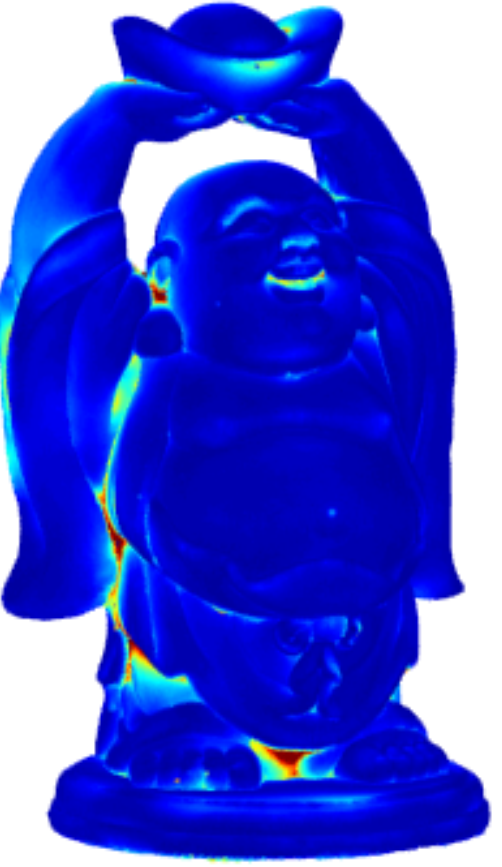}
		&  \includegraphics[height= 0.32 \linewidth]{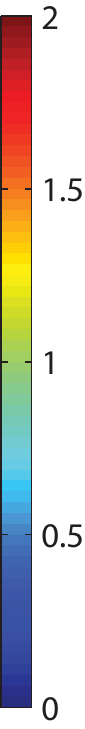}
	 \\
		\includegraphics[height= 0.155 \linewidth]{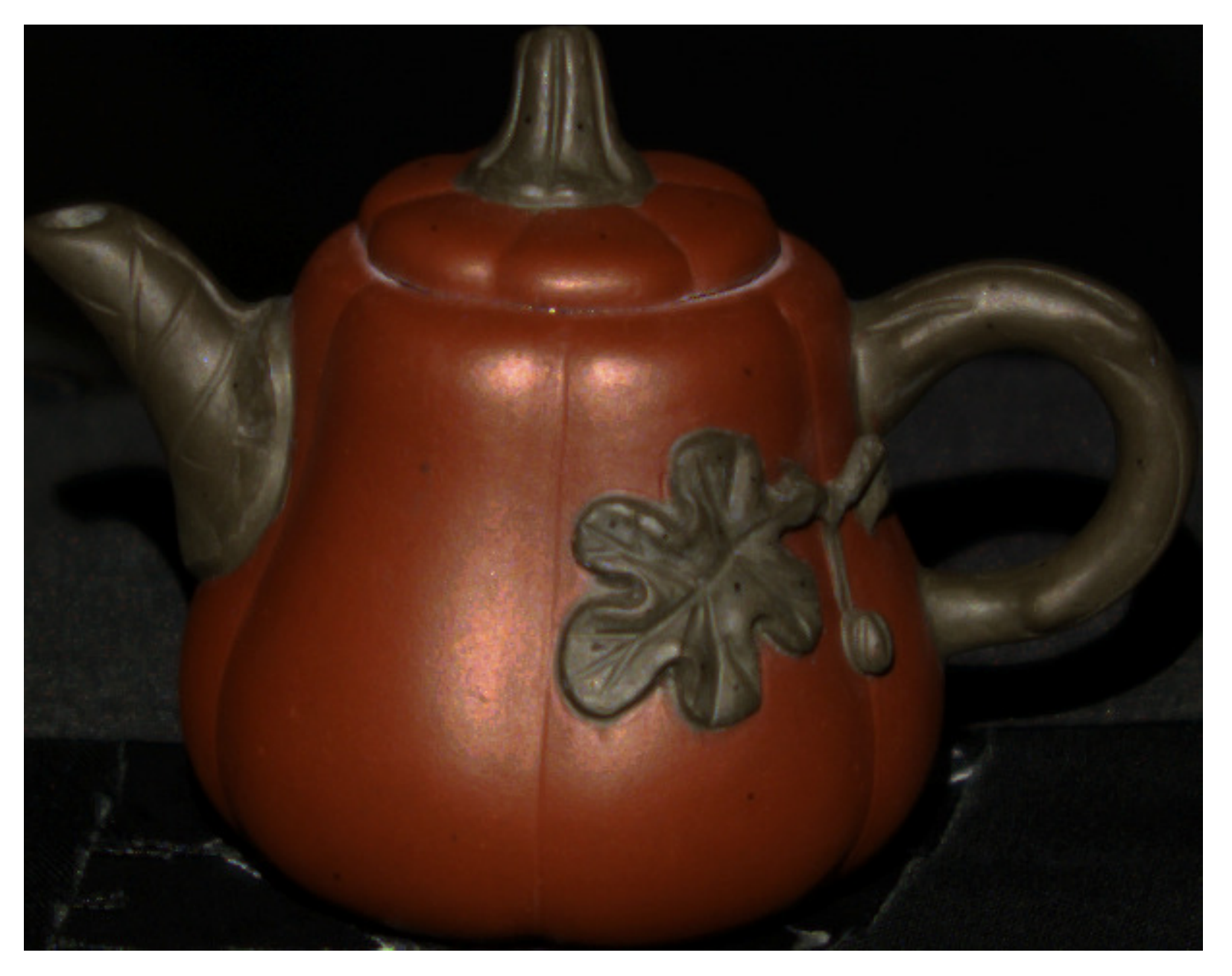}
       & \includegraphics[height= 0.155 \linewidth]{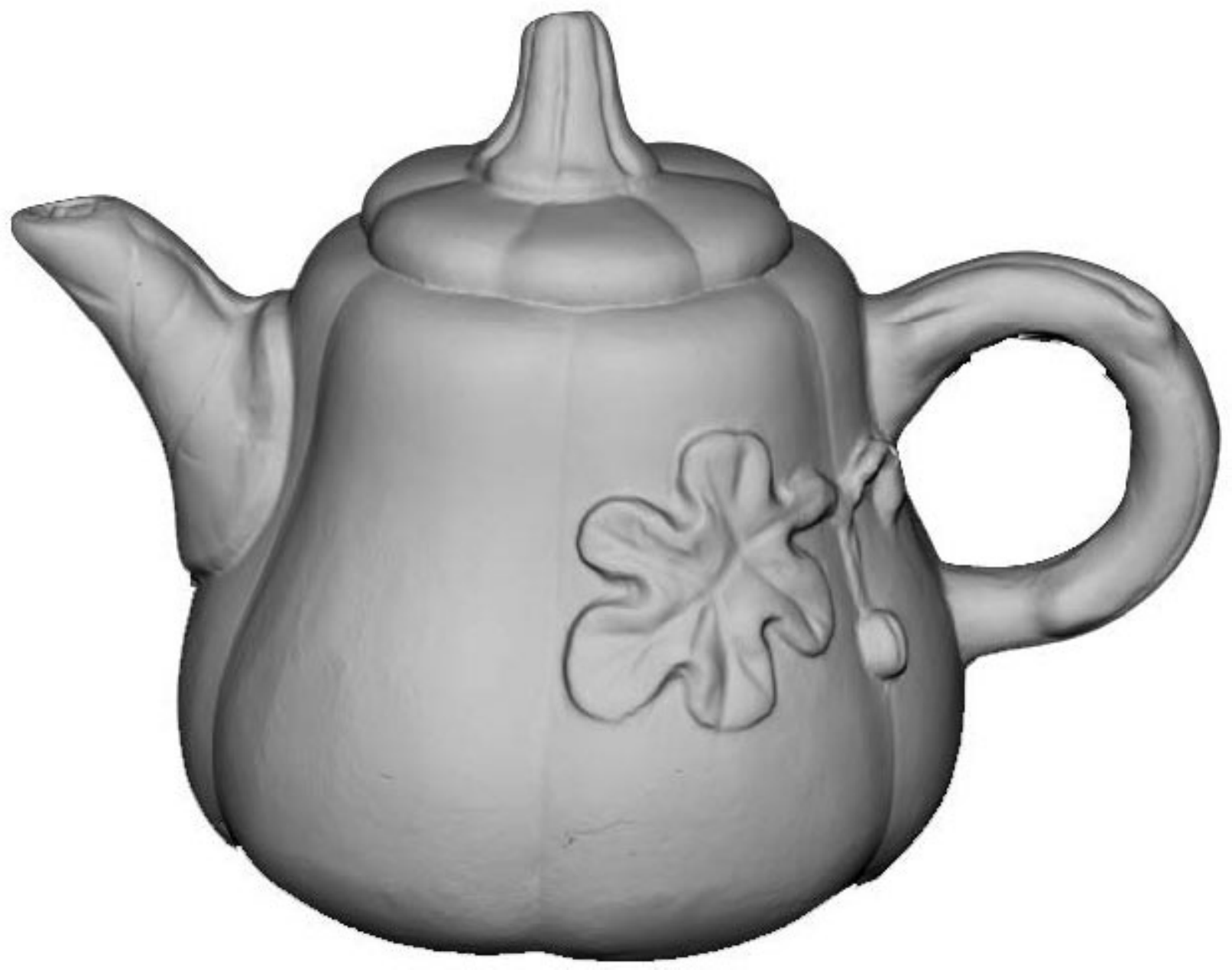}
       & \includegraphics[height= 0.155 \linewidth]{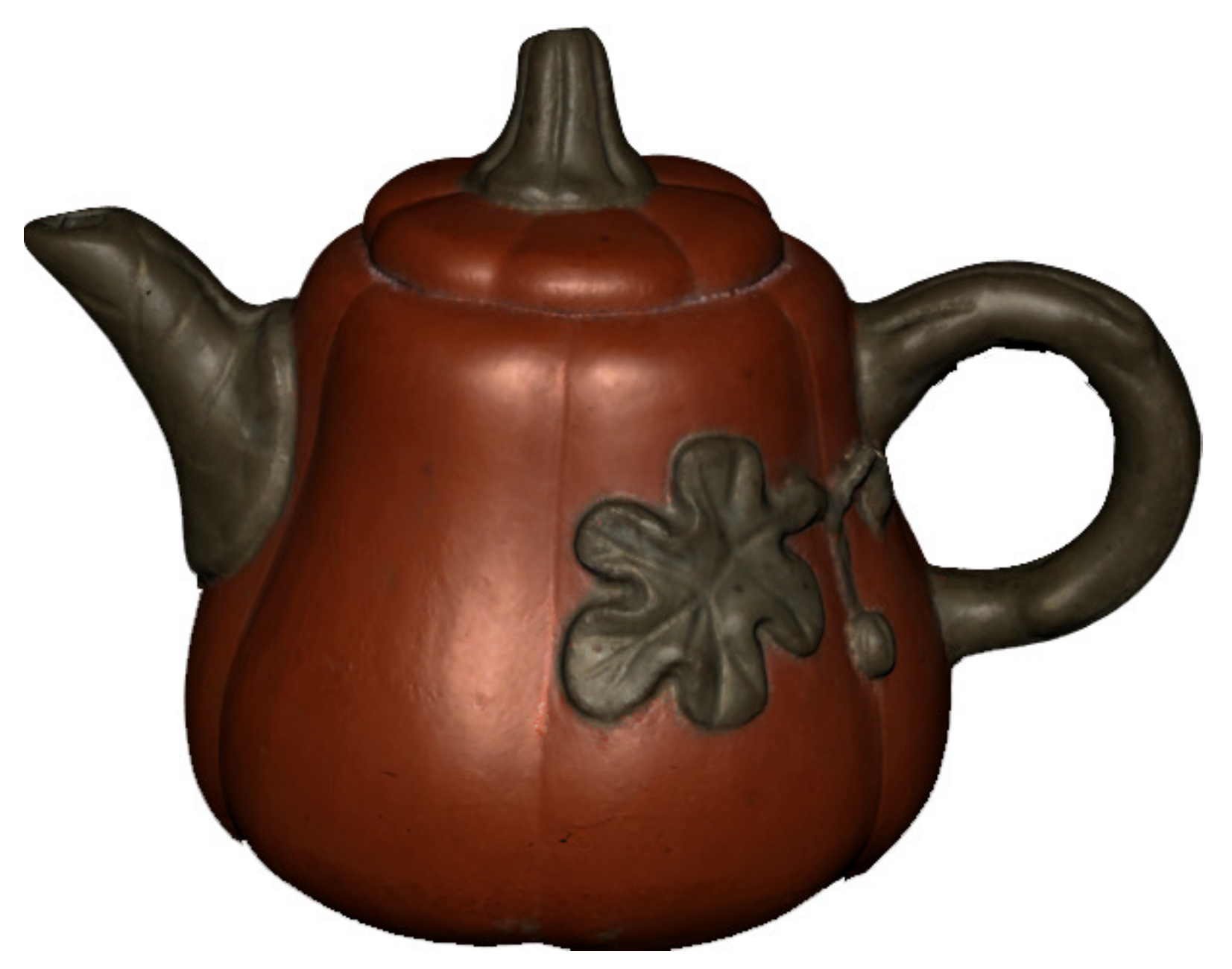}
       & \includegraphics[height= 0.155 \linewidth]{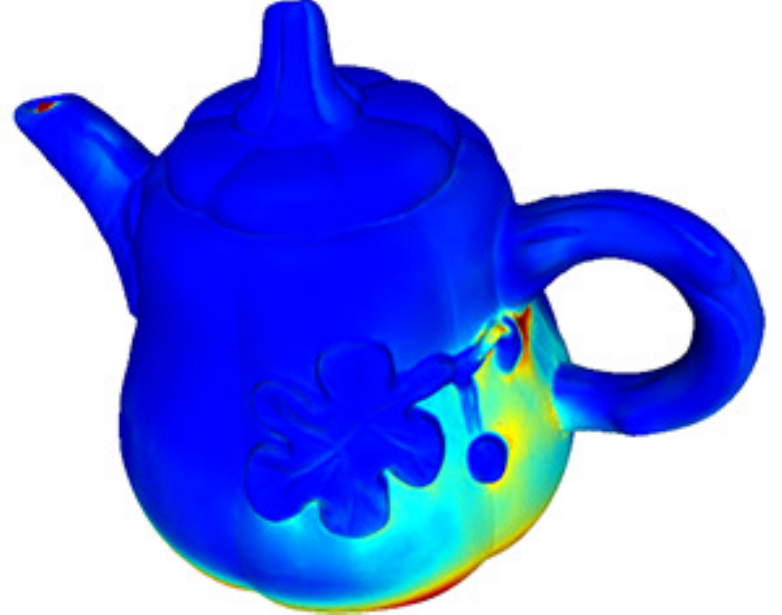}
       \\
		\includegraphics[height= 0.13 \linewidth]{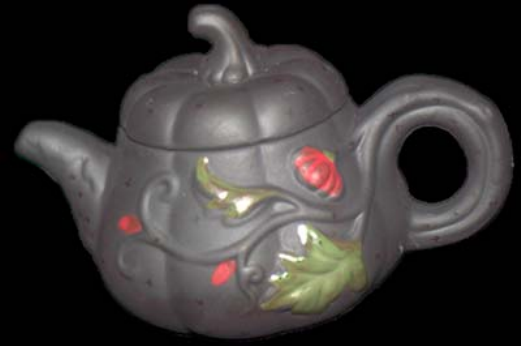}
		& \includegraphics[height= 0.14 \linewidth]{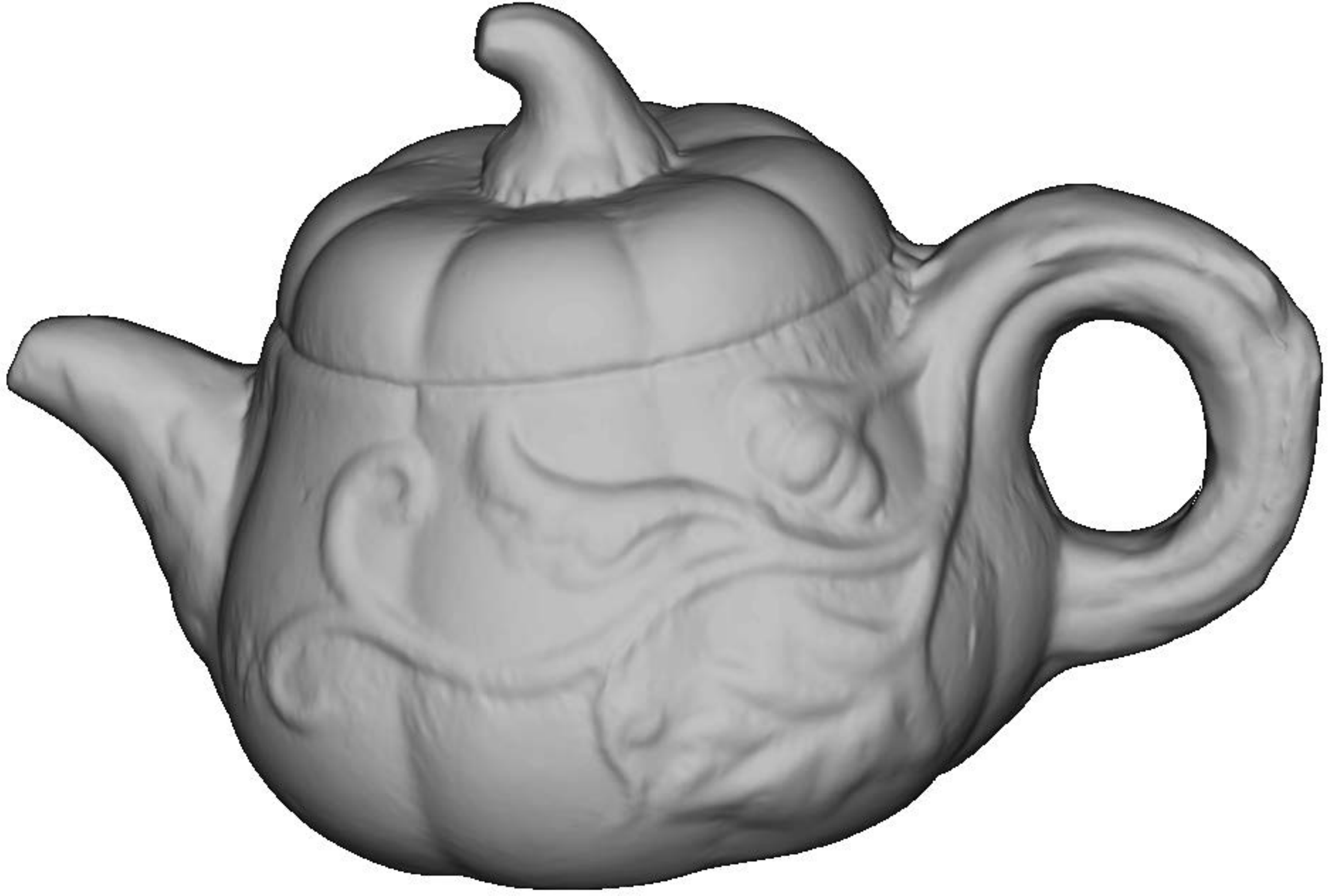}
		& \includegraphics[height= 0.14 \linewidth]{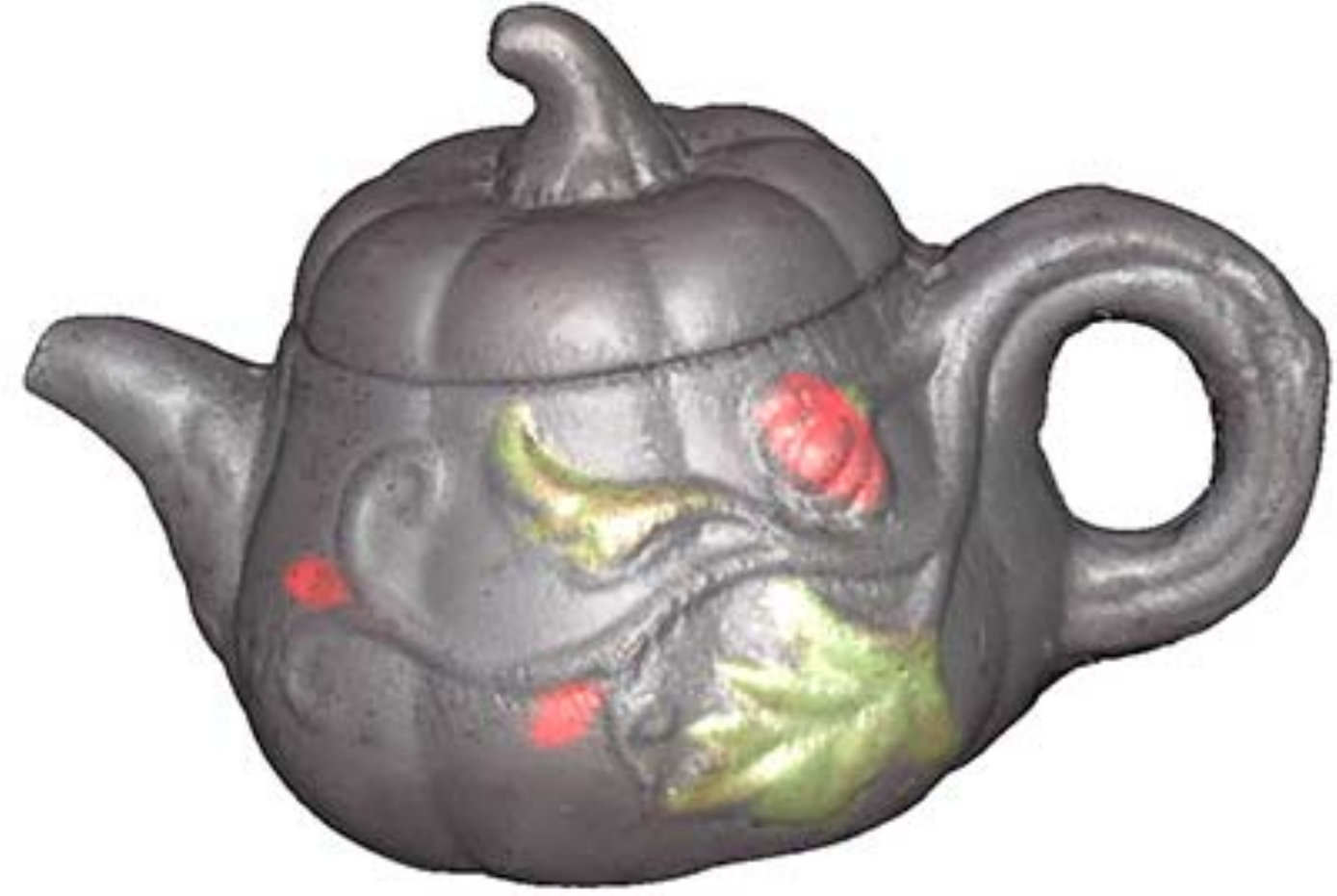}
		&\includegraphics[height= 0.14 \linewidth]{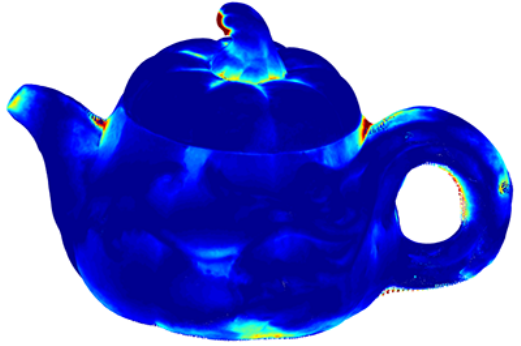}
		\\
		\includegraphics[height= 0.24 \linewidth]{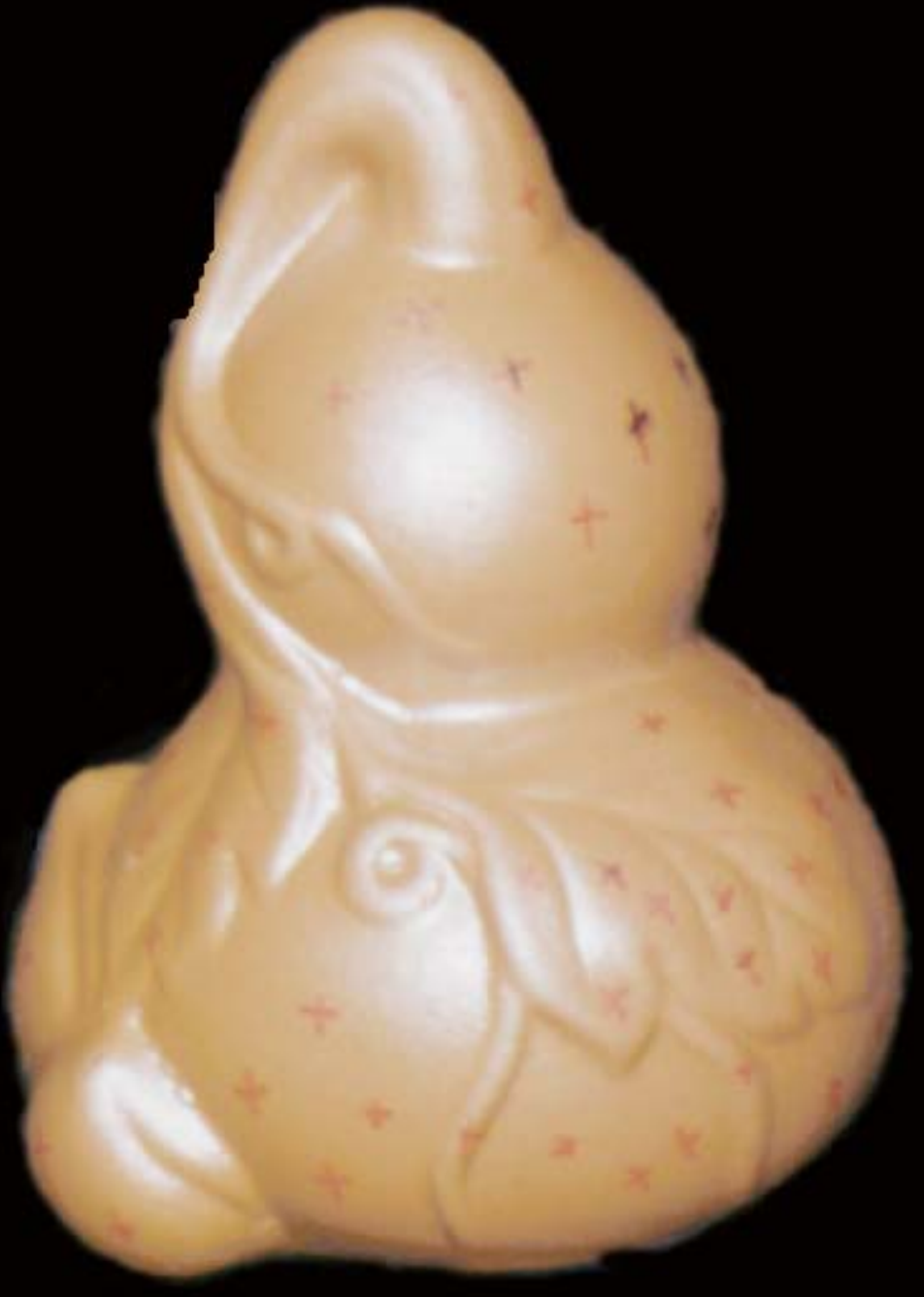}
		& \includegraphics[height= 0.24 \linewidth]{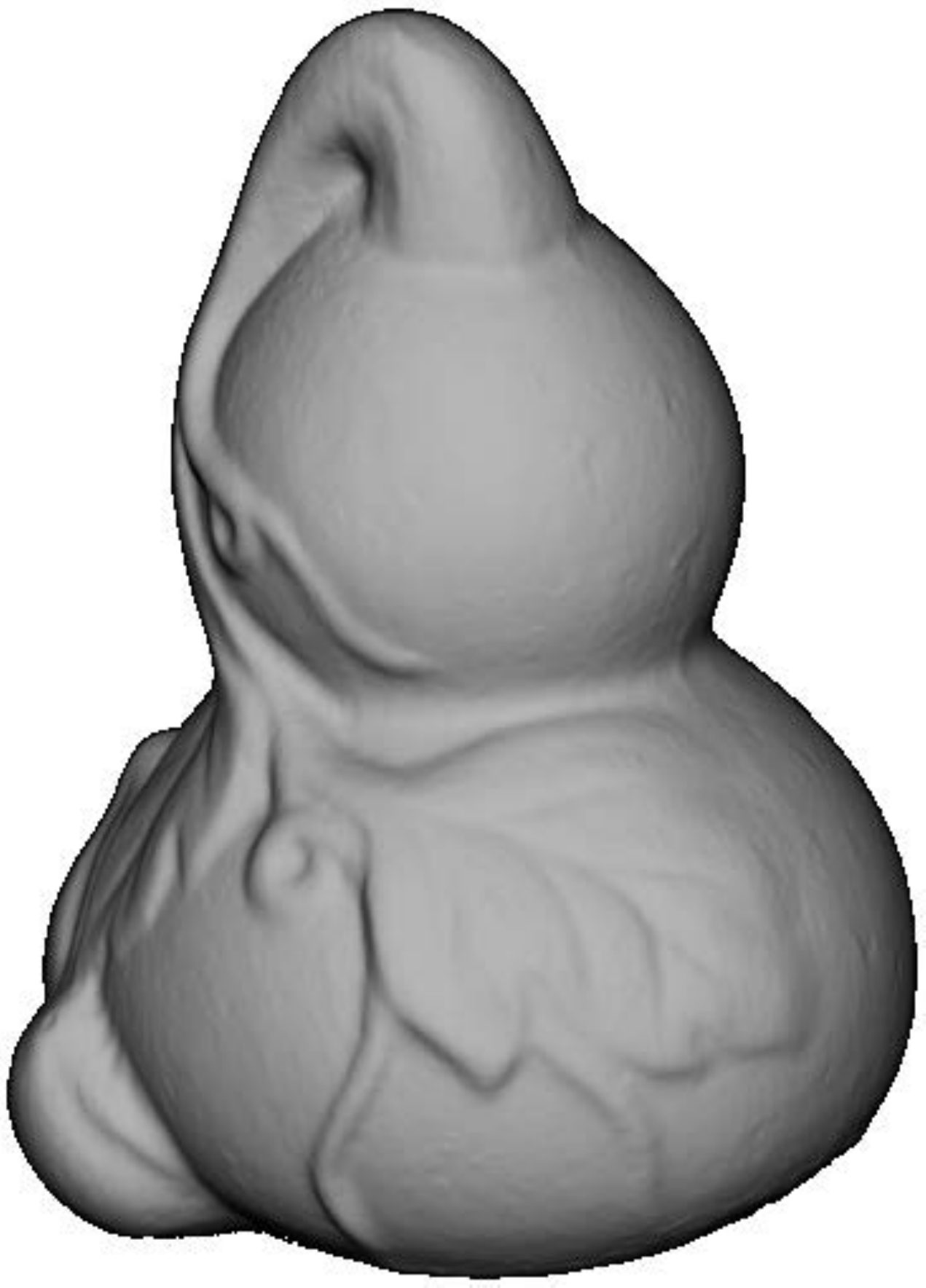}
		& \includegraphics[height= 0.24 \linewidth]{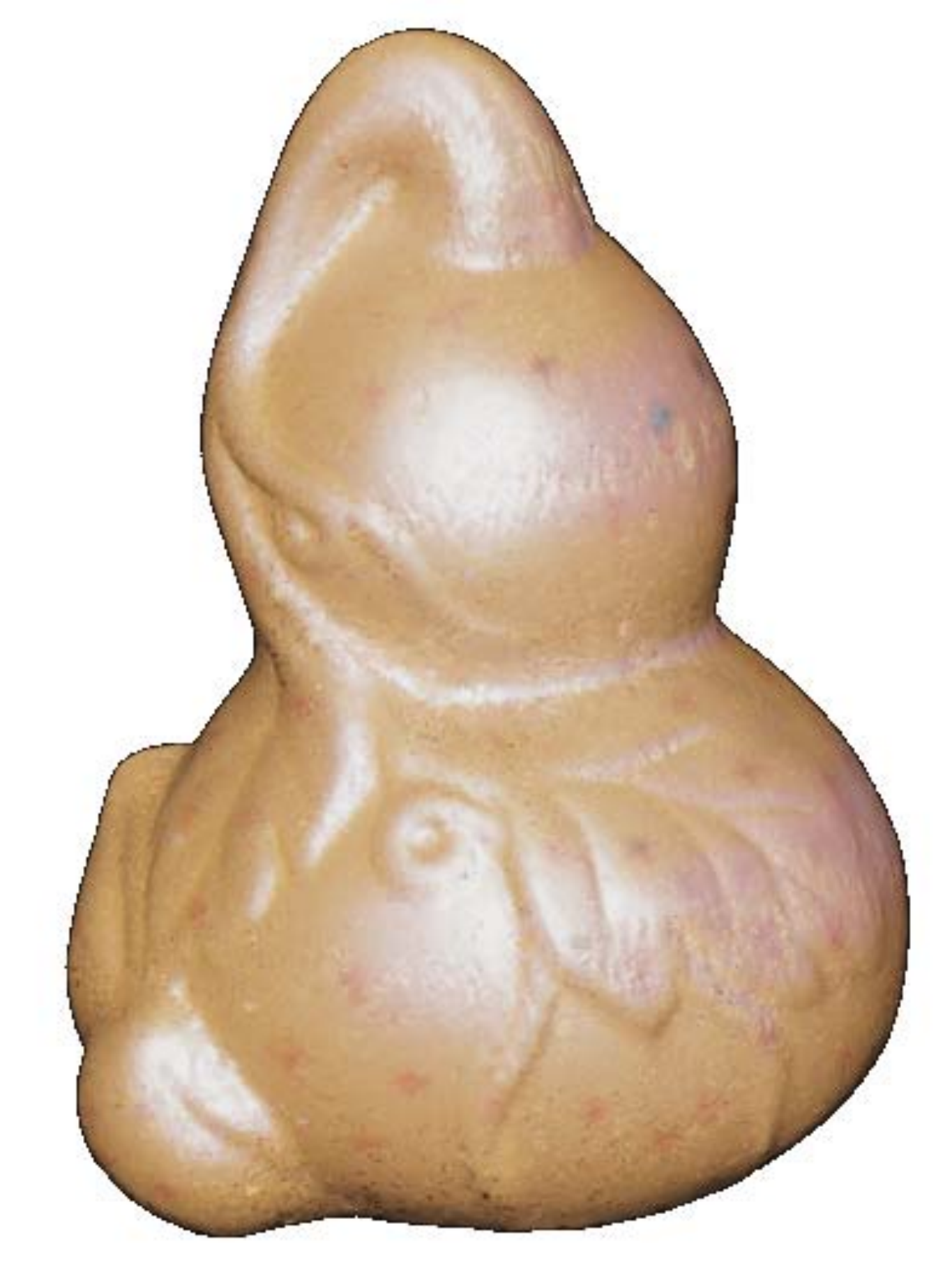}
		&\includegraphics[height= 0.24 \linewidth]{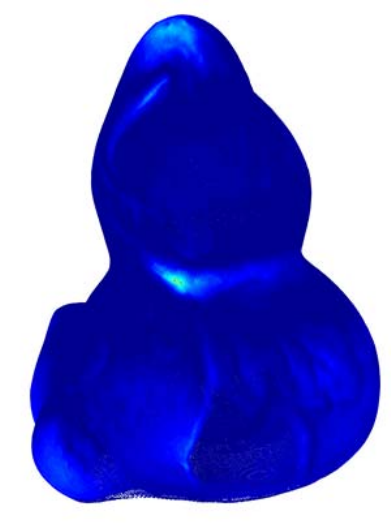}
		\\
		\includegraphics[height= 0.28 \linewidth]{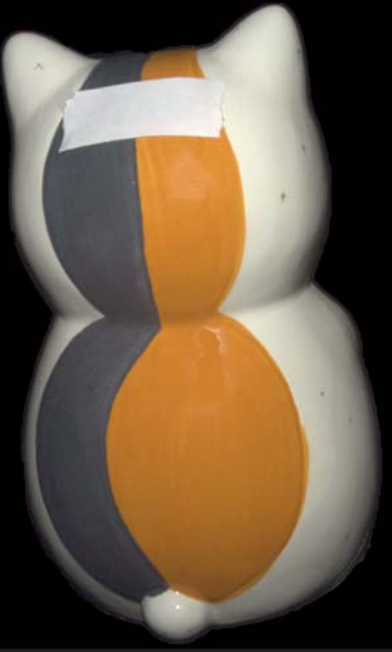}
		& \includegraphics[height= 0.28 \linewidth]{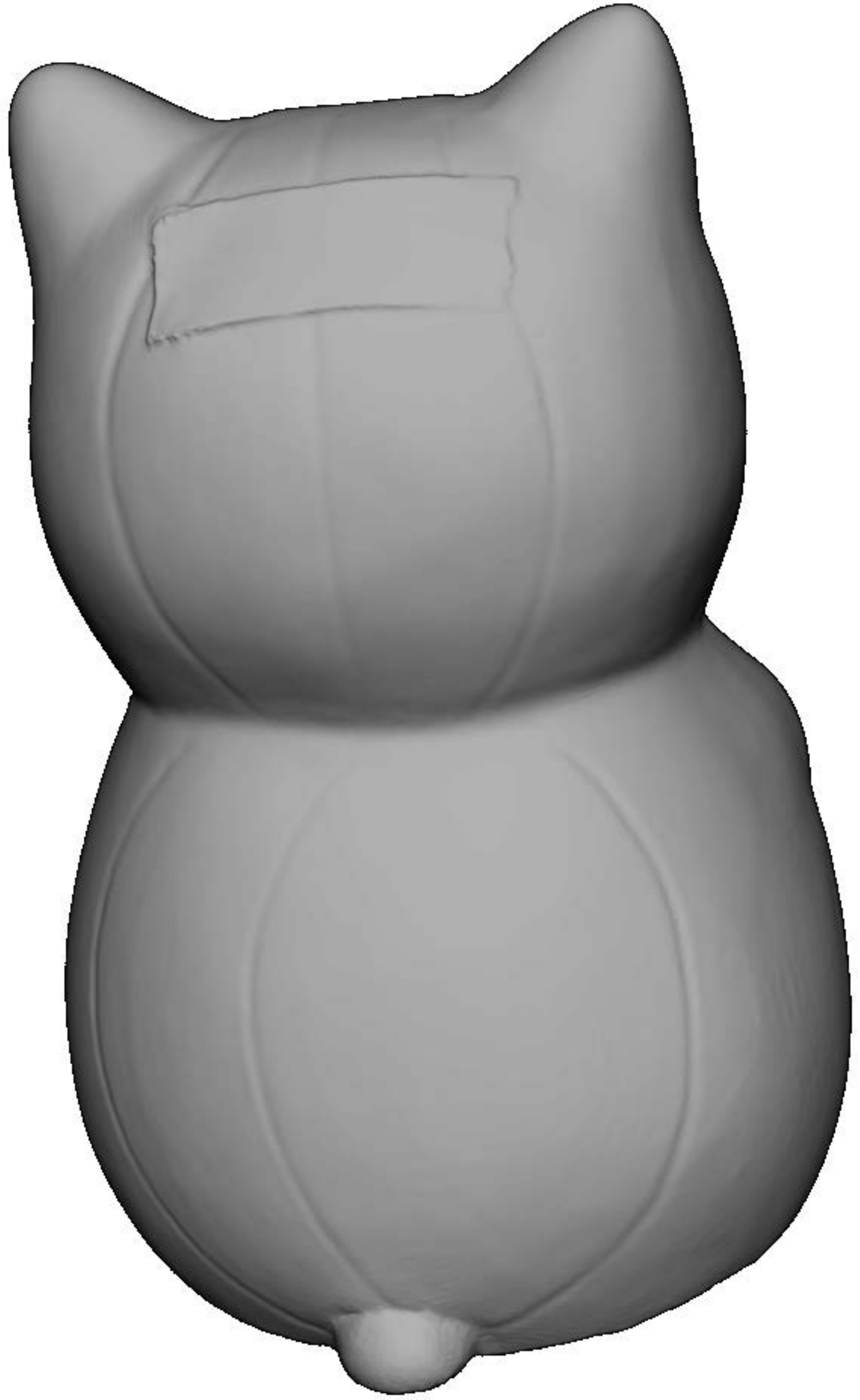}
		& \includegraphics[height= 0.28 \linewidth]{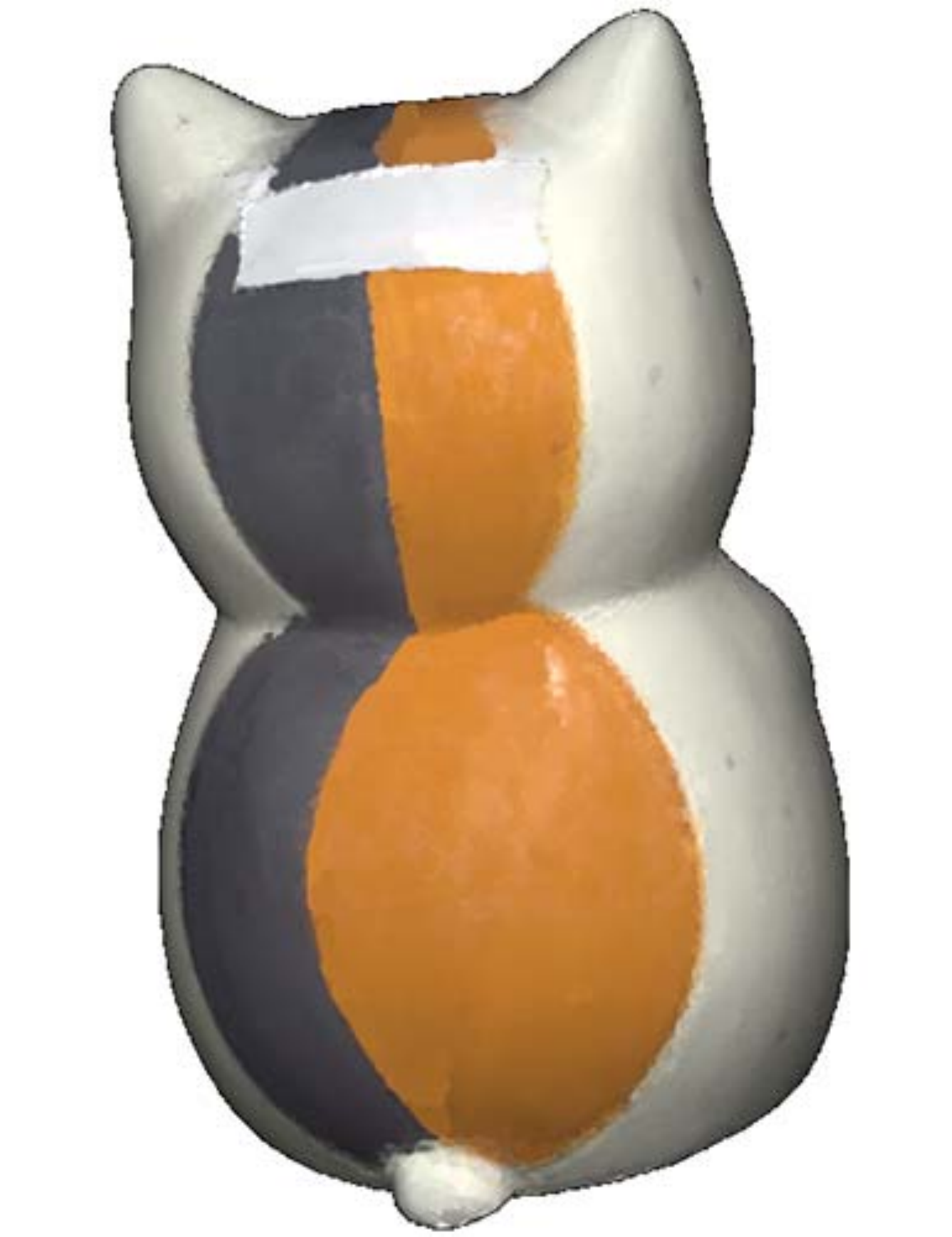}
		&\includegraphics[height= 0.27 \linewidth]{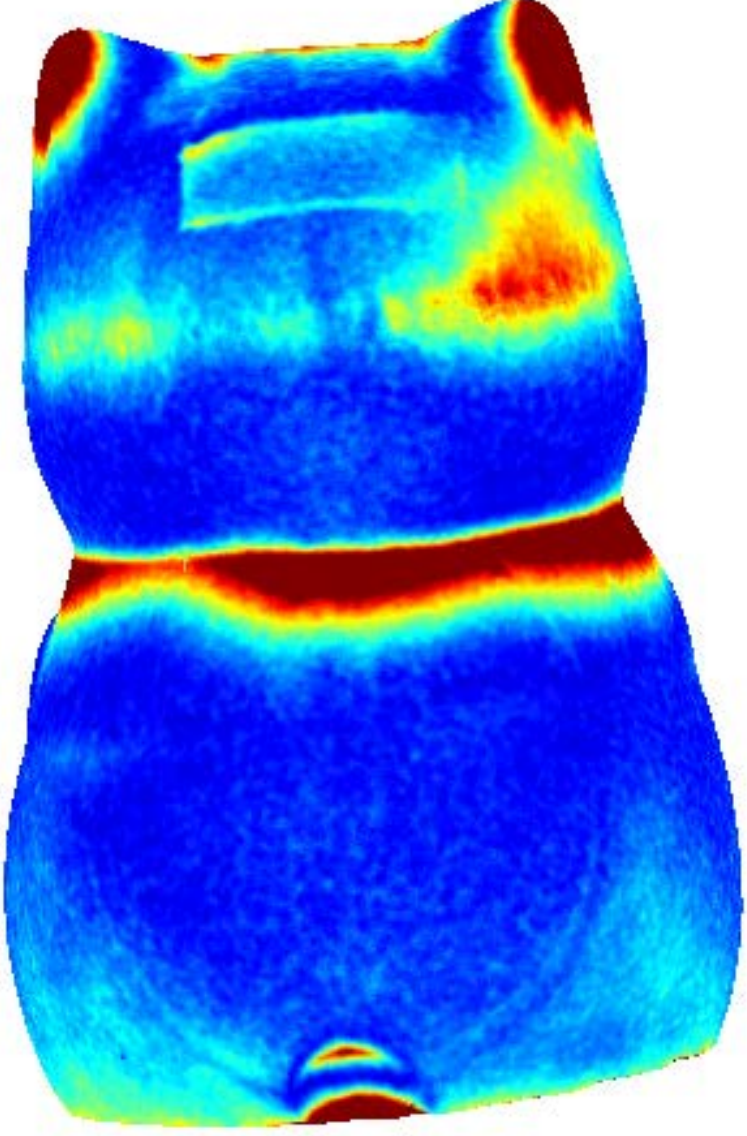}
		\\
		\includegraphics[height= 0.13 \linewidth]{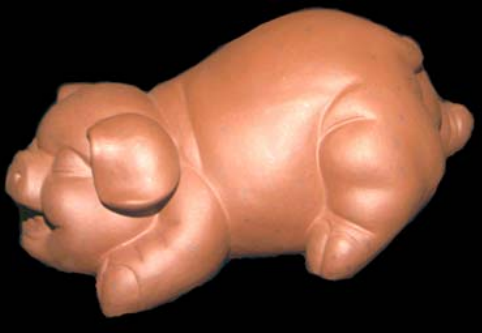}
		& \includegraphics[height= 0.13 \linewidth]{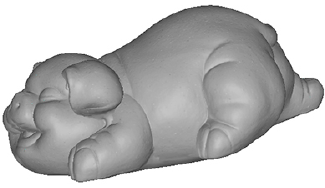}
		& \includegraphics[height= 0.14 \linewidth]{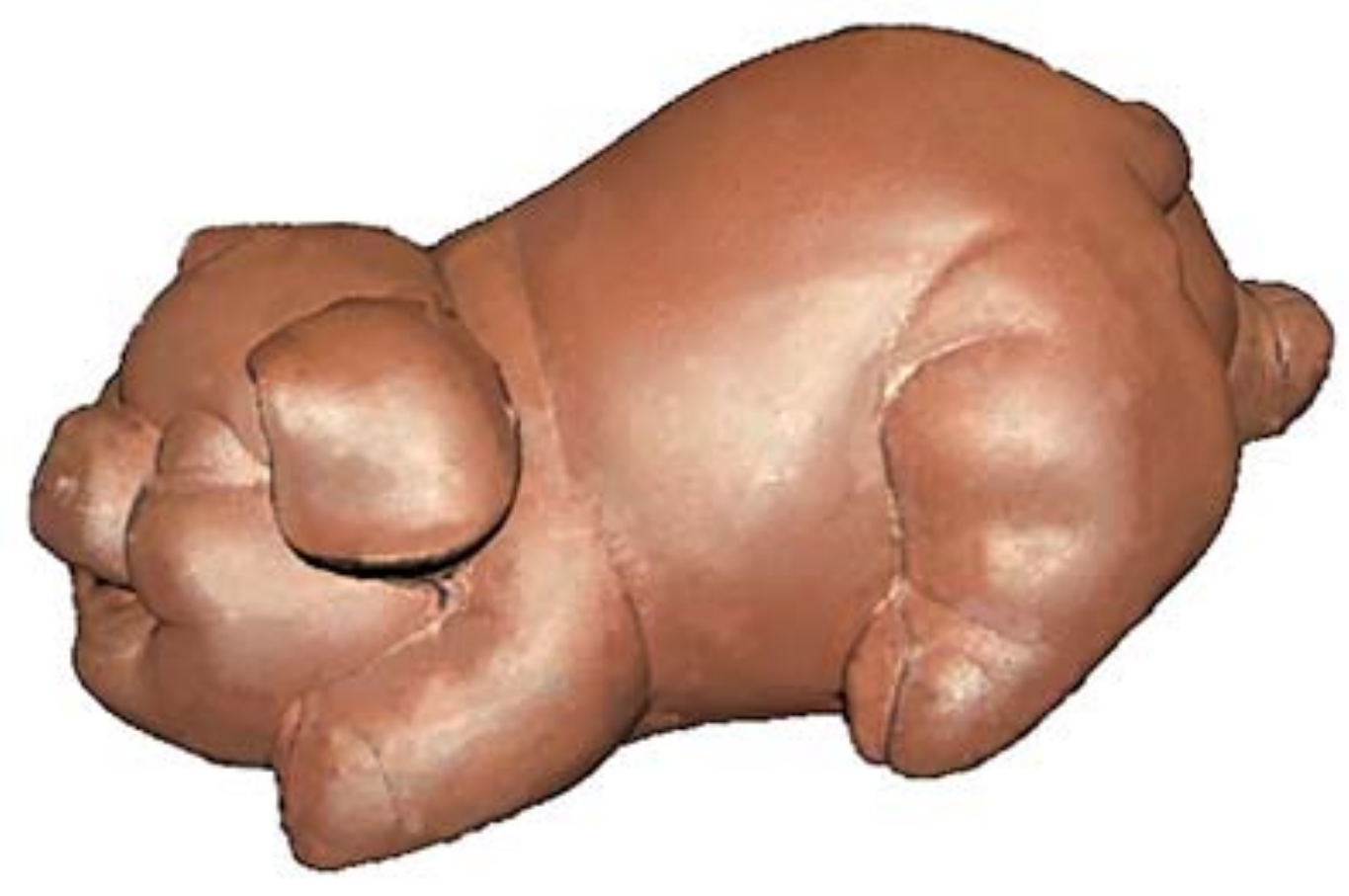}
		&\includegraphics[height= 0.14 \linewidth]{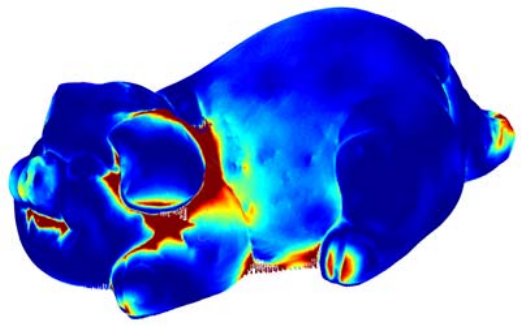}
       \\
(a) & (b) & (c) & \multicolumn{2}{c}{(d)}
\end{tabular}
	\caption{Results from the different data capture setups. (a) one of the input
	images; (b) the recovered shape rendered with uniform diffuse
	shading; (c) a rendering with the recovered reflectance model
	from the same viewpoint and lighting condition as the image in (a); (d) the color-coded shape error (in millimeters) compared
	to laser-scanned `ground truth'. The 1st-2nd rows are from the studio setup  and the 3th-6th rows are from the desktop setup.
}\label{fig:all_results}
\end{figure*}

\section{Experiments}
\label{sec:Results}
\subsection{Evaluation Using Data Capture by Our Setups}
In our experiments, the 3D points obtained from the structure-from-motion algorithm were often noisy.
We only keep points with reprojection error less than $0.5$ pixels.
Typically, about 200 initial points are obtained for each example.
Our system can also easily incorporate manual intervention in the form of matched feature points to handle textureless regions.
To provide a `ground truth' validation, all experimental objects were scanned using a Rexcan III industrial scanner, which is accurate to 10 microns.
Our results are registered with the scanned shapes using the iterative closest point (ICP) algorithm \cite{Besl1992}.

\subsubsection{Quantitative Shape Reconstruction Errors}
Some examples are provided in \figref{all_results} with a sample input image shown in (a) for each example.
From top to bottom, we refer these examples as `\emph{Buddha-S}', `\emph{Teapot2-S}', `\emph{Teapot3-D}', `\emph{Gourd-D}', `\emph{Cat2-D}', and `\emph{Pig-D}' respectively for future reference,
where the suffix \emph{-S} and \emph{-D} stand for the studio and desktop setups respectively.
To better visualize the recovered shape, we render it with uniform diffuse shading in (b).
Most of the geometry details are successfully captured by our methods, as exemplified by the wrinkles of the `\emph{Buddha-S}' clothes.
\figref{all_results} (c) is a rendering according to the captured reflectance from the same viewpoint and lighting condition as the input image in (a).
The rendered image closely resembles the input image, indicating high accuracy in both geometry and reflectance.
To provide a quantitative evaluation on shape capture, we visualize the shape reconstruction error (measured in millimeters) in (d).
Typically, larger errors are associated with concavities, which have fewer image observations due to occlusion and are also affected by stronger inter-reflections.


The `\emph{Buddha-S}' and `\emph{Teapot2-S}' examples were captured with the studio scanner setup.
The `\emph{Buddha-S}' example contains many discontinuities at clothes folds and large concavities at the shoulder.
The polished wooden `\emph{Buddha-S}' has focused and strong highlight, while the clay `\emph{Teapot2-S}' has soft and extended highlight.
Our method consistently performed well on both of them.
The `\emph{Teapot2-S}' example has relatively larger error at one side, mainly due to the imprecise SfM reconstruction at that area.
We captured four examples by the desktop scanner, including `\emph{Teapot3-D}', `\emph{Gourd-D}', `\emph{Cat2-D}', and `\emph{Pig-D}'. In particular, the `\emph{Cat2-D}' and `\emph{Pig-D}' examples are captured with the desktop scanner setup spanning about 30 degrees FoV in the camera, with significant perspective camera effects.
They also present a variety of different materials, where the `\emph{Cat2-D}' has focused highlight and the `\emph{Pig-D}' has softer and more extended highlight.

\begin{figure*}\centering
	\begin{tabular}{@{\hspace{1mm}}c@{\hspace{1mm}}c@{\hspace{1mm}}c@{\hspace{1mm}}c@{\hspace{1mm}}c@{\hspace{1mm}}c}
		  \includegraphics[width=0.15\linewidth]{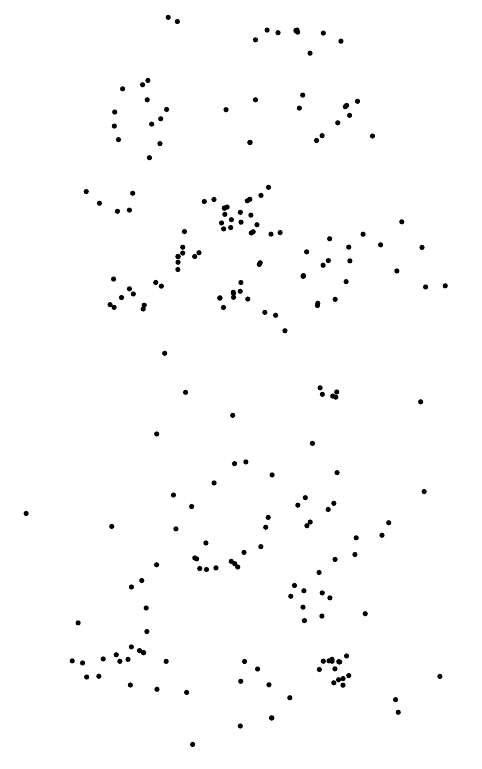}
		& \includegraphics[width=0.15\linewidth]{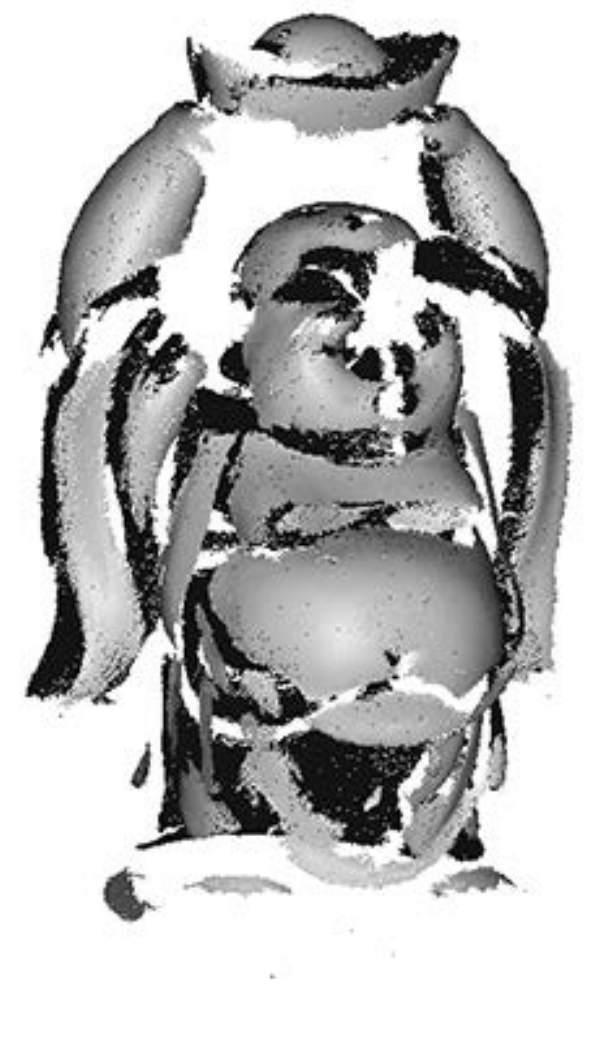}
		& \includegraphics[width=0.15\linewidth]{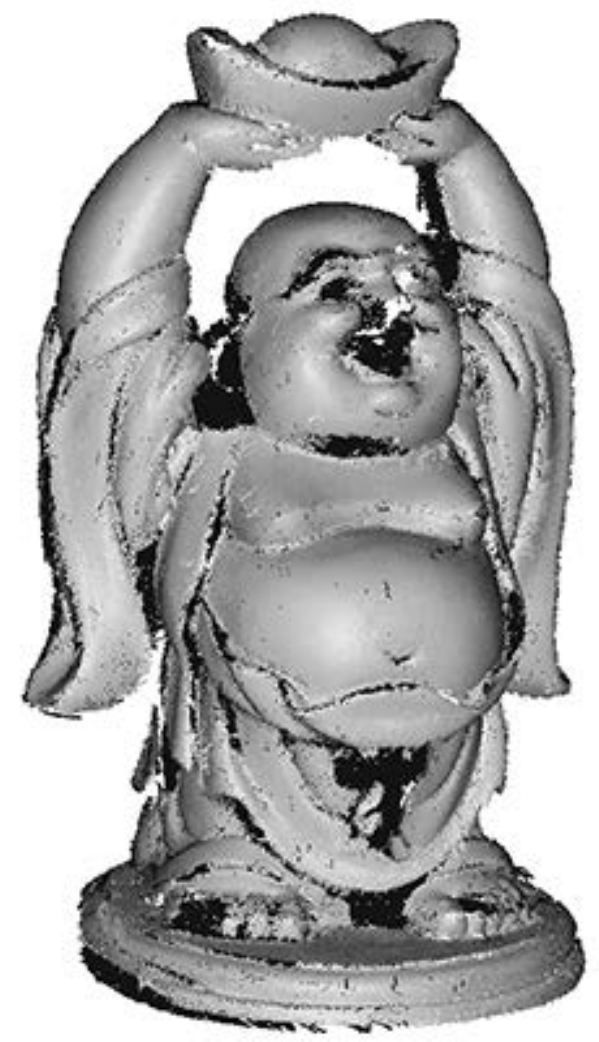}
		& \includegraphics[width=0.15\linewidth]{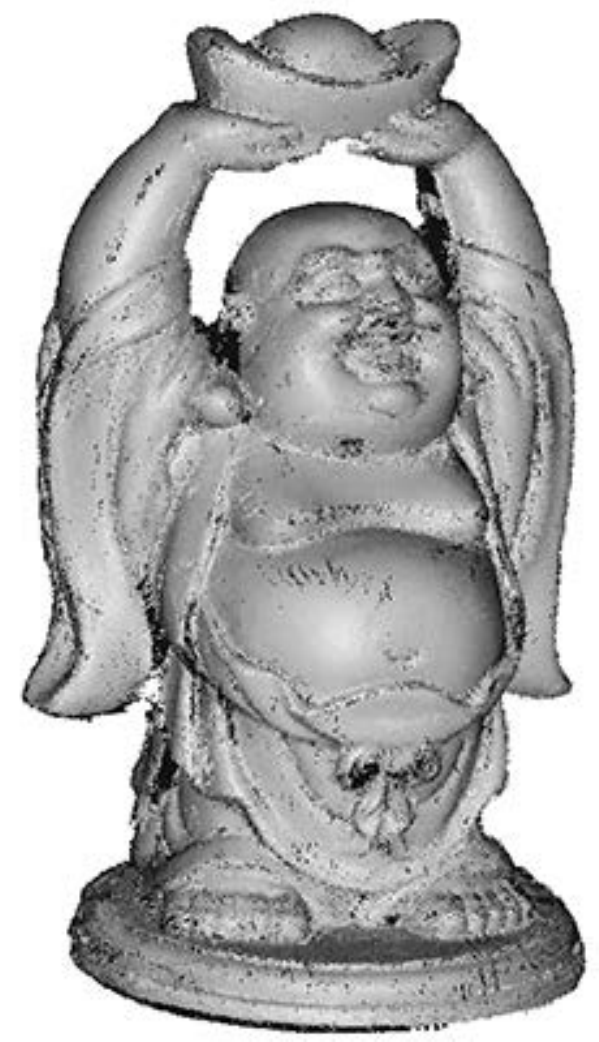}
		& \includegraphics[width=0.15\linewidth]{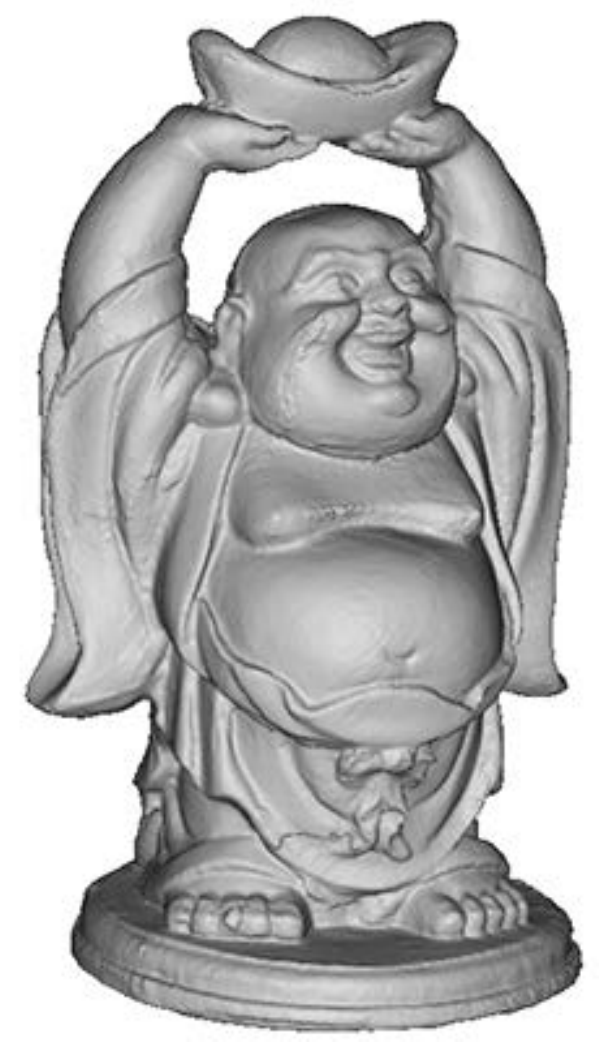}
        & \includegraphics[width=0.15\linewidth]{buddha/buddha_geometry.pdf}
		\\
		  \includegraphics[width=0.15\linewidth]{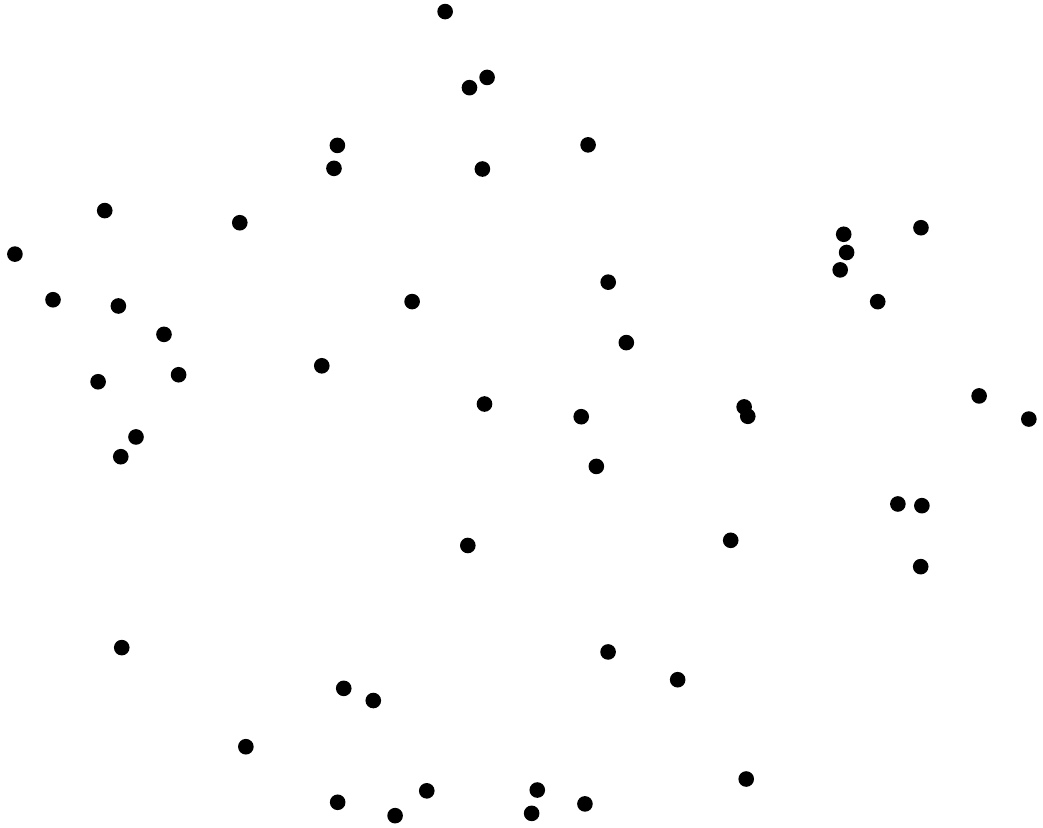}
		& \includegraphics[width=0.15\linewidth]{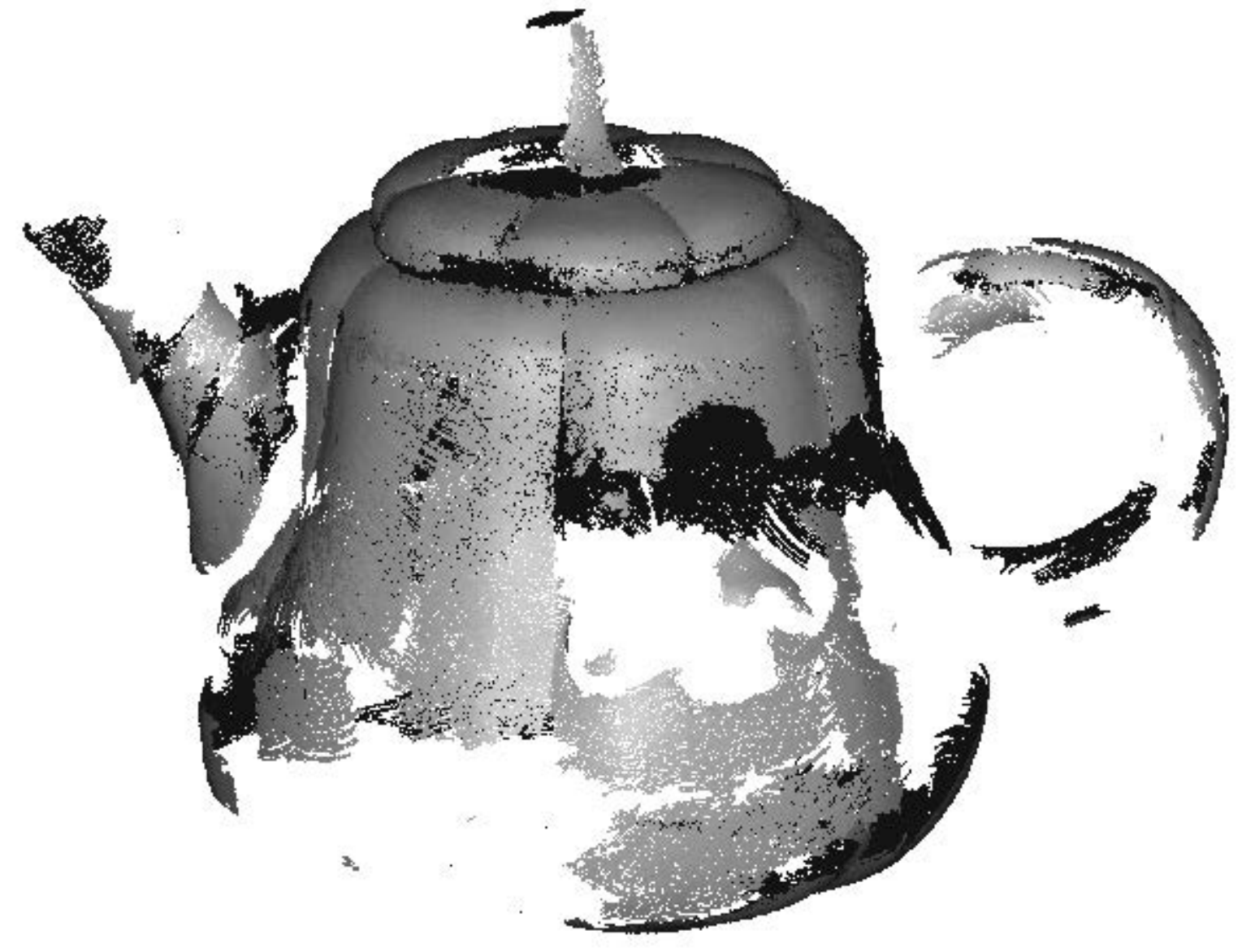}
		& \includegraphics[width=0.15\linewidth]{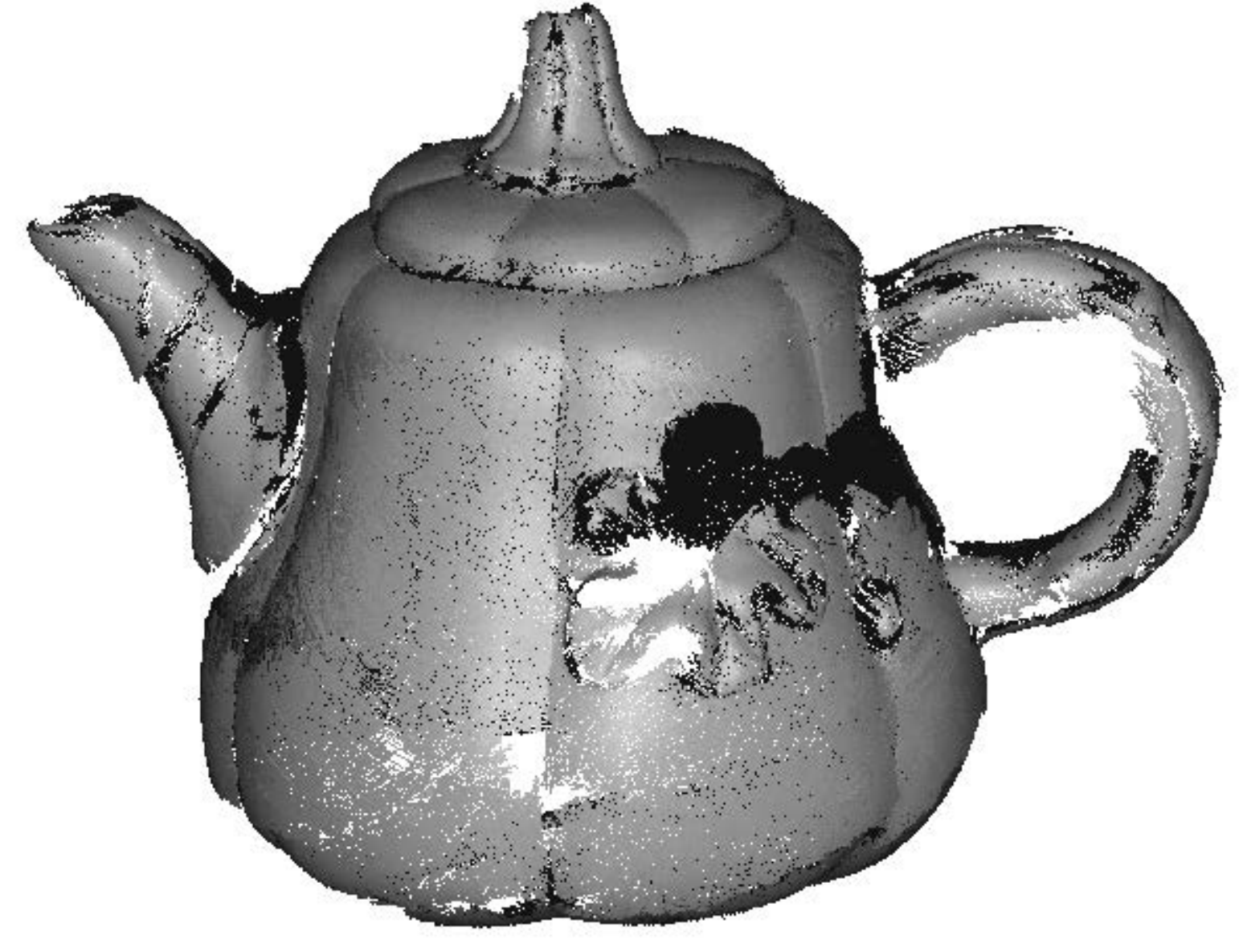}
		& \includegraphics[width=0.15\linewidth]{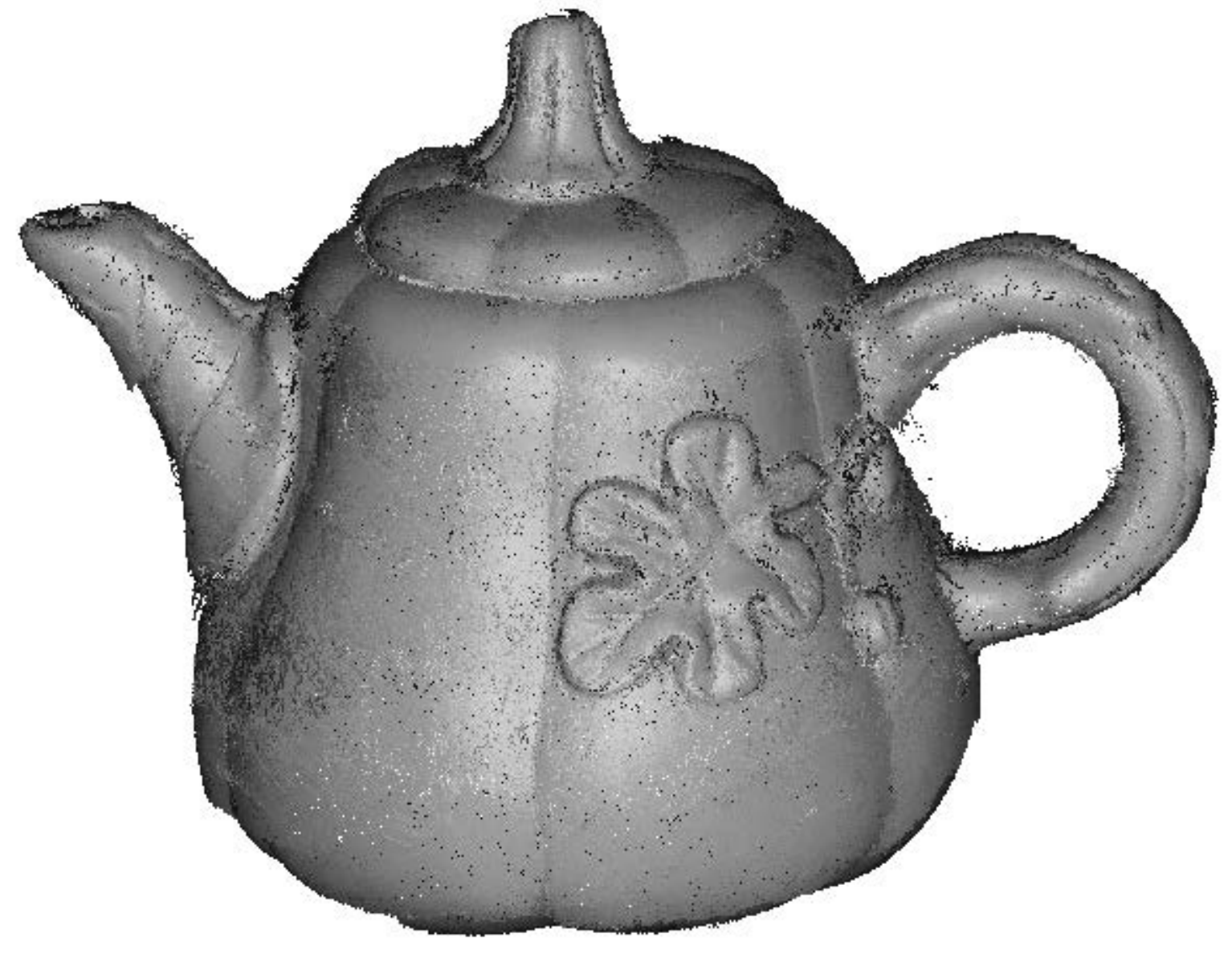}
		& \includegraphics[width=0.15\linewidth]{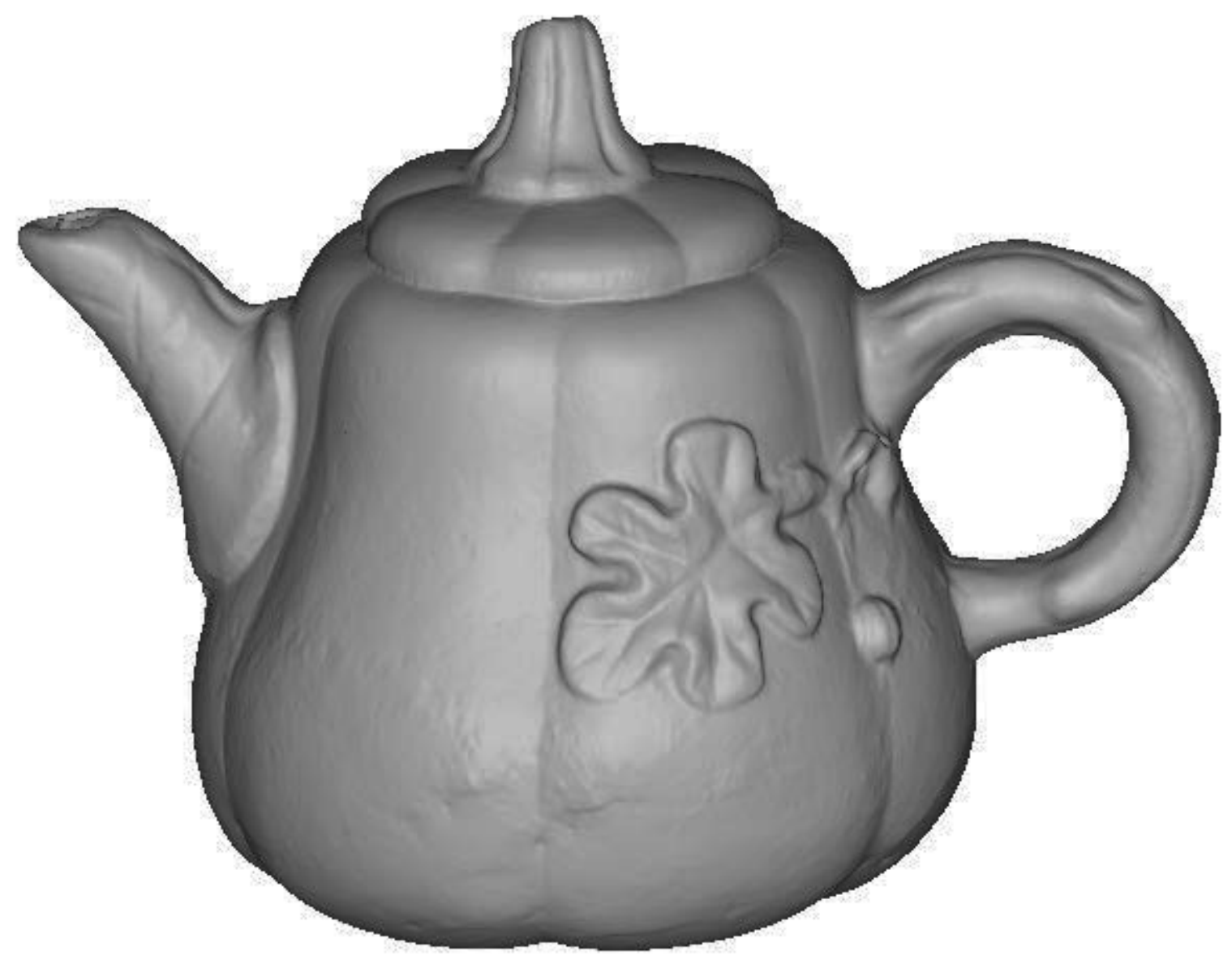}
        & \includegraphics[width=0.15\linewidth]{teapot/teapot_geometry.pdf}
		\\
		  \includegraphics[width=0.15\linewidth]{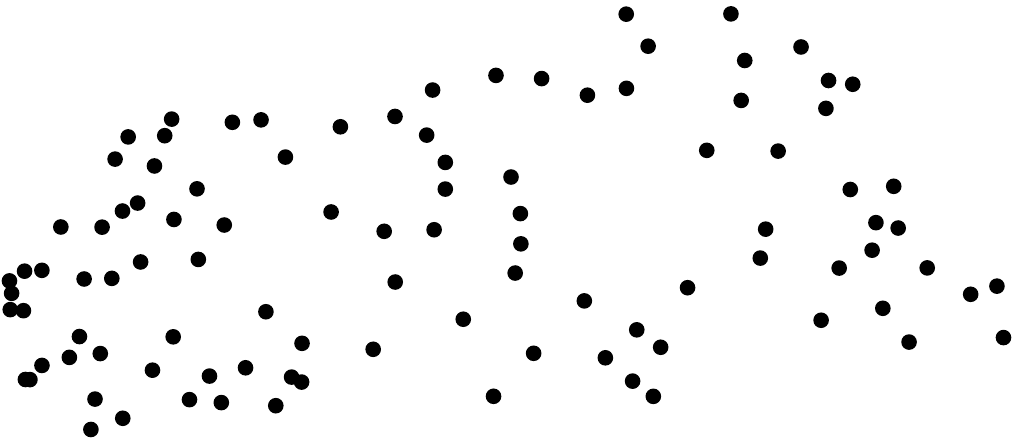}
		& \includegraphics[width=0.15\linewidth]{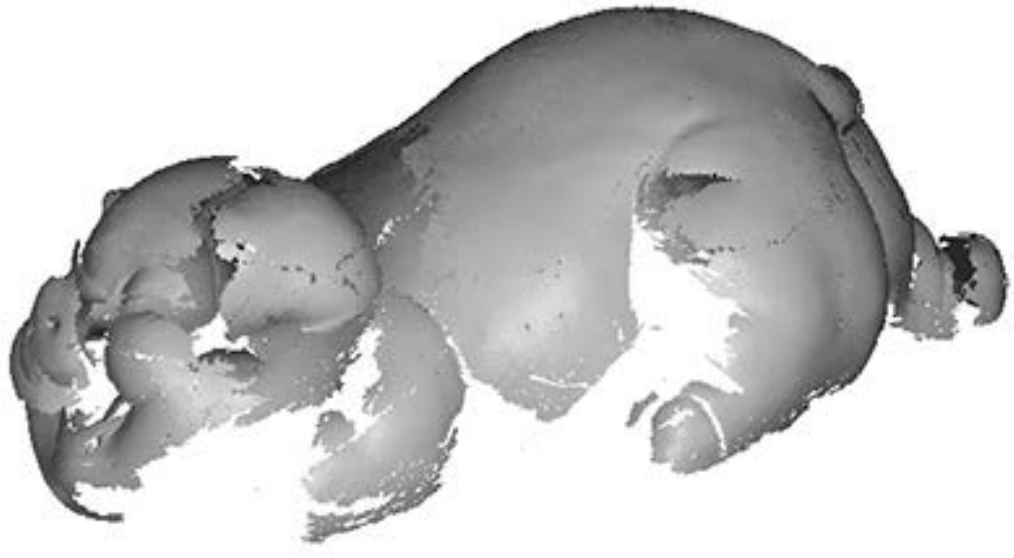}
		& \includegraphics[width=0.15\linewidth]{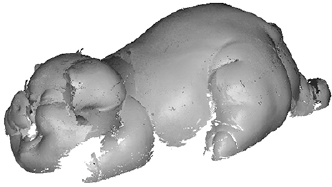}
		& \includegraphics[width=0.15\linewidth]{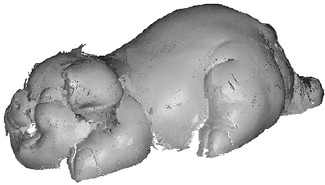}
		& \includegraphics[width=0.15\linewidth]{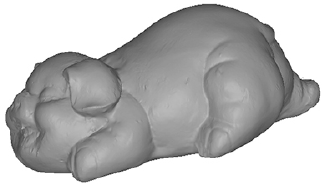}
        & \includegraphics[width=0.15\linewidth]{pig/pig_geometry1.pdf}
        \\
        (a) & (b) & (c) & (d) & (e) & (f)
	\end{tabular}
	\caption{ Intermediate results of depth propagation. (a) initial 3D points reconstructed by structure-from-motion;
		(b) 3D points obtained by the first iteration of depth propagation;
        (c) and (d) 3D points obtained after the middle and last iteration of depth propagation;
        (e) mesh after Poisson surface reconstruction from the points in (d);
        (f) final shapes after jointly optimizing shape and normal.}
        \label{fig:step_by_step}
\end{figure*}

Quantitative shape errors for all examples in \figref{all_results} are summarized in \tabref{errors}. The studio setup achieves the best accuracy of around $0.5$ millimeters mean error, but requires a large and bulky capture setup.
The desktop setup balances data capture flexibility and shape accuracy. 

\begin{figure*}\centering
	\begin{tabular}{@{\hspace{1mm}}c@{\hspace{1mm}}c@{\hspace{1mm}}c}
     \vspace{-0.2cm}   
		  \includegraphics[height= 0.24 \linewidth]{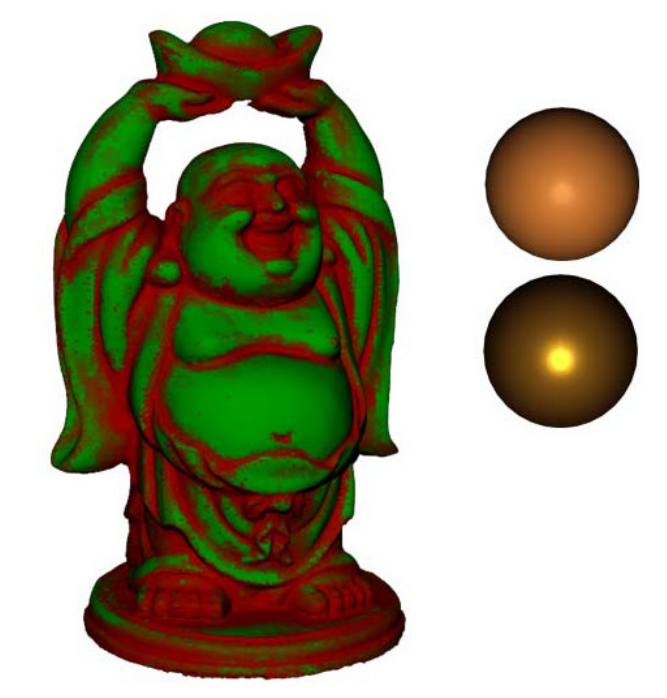}
		& \includegraphics[height= 0.20 \linewidth]{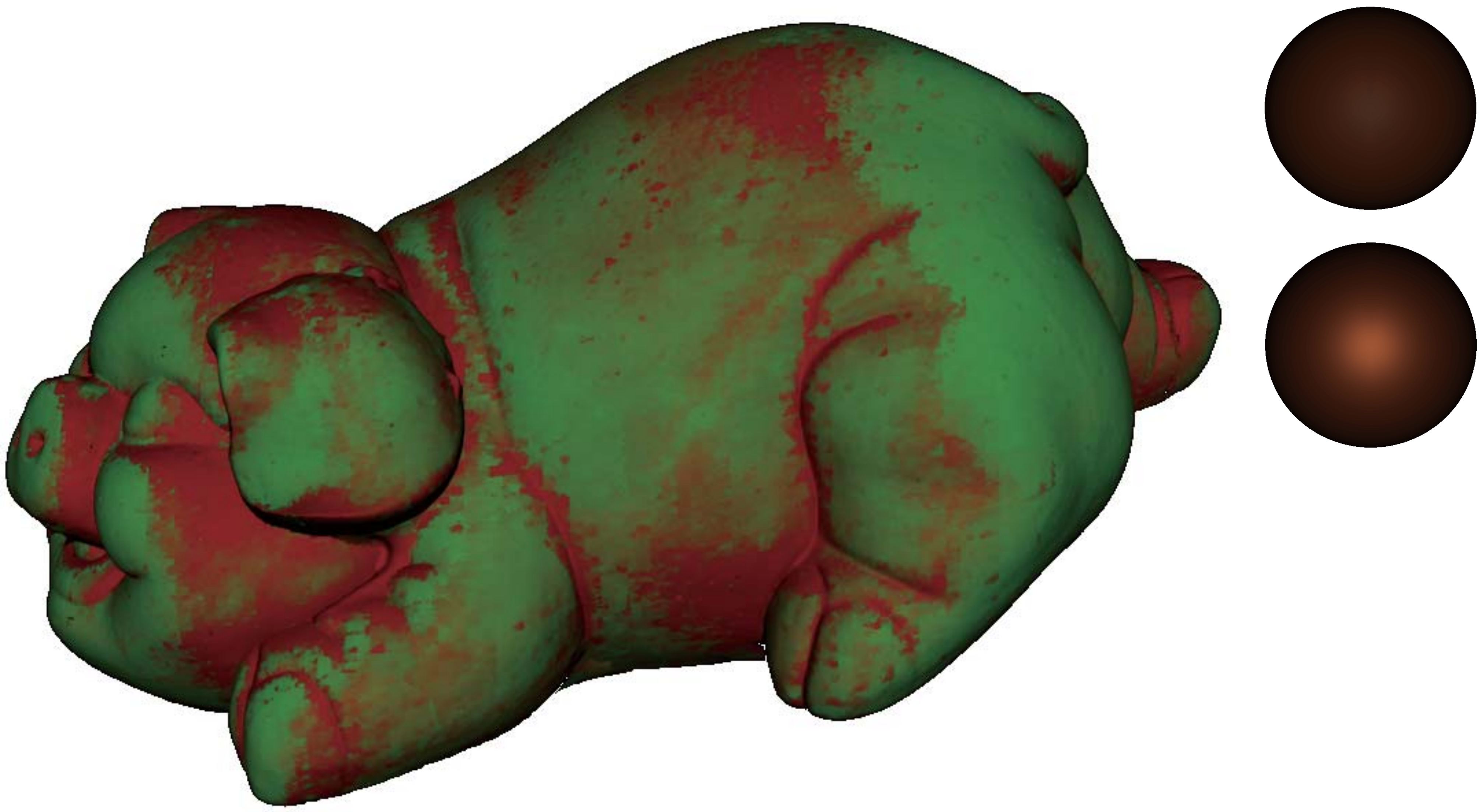}
		 & \includegraphics[height= 0.21 \linewidth]{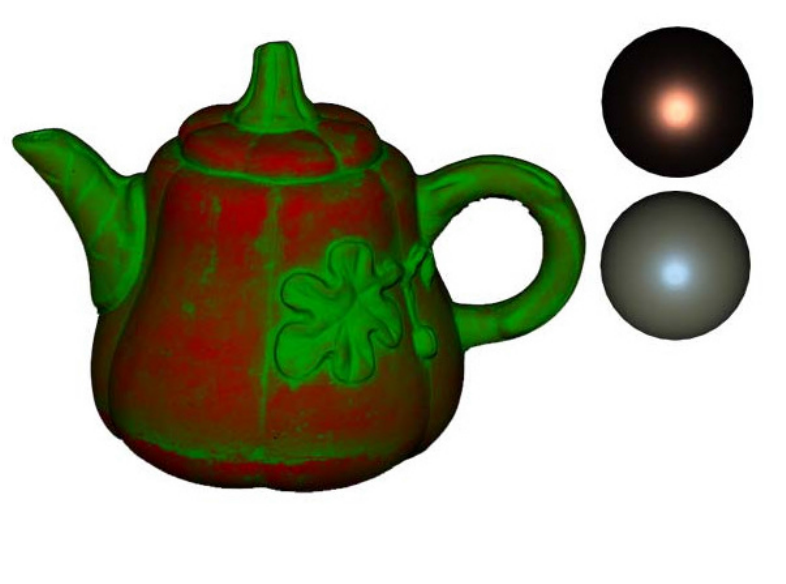}
	\end{tabular}
	\caption{The normalized BRDF mixture weights are visualized in different color channels. The corresponding basis BRDFs are used to render a sphere on the side.} \label{fig:results_brdf}
\end{figure*}

\subsubsection{Intermediate Results}
\figref{step_by_step} shows some intermediate results  at different stages of depth propagation for the examples `\emph{Buddha-S}', `\emph{Teapot2-S}', and `\emph{Pig-D}'.
Results shown in \figref{step_by_step} (a) are initial 3D points obtained from structure-from-motion.
These points are the initial seeds for depth propagation. We choose a tight threshold to get rid of unreliable points.
Results shown in (b) are the 3D points after the 1st iteration of depth propagation, where a large portion of the surface has been reconstructed.
(c) and (d) are the 3D points after half and all of the propagation iterations respectively.
The black points in (b) and (c) are those with a surface normal facing away from the camera.
Typically, it takes 5-6 iterations of depth propagation to cover the complete surface.
It is interesting to see that our method can propagate dense depth through a very small number of seed points in (a).
Quick shape change makes the propagation slower, as the iso-depth contours tend to break at shape discontinuities.
For example, half of the propagation iterations are spent to cover the mouth of `\emph{Buddha-S}' and the flower decoration on `\emph{Teapot2-S}'.
Results shown in (e) are the initial surface meshes after Poisson surface reconstruction.
Some shape errors are visible, e.g., on the face of the `\emph{Buddha-S}' example, which are due to the drifting errors of depth propagation.
Results shown in (f) are the final optimized shapes by the method in \cite{Nehab2005}.
The surfaces become clearly smoother after optimization.

\figref{results_brdf} visualizes the BRDF mixture weights and the basis BRDFs for the `\emph{Buddha-S}', `\emph{Pig-D}', and `\emph{Teapot2-S}' examples.
The red and green channels are the normalized mixture weight of the first and second basis BRDFs.
Each basis BRDF is applied to render a sphere under front lighting for reference.
Most of our examples are consisted of a shiny and a less shiny basis BRDFs.
This can be seen clearly from the `\emph{Buddha-S}' example.

\subsection{Performance Analysis of Our Method}\label{sec:results_quant}

\subsubsection{Number of Image at Each Viewpoint}
We evaluate the accuracy of captured shape with different numbers of input images from each viewpoint. Similar analysis on BRDF capturing can be found in the conference version \cite{Zhou2013}.

\begin{figure}\centering
	\includegraphics[width=0.95\linewidth]{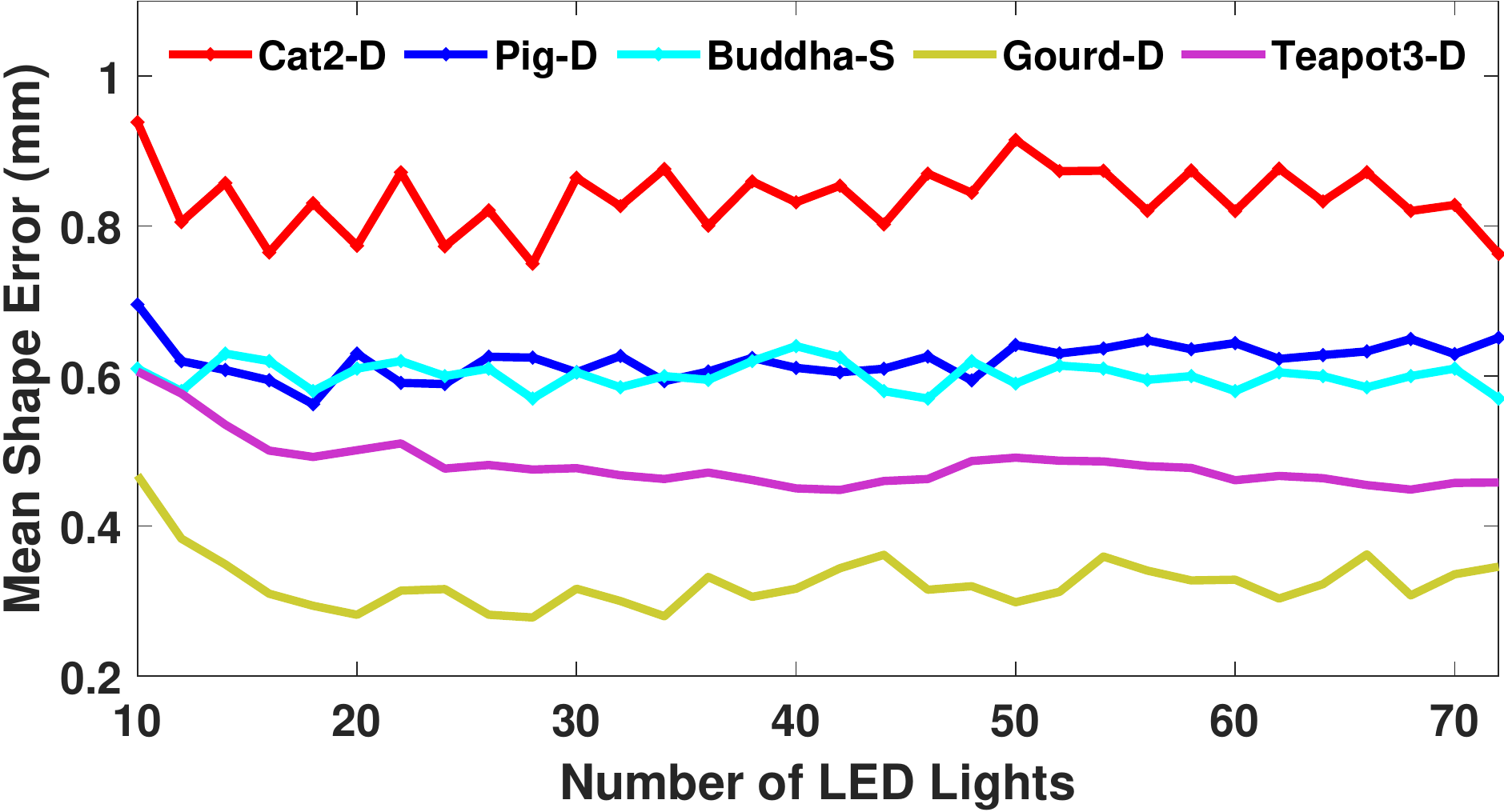}
	\caption{The mean shape error does not change significantly with different number of input images per viewpoint.}
\label{fig:error_vs_light}
\end{figure}

\figref{error_vs_light} shows the mean shape error (in millimeters) with different numbers of input images in each viewpoint, starting with 10 images per view.
We always chose equal number of uniformly distributed lights on both the outer and inner circles, i.e., starting with 5 LED lights for each circle.
This is because our Fourier series fitting requires at least 5 LEDs from each viewpoint.
In most of the examples, the mean shape error does not change significantly for different numbers of input images.
The errors of `\emph{Teapot3-D}' and `\emph{Gourd-D}' drop slightly when the number of input image per viewpoint increases to 20.
`\emph{Cat2-D}' has the largest error, most likely because it spans the largest FoV to the camera.

\subsubsection{Perspective vs Orthographic Camera}
\begin{figure}\centering	
	\begin{tabular}{@{\hspace{1mm}}c@{\hspace{1mm}}c@{\hspace{1mm}}c}
		\hspace{-2mm}	
		\includegraphics[height=0.40 \linewidth] {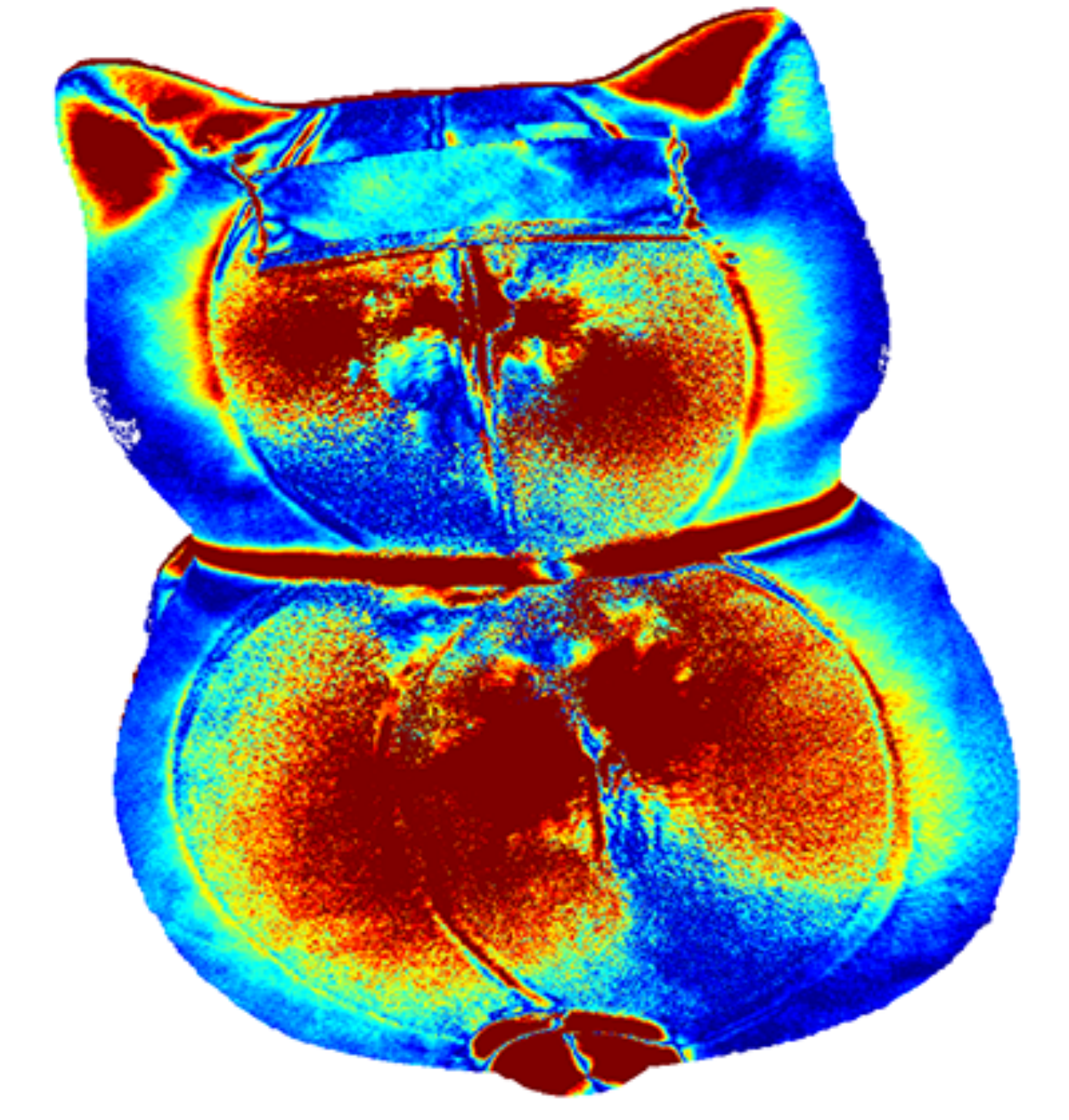}
		& \includegraphics[height=0.40 \linewidth] {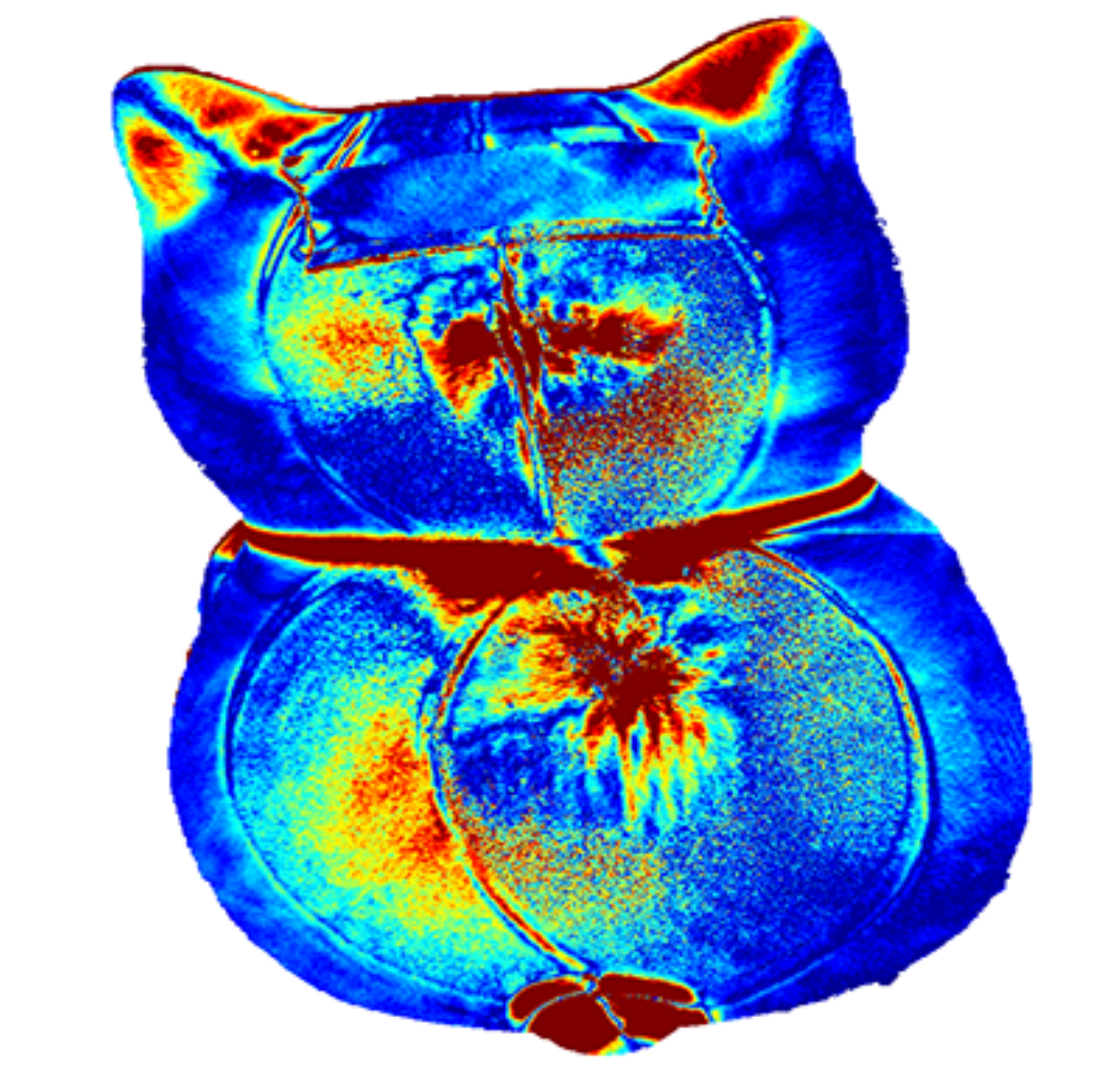} 		
		& \includegraphics[height=0.40 \linewidth]{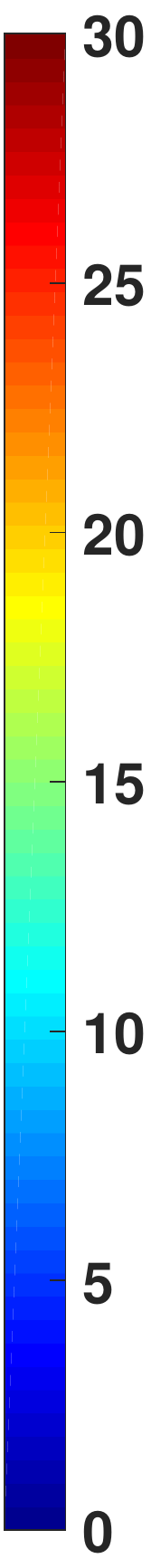}	\\
		
		\includegraphics[height=0.35 \linewidth] {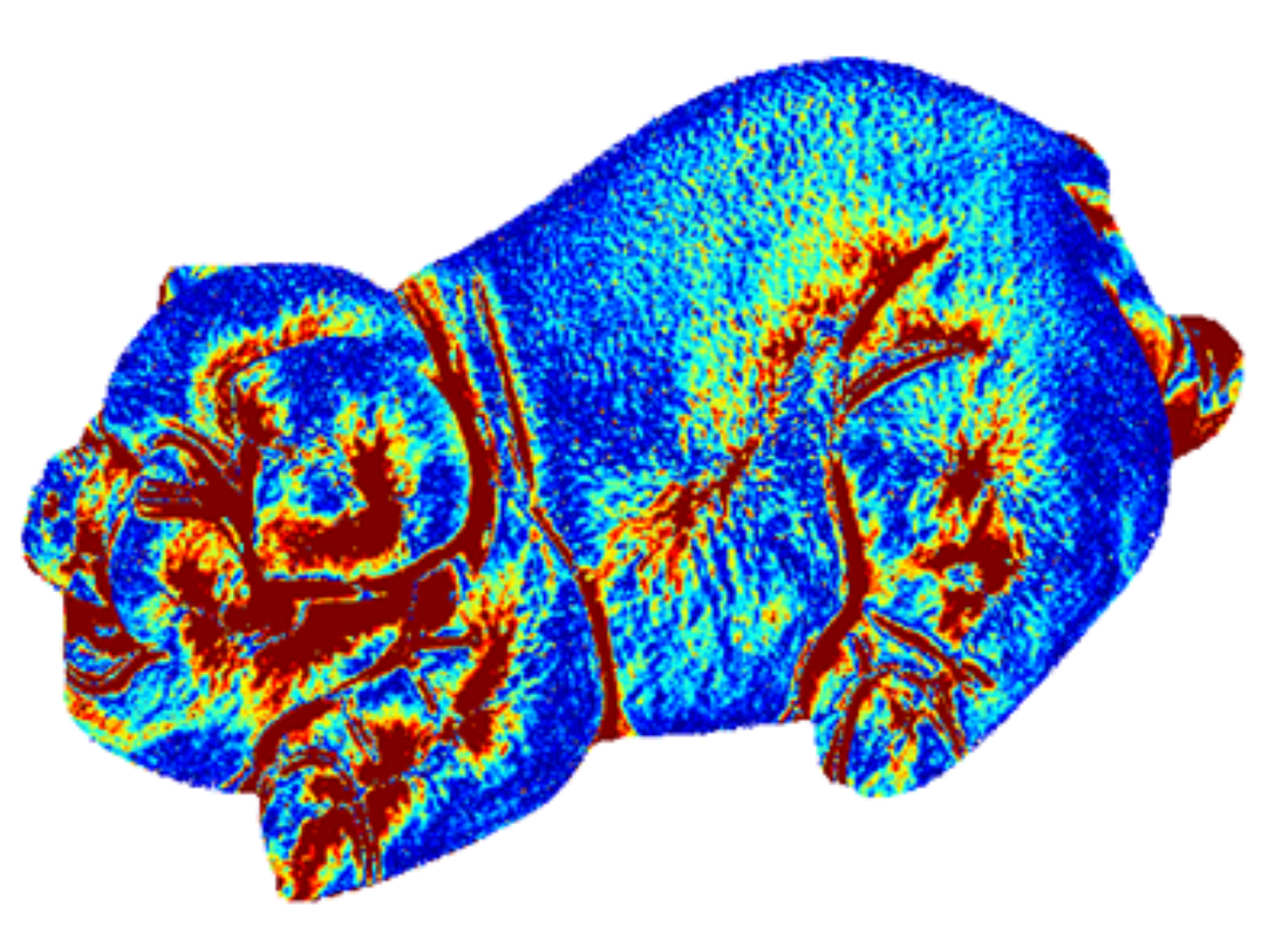}
		& \multicolumn{2}{c}{\includegraphics[height=0.35 \linewidth] {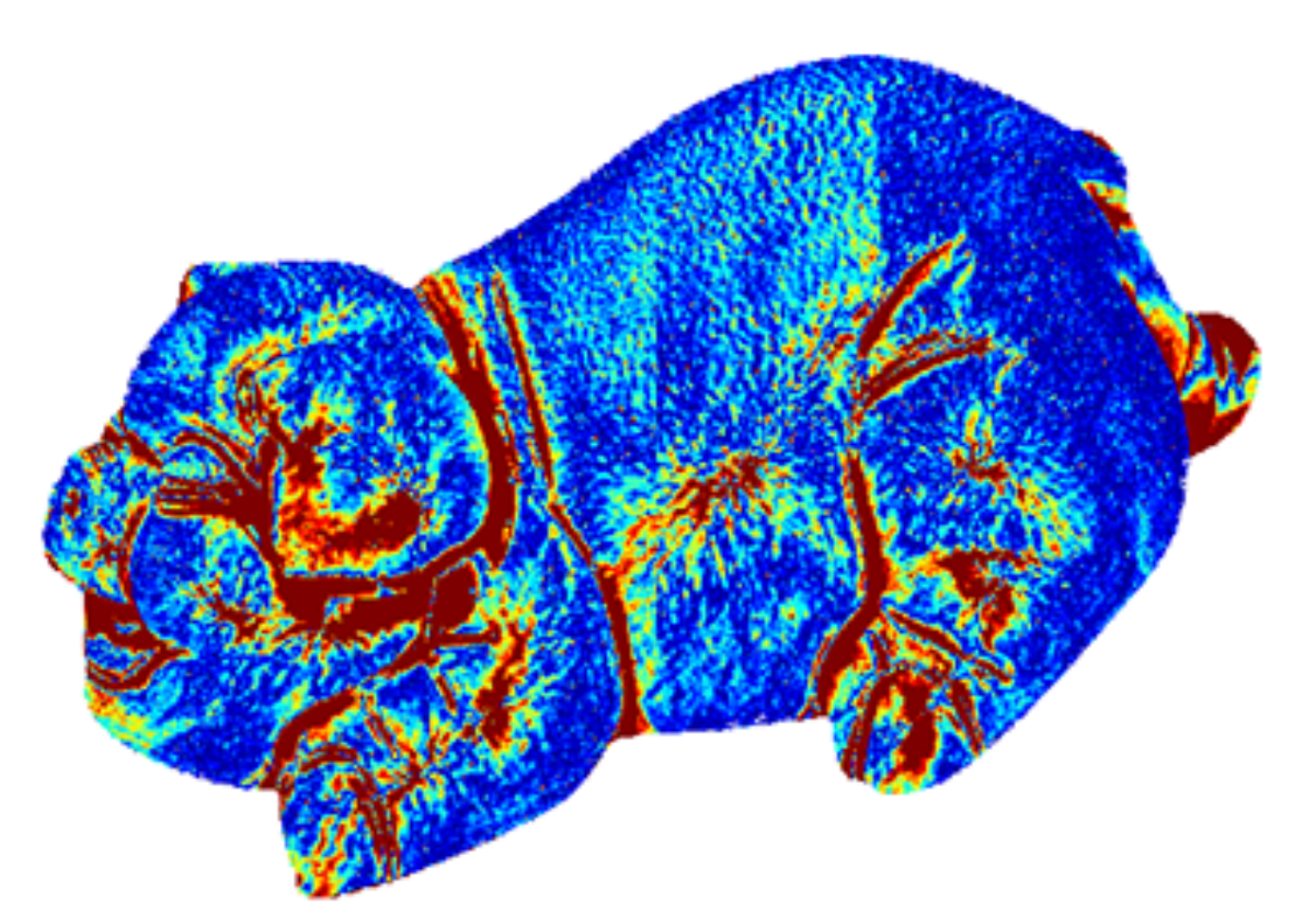}}
	\end{tabular}
	\caption{The color coded azimuth angle errors with and without the sub-dividing method presented in \secref{perspective}.
		The 1st and 2nd rows are the results of the `\emph{Pig-D}' and `\emph{Cat2-D}' examples respectively.
		On the left is the result obtained without sub-dividing (`orthogonal camera'), and on the right is the one with sub-dividing (`perspective camera').
		It is clear that the sub-dividing approach can significantly reduce errors in azimuth angles.
	}\label{fig:azimuth_perspective}

\end{figure}

\begin{figure}\centering
\includegraphics[height = 0.45 \linewidth]{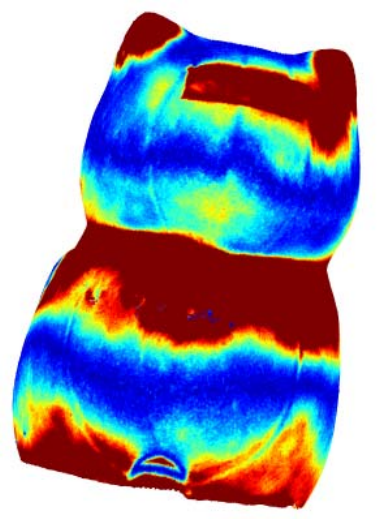}
\includegraphics[height= 0.4 \linewidth]{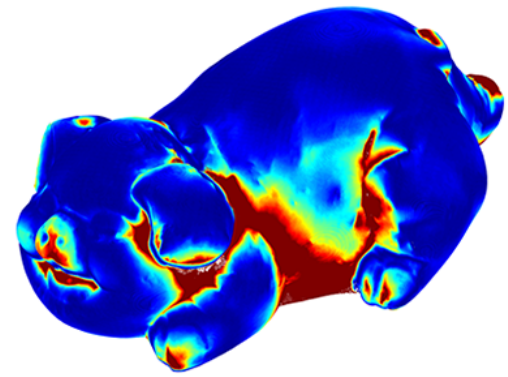}
\includegraphics[height= 0.45 \linewidth]{Figures/shapeerror_visualization.pdf}
\caption{The color coded shape reconstruction errors without the sub-dividing method (`orthographic camera') in \secref{perspective} for the `\emph{Pig-D}' and `\emph{Cat2-D}' examples.
Please refer to \figref{all_results} for their shape errors with sub-dividing (`perspective camera').
}	\label{fig:results_fov22}

\end{figure}

To evaluate our sub-dividing method for perspective cameras as presented in \secref{perspective},
we test the `\emph{Pig-D}' and `\emph{Cat2-D}' examples with and without sub-dividing the image plane.
\figref{azimuth_perspective} shows the error in estimated azimuth angles.
As we can see, the `perspective camera' model produces much more accurate results.
It reduces the mean azimuth angle error from  $12.1$ degrees to $9.9$ degrees for the `\emph{Pig-D}' example,
and from $19.5$ degrees to $12.6$ degrees for the `\emph{Cat2-D}' example respectively.
\figref{results_fov22} shows the shape reconstruction errors.
Similarly, the `perspective camera' model reduces the mean shape  reconstruction error from $0.75$ millimeters to $0.57$ millimeters for the `\emph{Pig-D}' example,
and from $1.66$ millimeters to $0.76$ millimeters for the `\emph{Cat2-D}' example respectively.
This comparison demonstrates the effectiveness of the simple sub-dividing approach,
which helps to reduce the setup size to fit a desktop.
Note that the shape error in \figref{results_fov22} is much smoother than the azimuth angle error in \figref{azimuth_perspective},
because the shape is reconstructed by fusing azimuth angle information from many viewpoints and the later Poisson surface reconstruction  and shape optimization \cite{Nehab2005} all help to smooth out the noise.

\begin{figure}\centering
\includegraphics[width = 0.48 \linewidth]{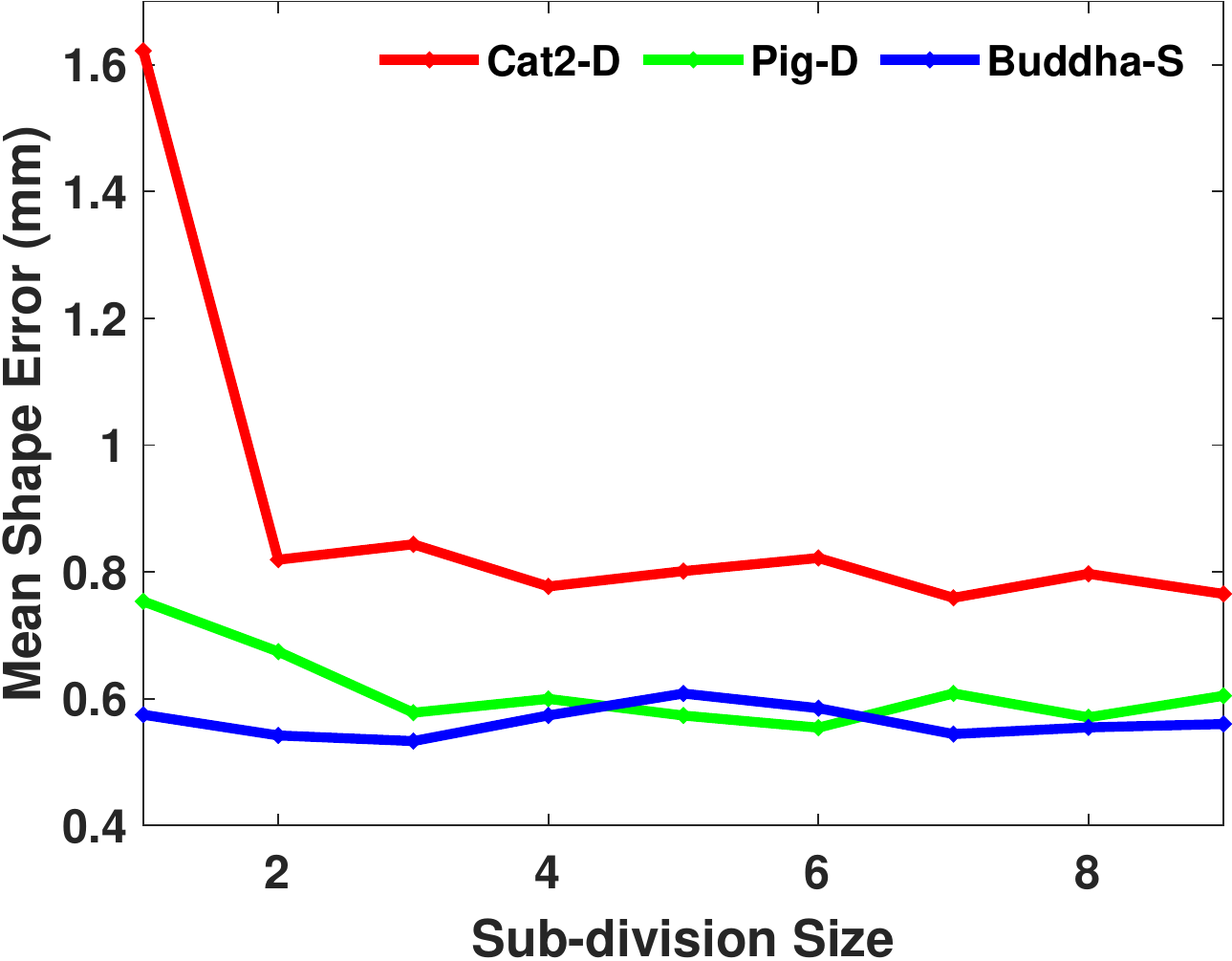}
\includegraphics[width = 0.48 \linewidth]{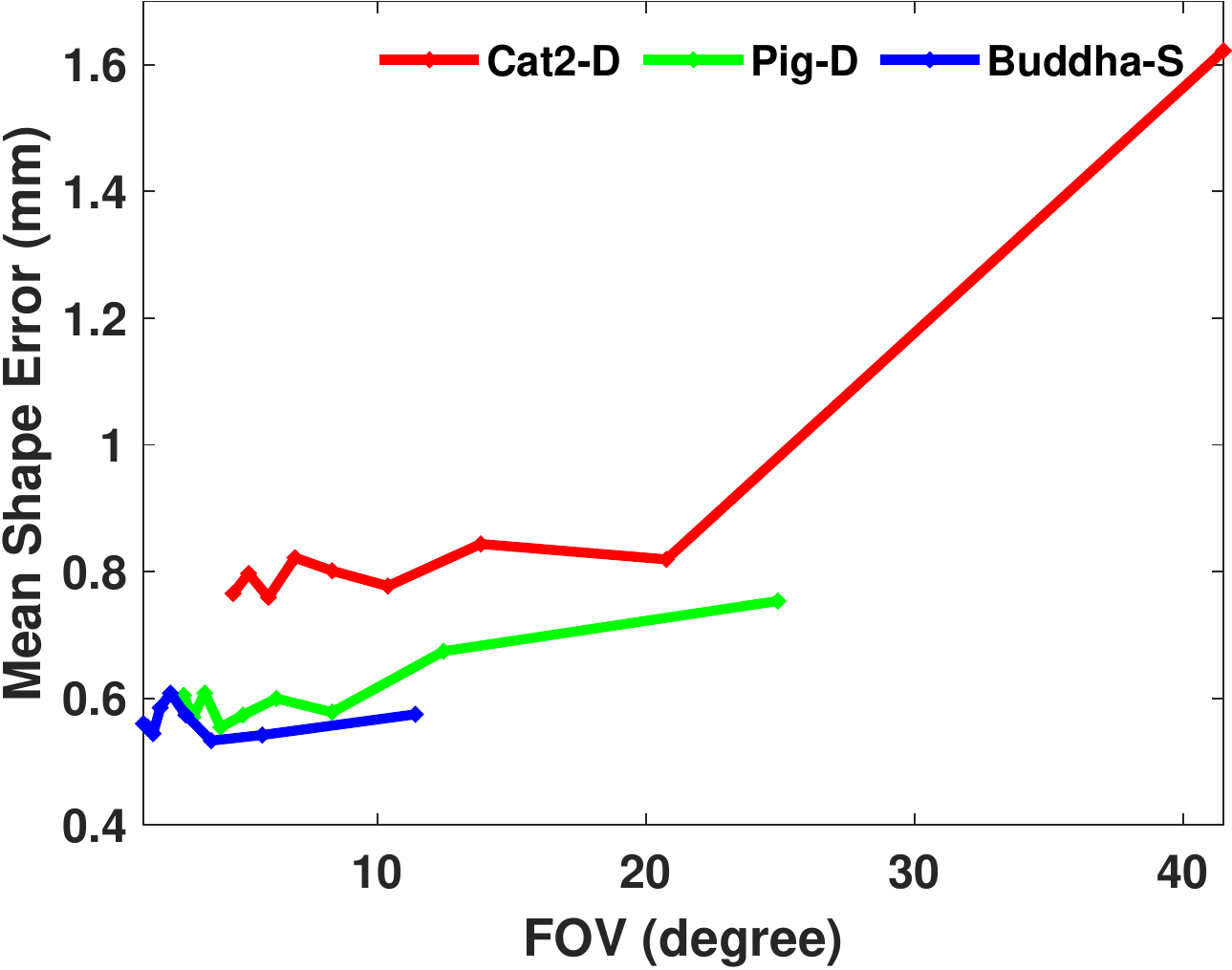}
\caption{Left: mean shape error with different numbers of sub-division to the image plane to model perspective cameras.
Right: mean shape error with different horizontal field-of-view (FoV) per sub-divided window.
}	\label{fig:error_vs_division}
\end{figure}

We further analyze the mean shape error with different numbers of sub-divisions to examine our algorithm for perspective cameras.
The left of \figref{error_vs_division} shows the mean shape error of the `\emph{Pig-D}', `\emph{Cat2-D}', and `\emph{Buddha-S}' examples with different numbers of sub-divisions.
The shape error of the `\emph{Cat2-D}' is significantly reduced by this subdivision scheme, while the error of `\emph{Pig-D}' is only mildly decreased.
To understand this, we further plot the mean shape error with different horizontal FoV of each sub-divided window in the right of  \figref{error_vs_division}.
The `\emph{Buddha-S}' example always spans a FoV less than 10 degrees, and thus sub-division is unnecessary.
The `\emph{Pig-D}' example spans a 25-degree FoV. Thus, a $2\times 2$ sub-division will be sufficient.
The `\emph{Cat2-D}' example has a much larger FoV and a $3 \times 3$ sub-division is desired to achieve best results.
Empirically, we find the orthogonal camera assumption can be safely made, when the FoV is less than 10 degrees.

%


\subsubsection{Runtime Efficiency}
Our implementation is not optimized for speed. We finish all experiments on a computer with 24 GB RAM and an 8-core 3.0 GHz CPU. At each viewpoint, our Matlab code computes azimuth angles in 1 minute, and traces iso-depth contours in 1.5 minutes. Depth propagation takes 16 minutes (for 40 viewpoints), and the final shape optimization takes 1 minute. It takes about 15 minutes to compute the basis BRDFs from 5,000 samples with ACLS. Our output mesh typically has about 150,000 points with average spatial distance 0.095 millimeters. It takes another 45 minutes to compute their BRDF mixing weights. Much of the involved process including azimuth angle computation, iso-depth contour tracing, and BRDF mixing weight computation can be easily parallelized.

\subsection{Evaluation on `DiLiGenT' and `DiLiGenT-MV'}
We first compare our method with the representative single-view photometric stereo methods, IA14\cite{ikehata2014photometric}, ST14\cite{shi2014bi}, HS15\cite{han2015photometric}, SH17\cite{shen2017efficient}, and ZK19\cite{zheng2019numerical} on `DiLiGenT' for normal estimation accuracy. Then we evaluate our method and another representative MVPS method PJ16\cite{park2016robust} on `DiLiGenT-MV' for both normal and shape accuracy. For all validated methods, we use the parameters provided in the original codes or suggested by the original papers. We use all 96 images from all 20 viewpoints to evaluate different methods, with an exception of \cite{park2016robust}, whose executable fails when being fed with over 10 images for each viewpoint. Since the FoV of the camera used for both datasets is less than 10 degrees, we employ the `orthogonal' algorithm for the following evaluation according to \secref{results_quant}.

\subsubsection{Quantitative Normal Accuracy on `DiLiGenT'}
We choose BEAR, BUDDHA, COW, POT2, and READING from the `DiLiGenT' dataset to validate our method. 
We projected the final results of multi-view based methods to get normals at different viewpoints to compare them with those single view based methods.
Results of single-view photometric methods are from the benchmark result on `DiLiGenT' online.

\begin{figure*}
	\includegraphics[height = 0.18 \linewidth]{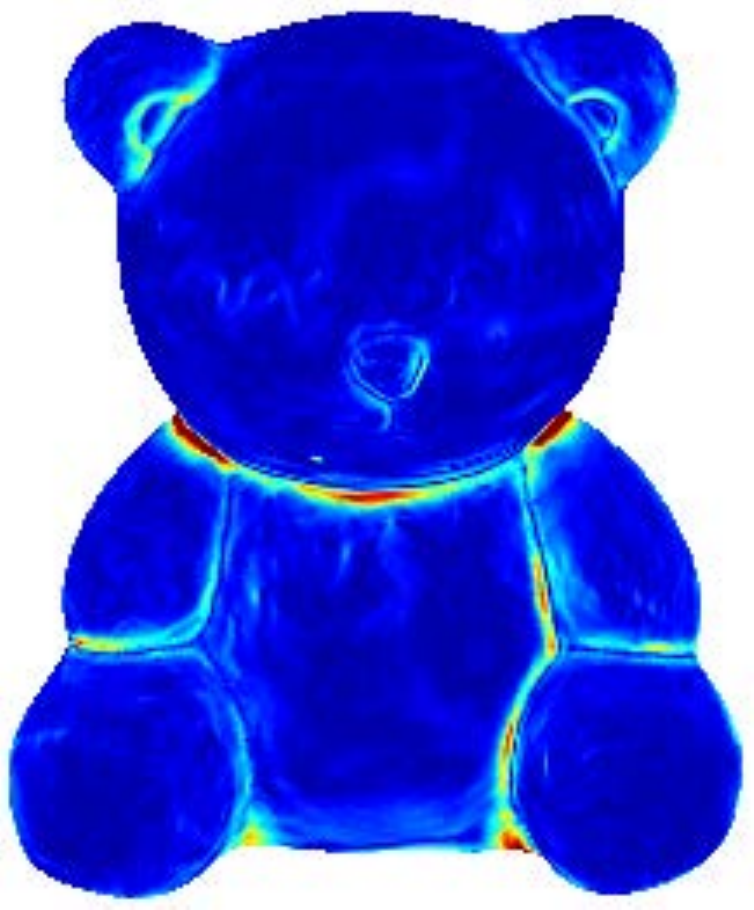}
\includegraphics[height = 0.18 \linewidth]{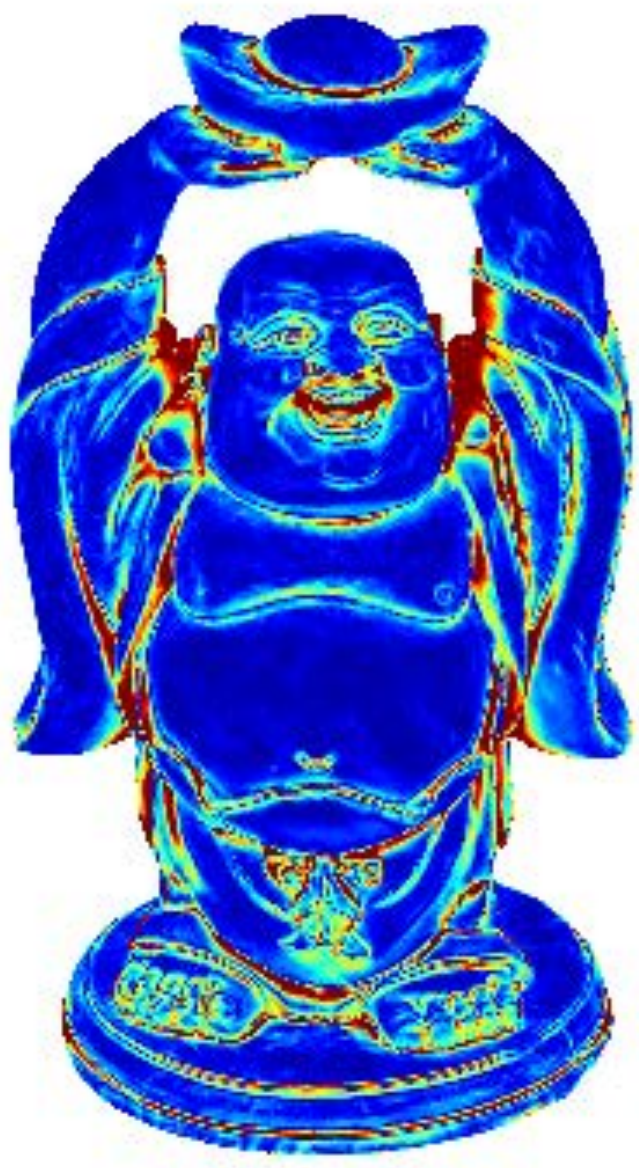}
\includegraphics[height = 0.18 \linewidth]{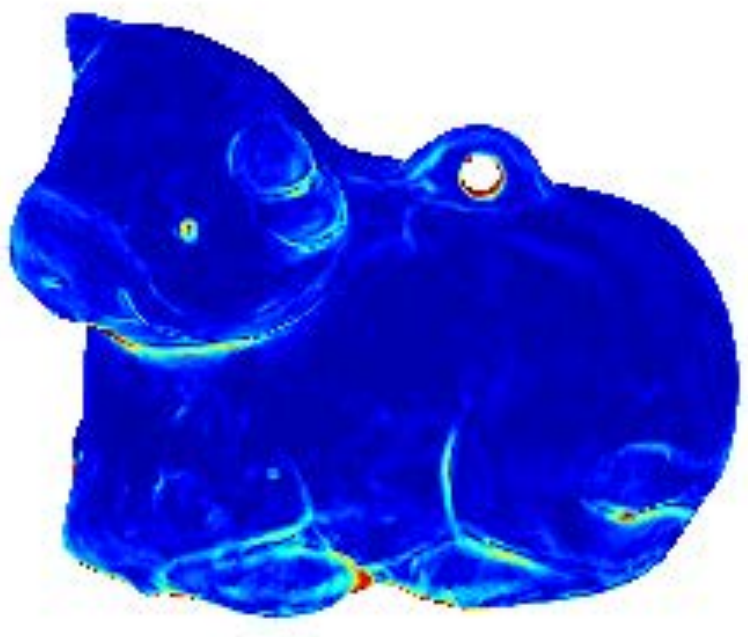}
\includegraphics[height = 0.18 \linewidth]{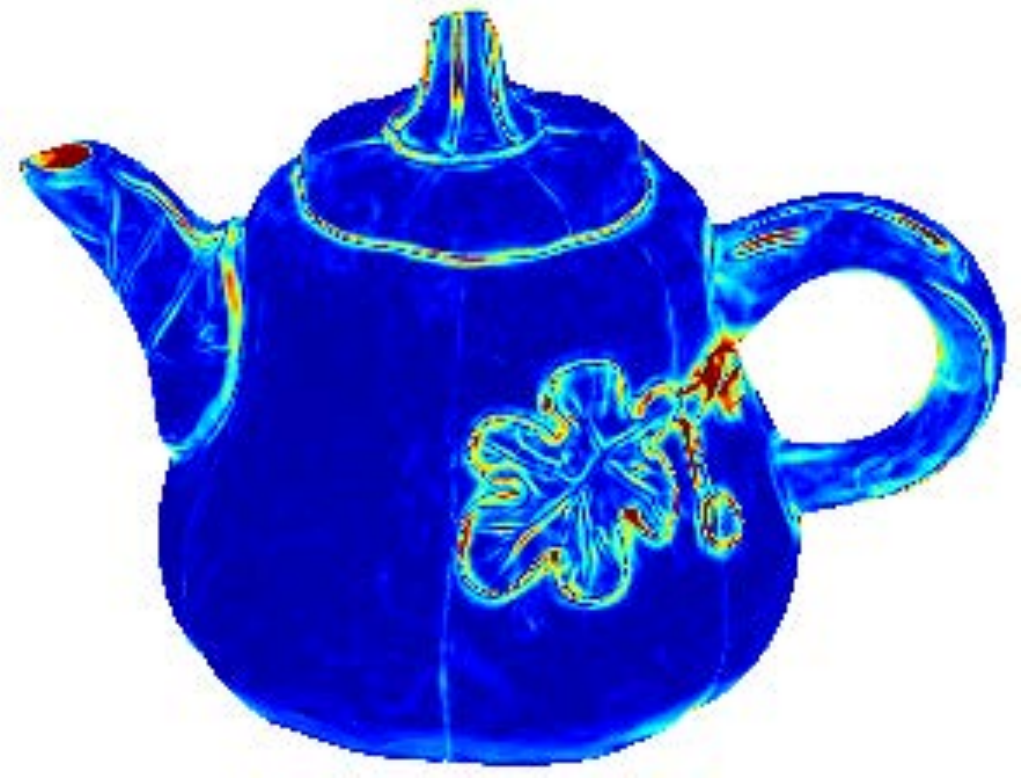}
\includegraphics[height = 0.18 \linewidth]{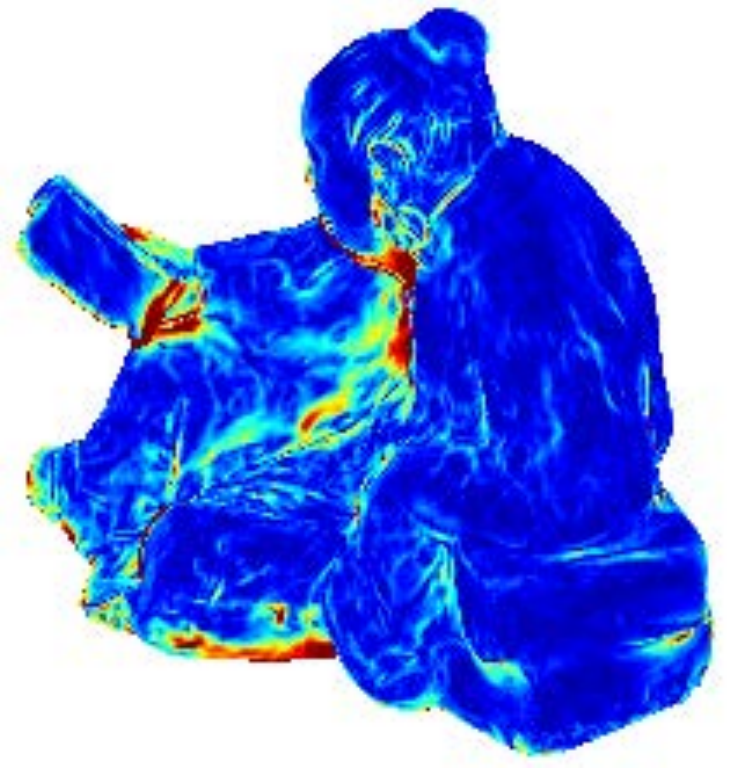}
\includegraphics[width = 0.03 \linewidth]{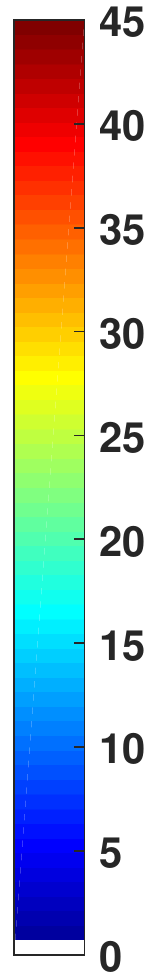}
	\caption{Normal error maps of  our method on `DiLiGenT-MV'. The error range is [0, 45] in degree.}
	\label{fig:normal_error_view1}
\end{figure*}

Multi-view methods (i.e., our method and PJ16\cite{park2016robust}) generate better results than single-view methods on most objects except for BUDDHA, 
as they have access to more views to overcome the difficulties in photometric stereo, e.g., correcting the low-frequency shape distortions. The BUDDHA example is an exception because the strong inter-reflections at wrinkles, which are not modeled in these two evaluated multi-view methods. This explains why some more sophisticated single-view algorithms produce better results. Our method outperforms other methods on the challenging example COW in \figref{normal_error_view1}, because the non-Lambertian material here satisfies the isotropy assumption well as discussed in \cite{shi2019benchmark}.

\begin{table*}
	\centering 
\caption{Mean normal angular errors (in degree) on `DiLiGenT'.} 
\vspace{-0.2cm}
	\begin{tabular}{ccccccccccc}
		\toprule[1pt]
		& LS\cite{woodham1979photometric} & IA14\cite{ikehata2014photometric} & ST14\cite{shi2014bi} & HS15\cite{han2015photometric} & SH17\cite{shen2017efficient} 
		& ZK19\cite{zheng2019numerical} & PJ16\cite{park2016robust} & Ours	
		\\
	\midrule[1pt]
		
		BEAR & 8.39 & 7.11 &  6.12 & 5.12 & 5.31 
		& 4.65 & 12.63 & \textbf{4.45}
		\\
		
		BUDDHA & 14.92 & 10.47 & 10.60 & 12.29 & 9.30 
		& \textbf{9.14} & 14.58 & 11.64
		\\
		
		COW & 13.05  & 13.95 & 6.51 & 17.20 & 16.79 
		& 15.85 & 13.24 & \textbf{4.13}
		\\
		
		POT2 & 14.65 & 8.77 & 8.78 & 9.80 & 8.43 
		& 8.09 & 15.31 & \textbf{6.79}
		\\
		
		READING & 19.80 & 14.19 & 13.63 & 14.56 & 13.00 
		& 12.77 & 12.23 & \textbf{8.74}
		\\
		\hline
	\end{tabular}
	\label{tab:benchmark_errors}
\end{table*}

\subsubsection{Quantitative Shape and Normal Accuracy on `DiLiGenT-MV'}
\begin{figure}
	\begin{tabular}{@{\hspace{1mm}}c@{\hspace{1mm}}c@{\hspace{1mm}}c@{\hspace{1mm}}c@{\hspace{1mm}}c}
		\includegraphics[width = 0.3 \linewidth]{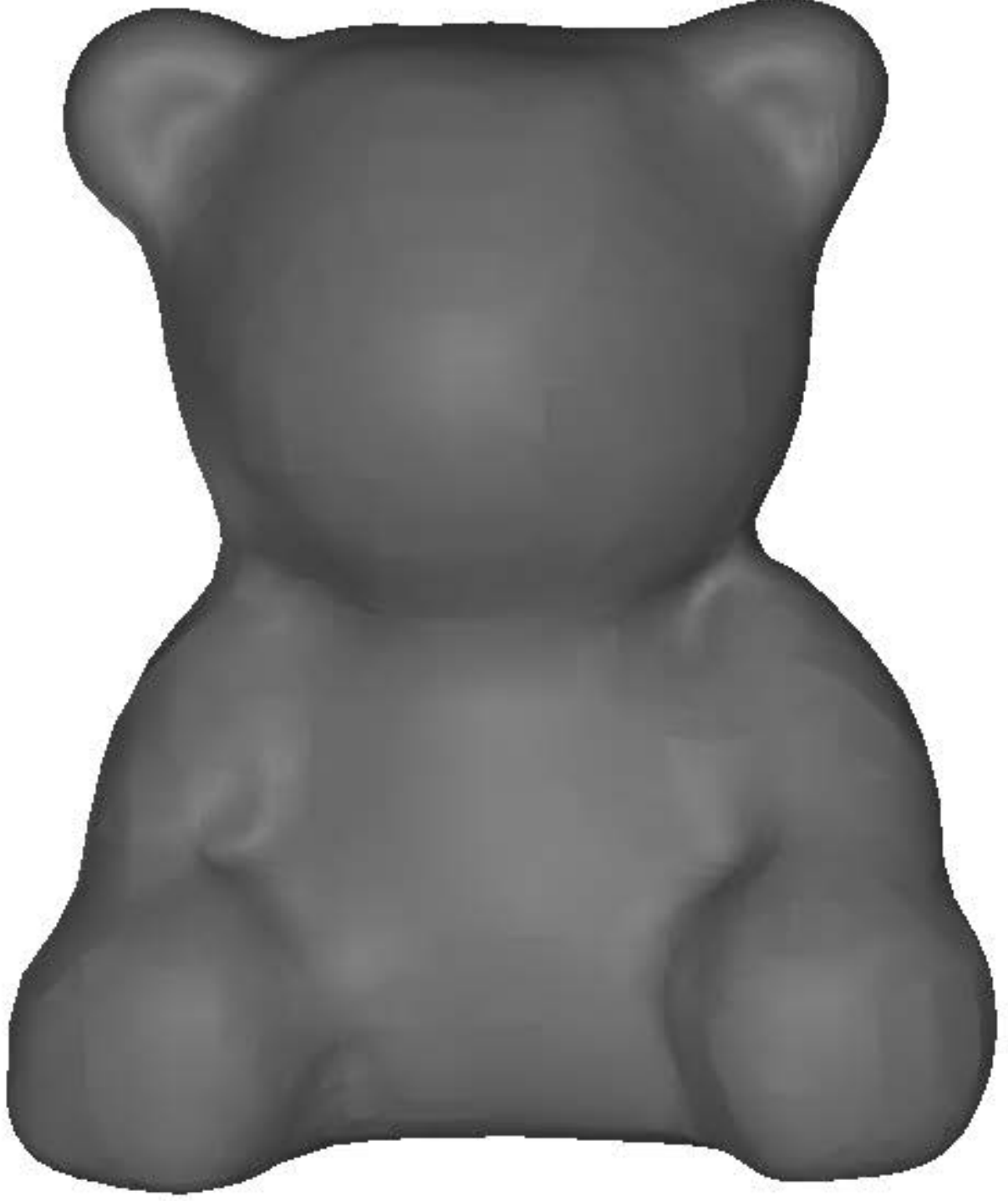}
		&\includegraphics[width= 0.3 \linewidth]{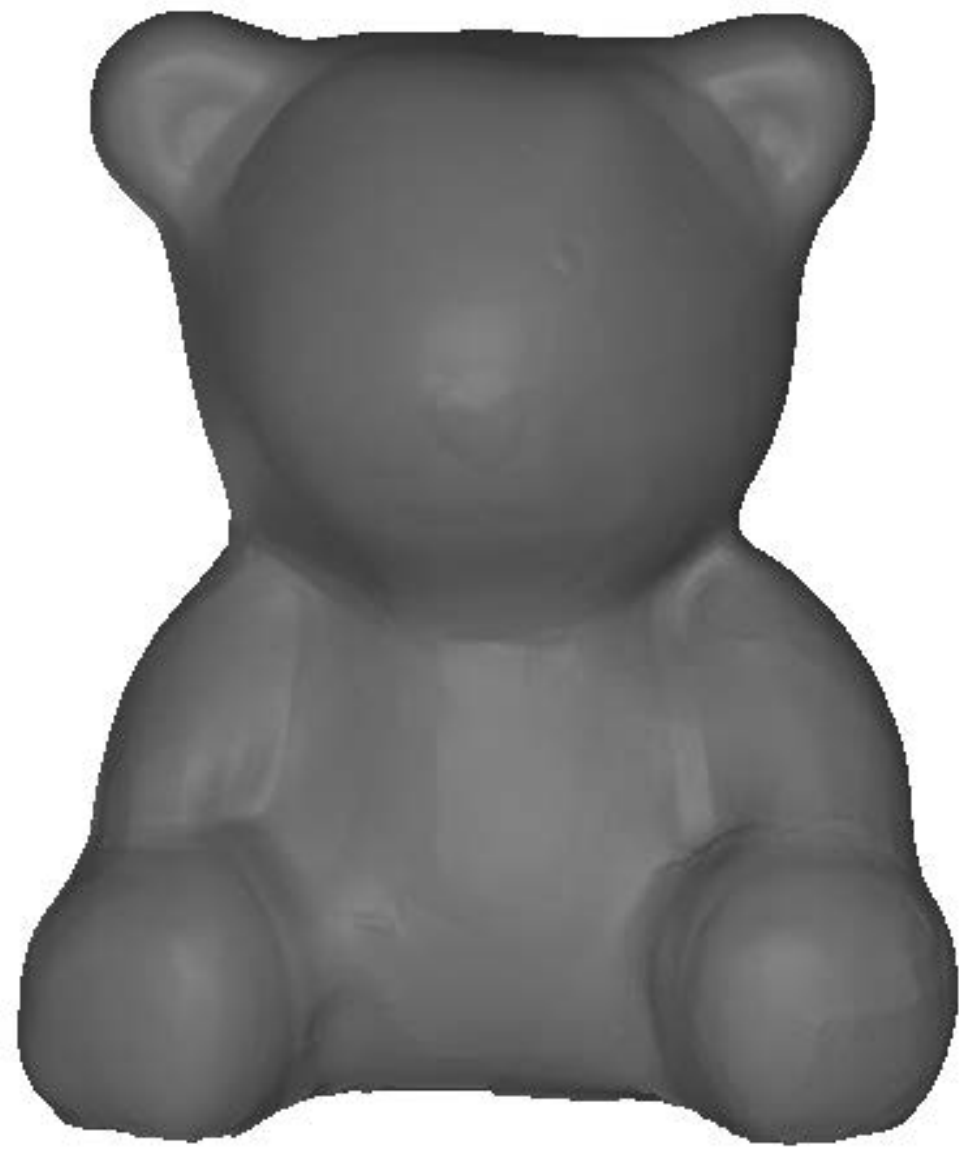}
		&\includegraphics[width= 0.3 \linewidth]{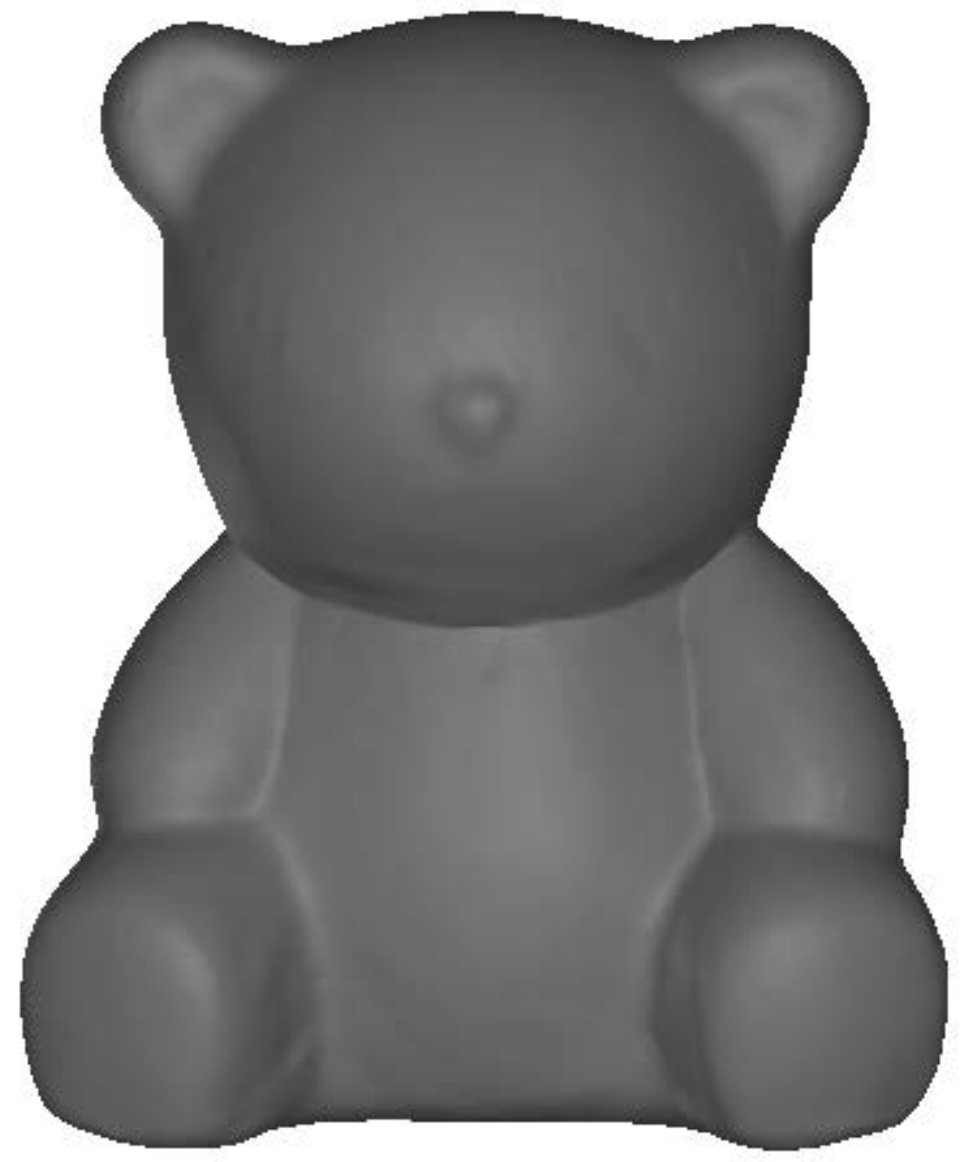}
		\\
		
		\includegraphics[width = 0.3 \linewidth]{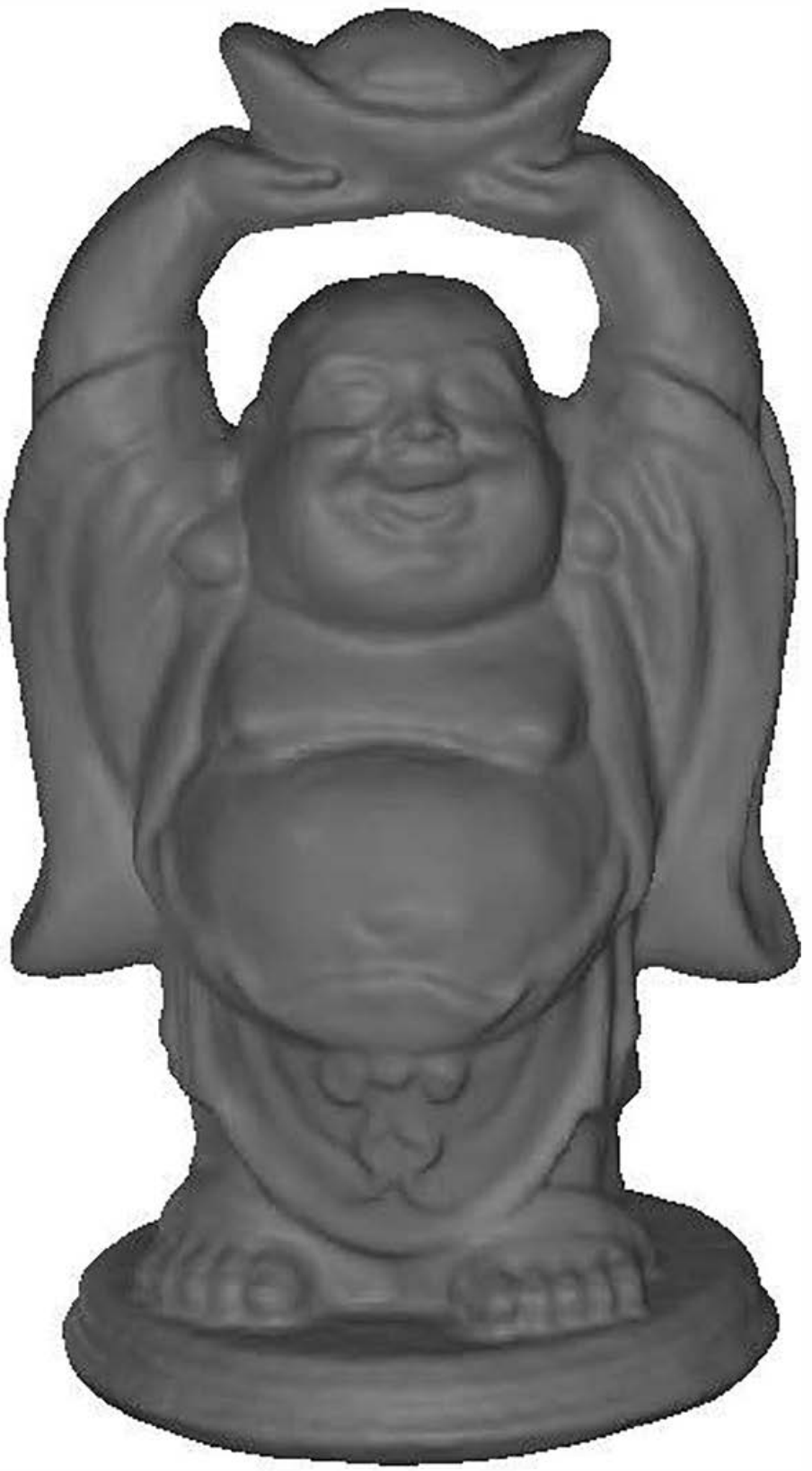}
		&\includegraphics[width= 0.31 \linewidth]{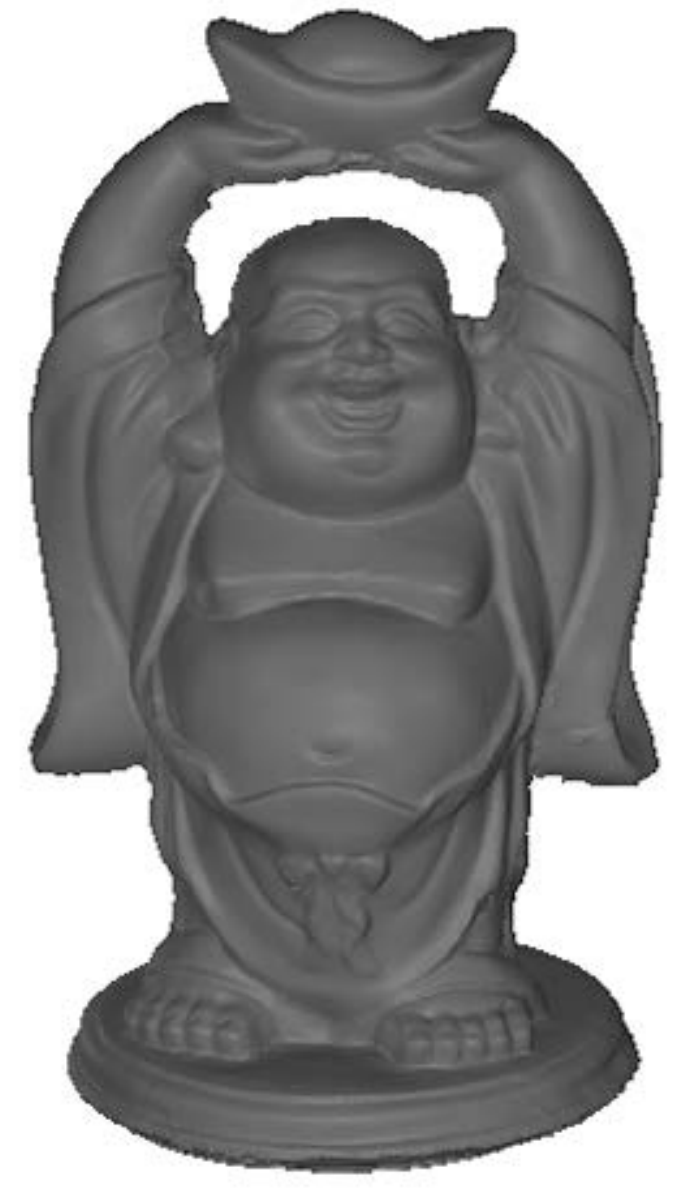}
		&\includegraphics[width= 0.31 \linewidth]{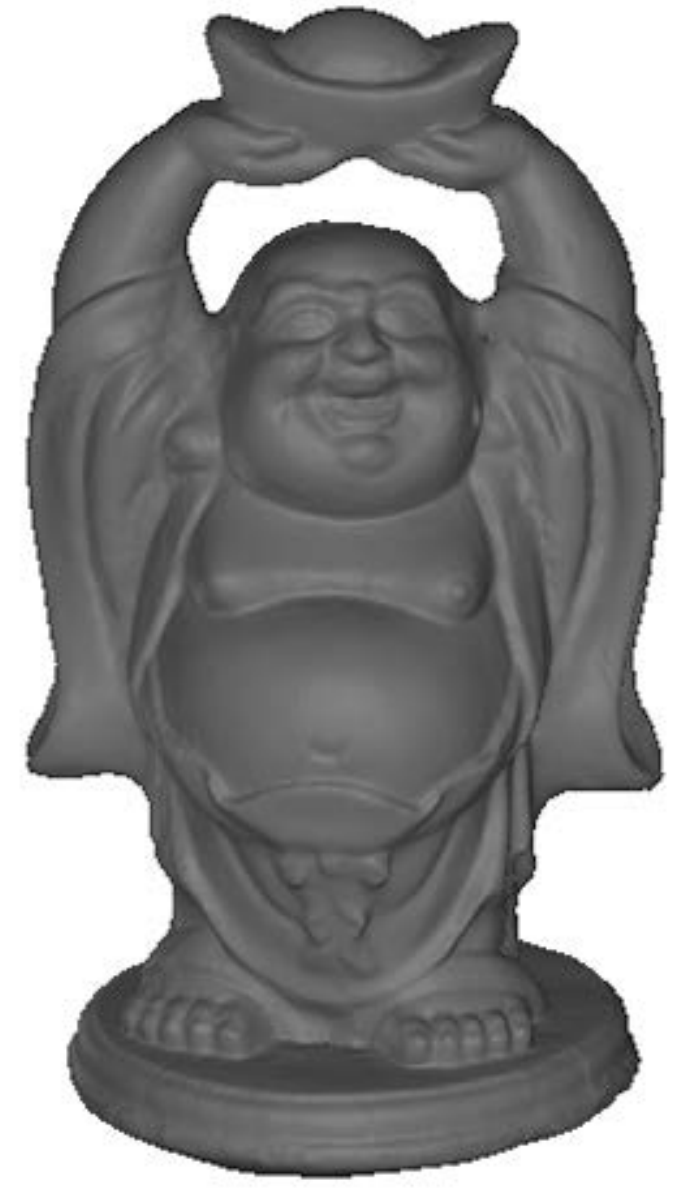}
		\\
		
		\includegraphics[width = 0.31 \linewidth]{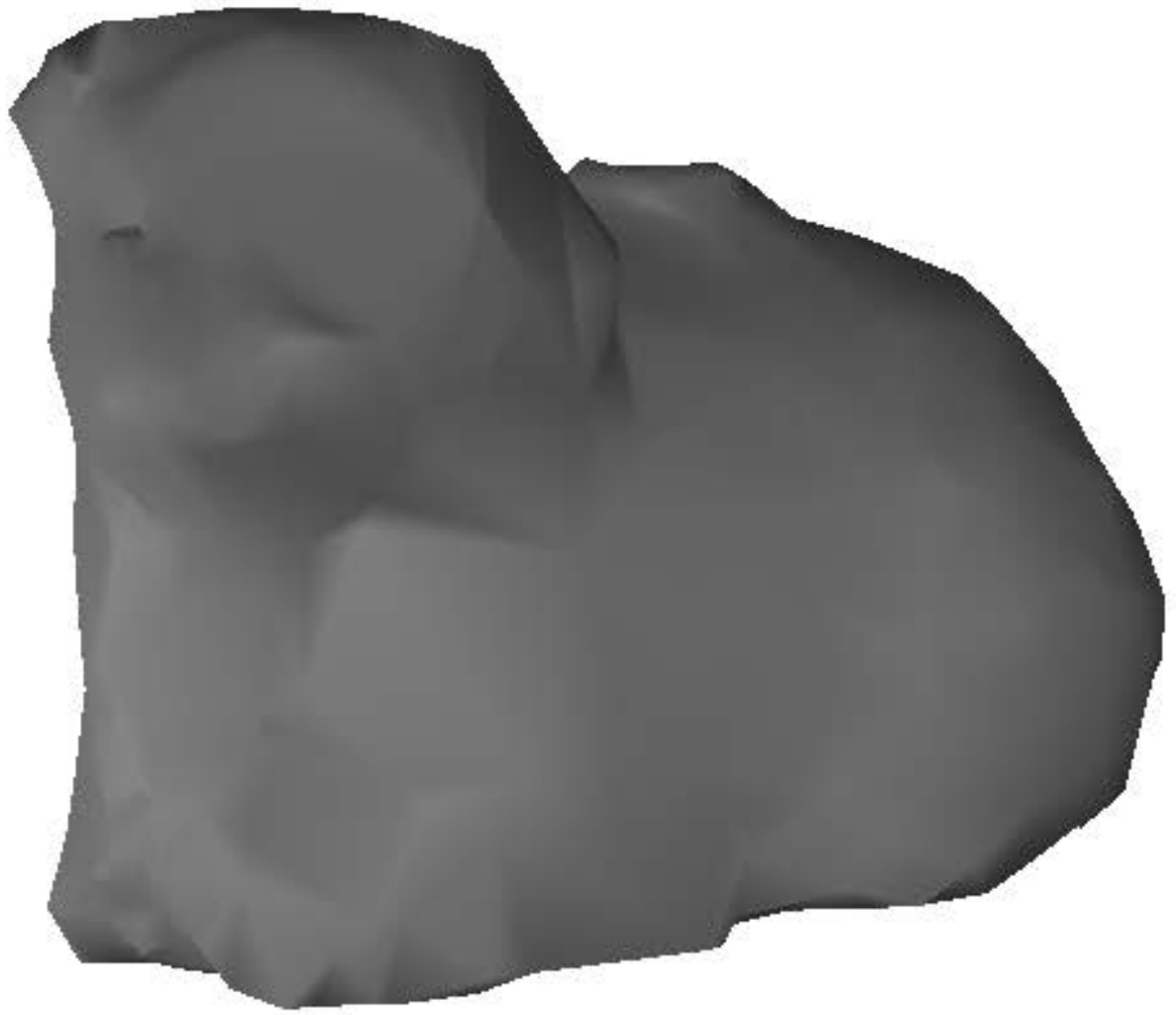}
		&\includegraphics[width= 0.3 \linewidth]{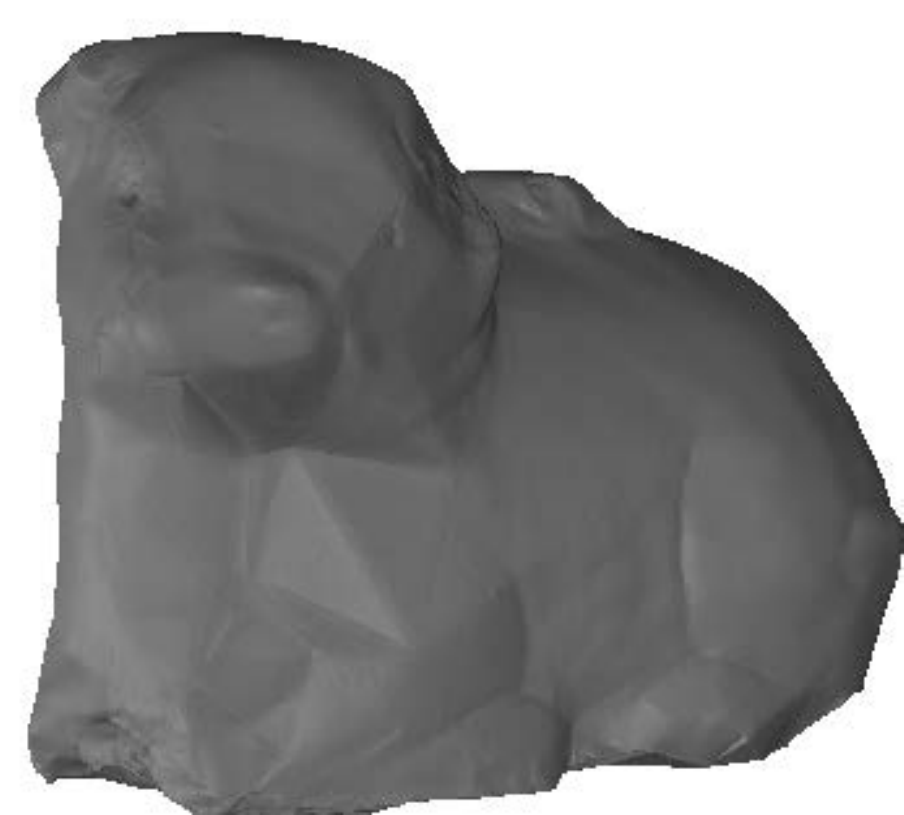}
		&\includegraphics[width= 0.3 \linewidth]{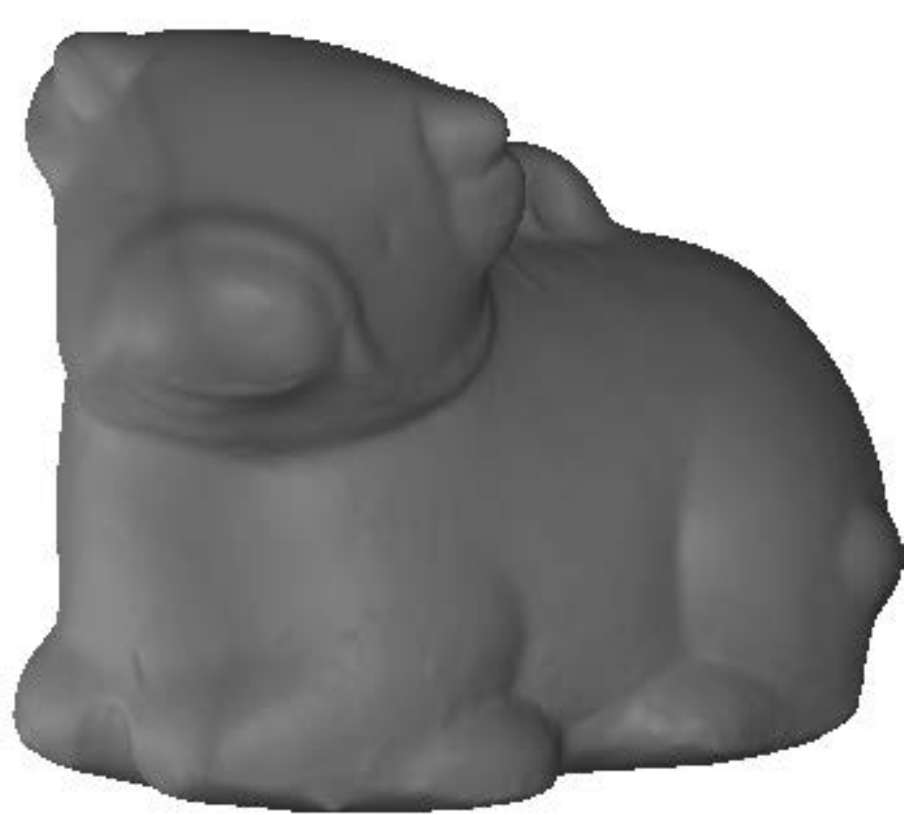}
		\\
				
		\includegraphics[width = 0.31 \linewidth]{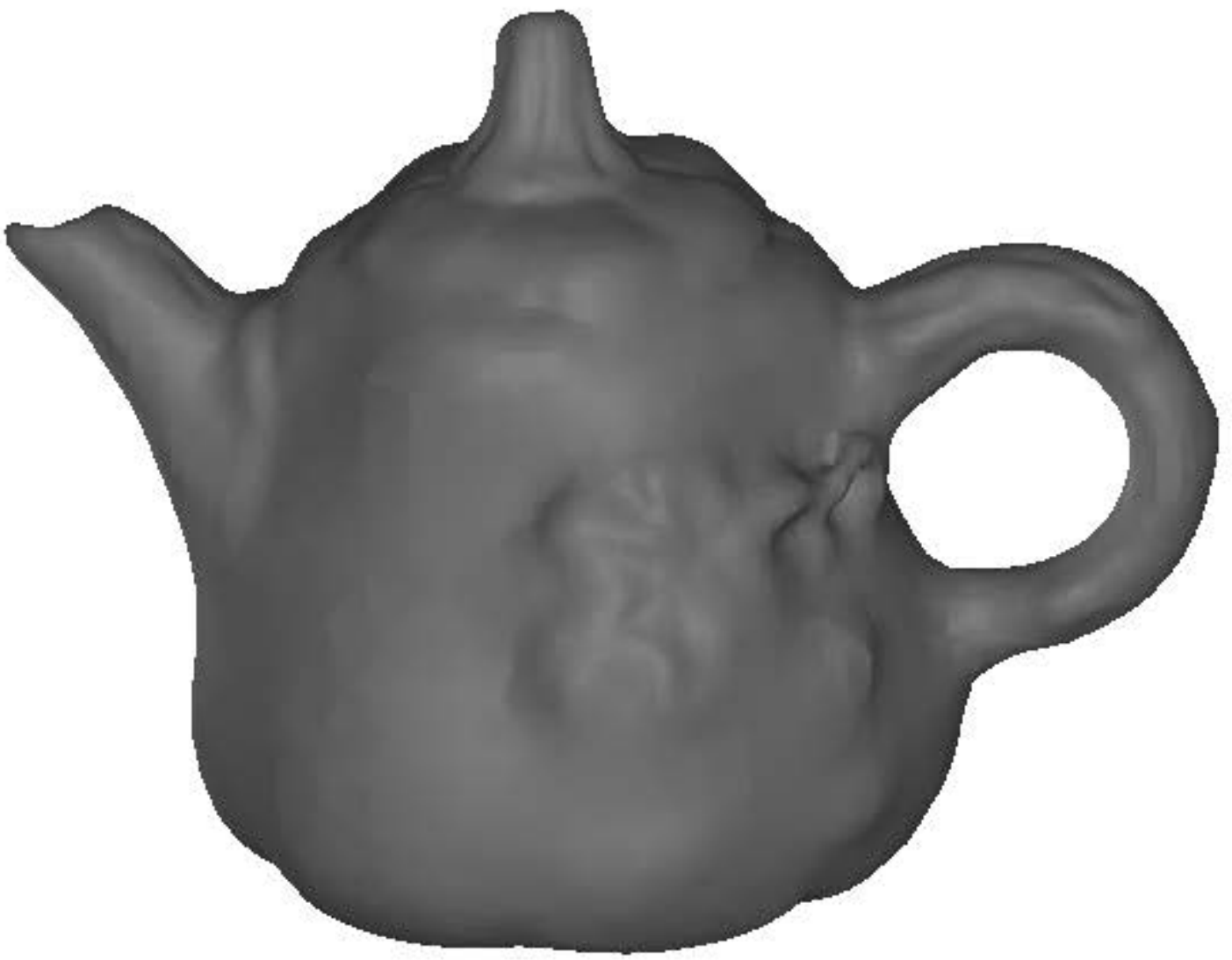}
		&\includegraphics[width= 0.3 \linewidth]{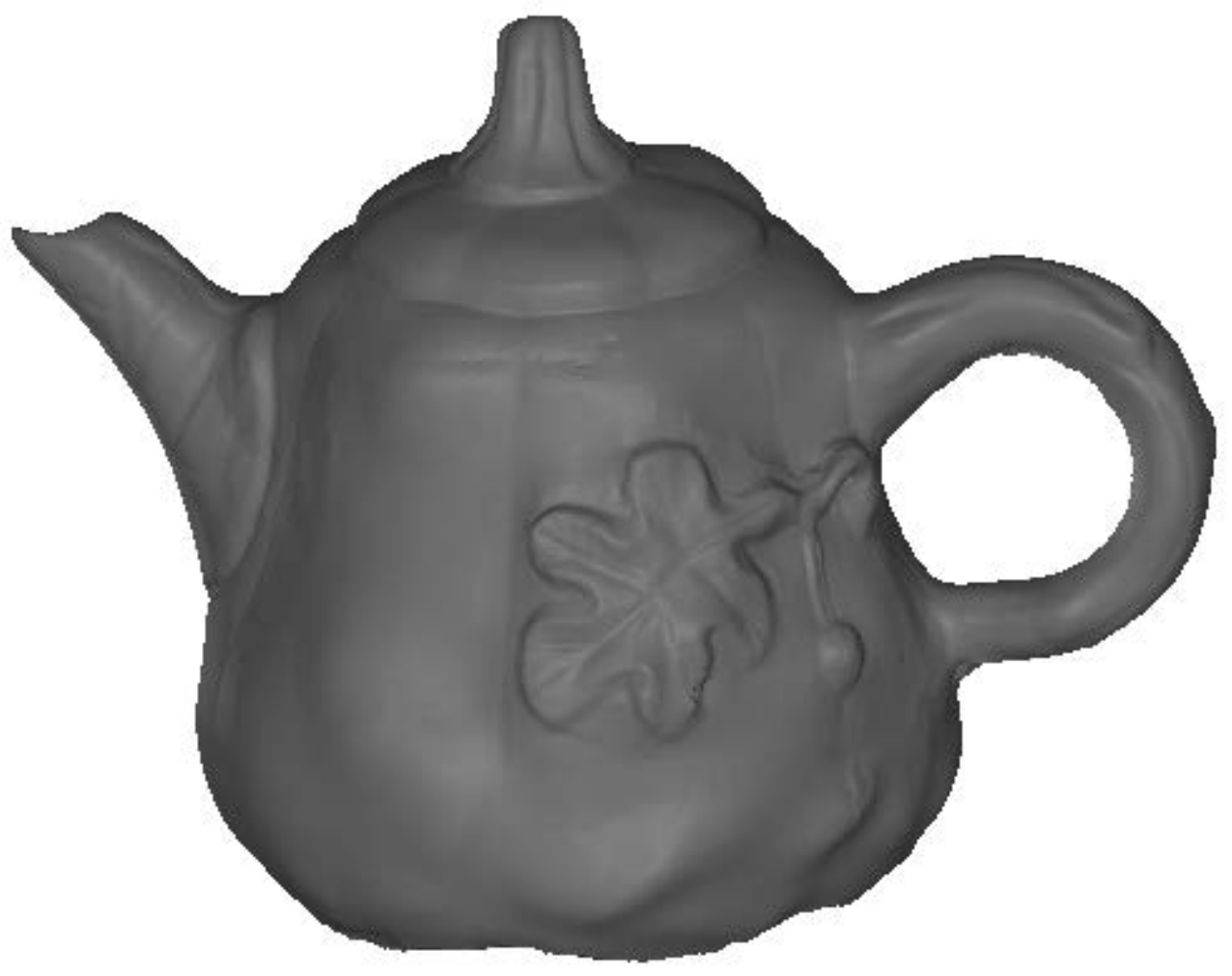}
		&\includegraphics[width= 0.3 \linewidth]{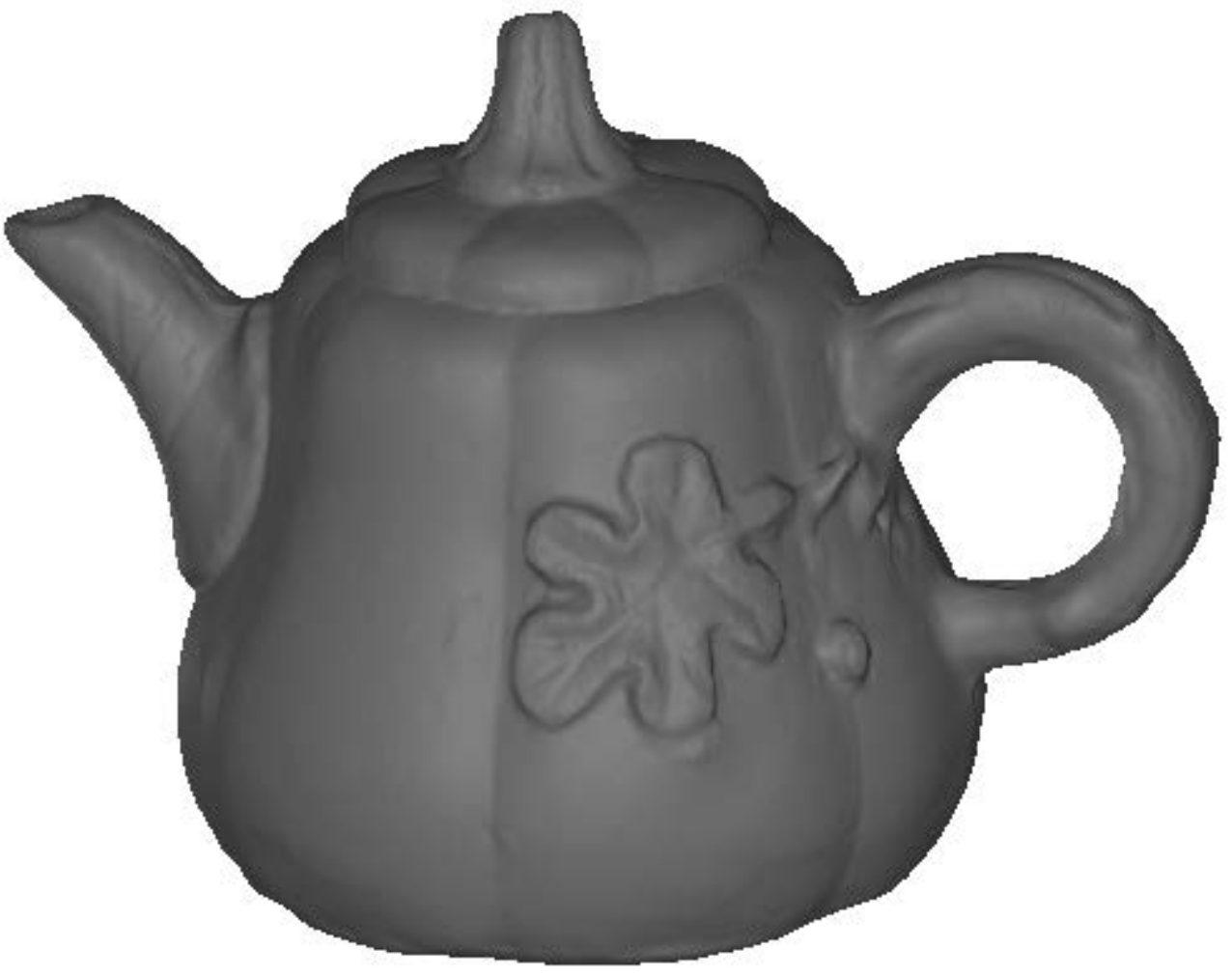}
		\\
				
		\includegraphics[width = 0.3 \linewidth]{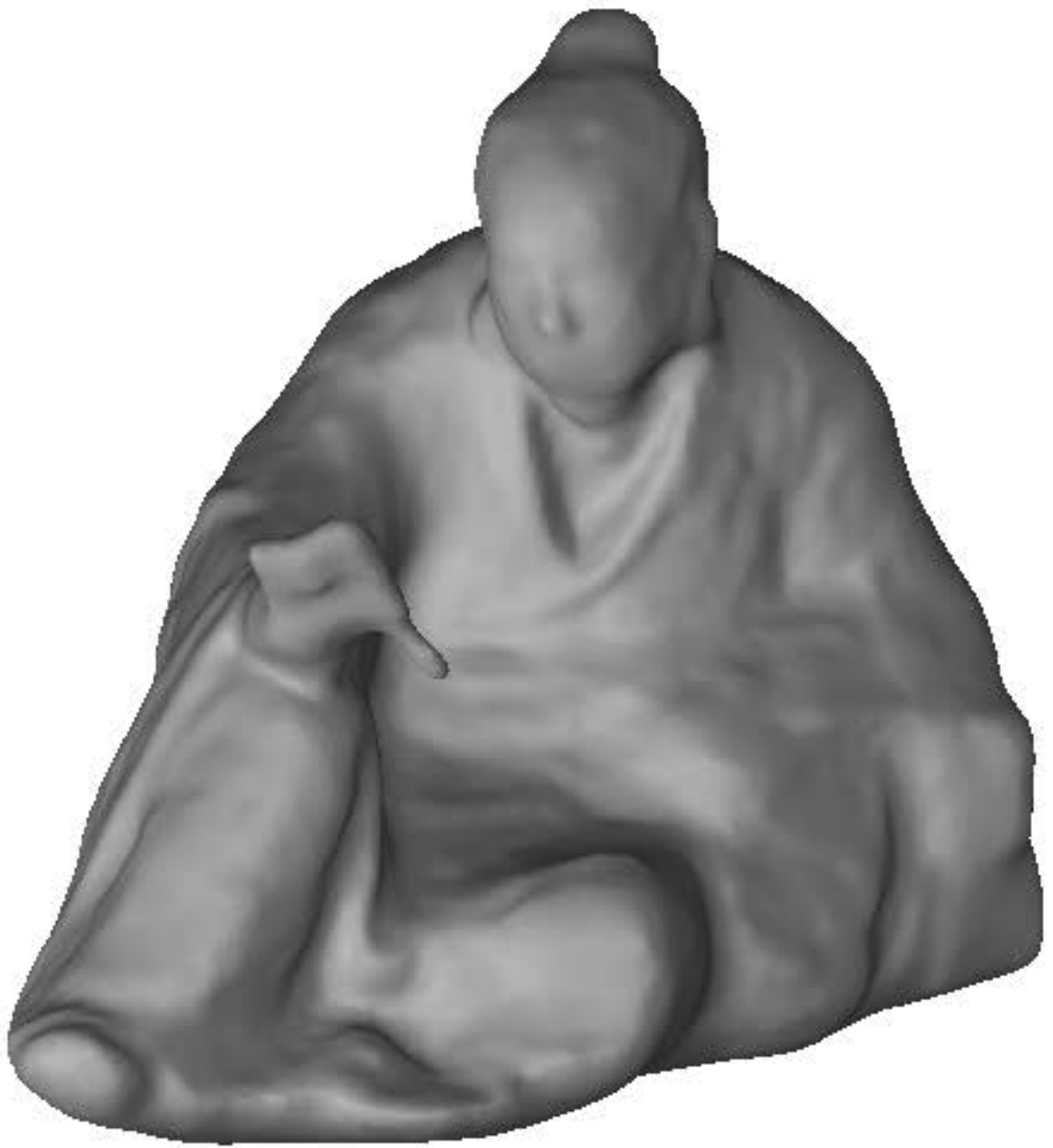}	
		&\includegraphics[width= 0.3 \linewidth]{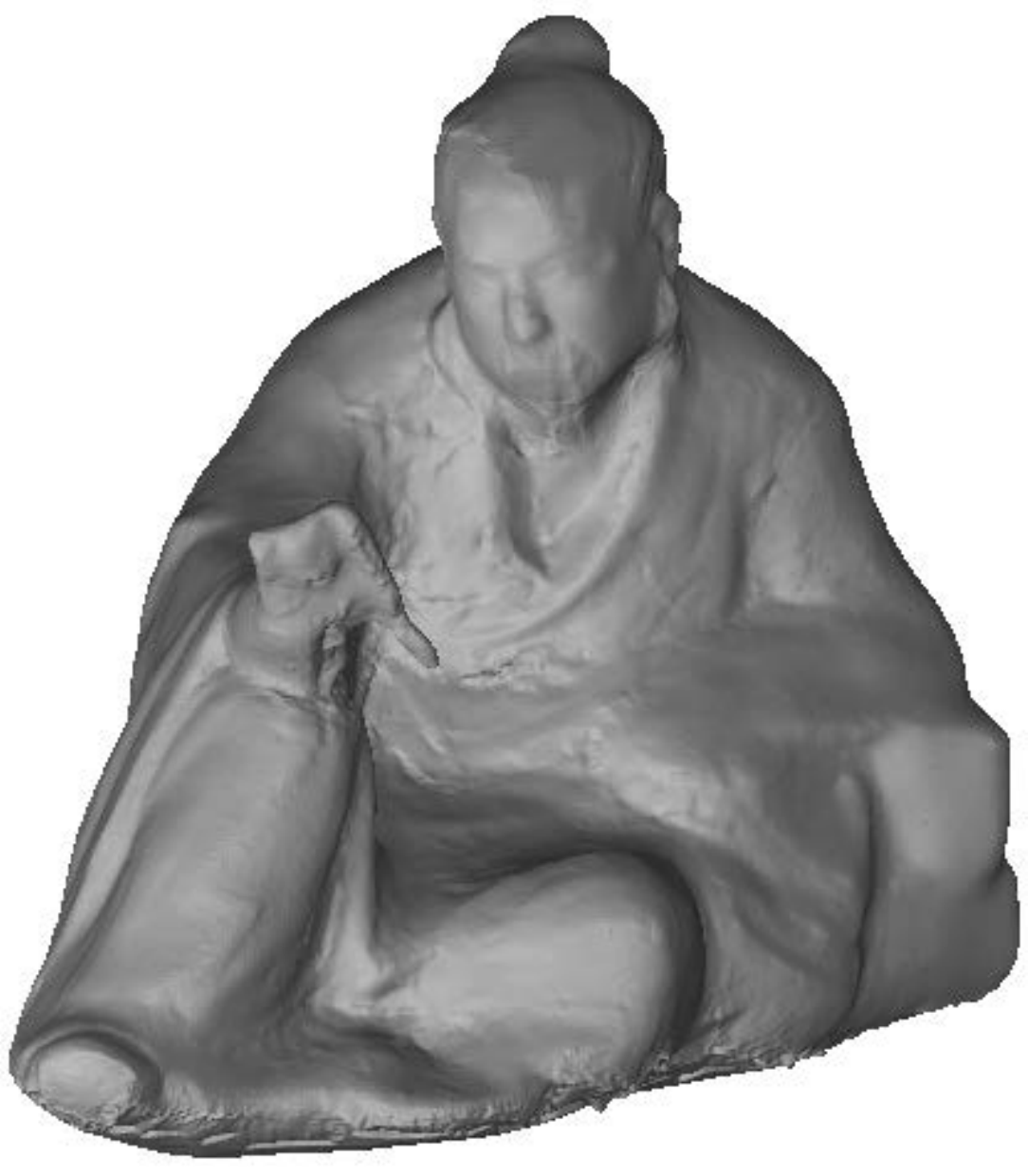}
		&\includegraphics[width= 0.3 \linewidth]{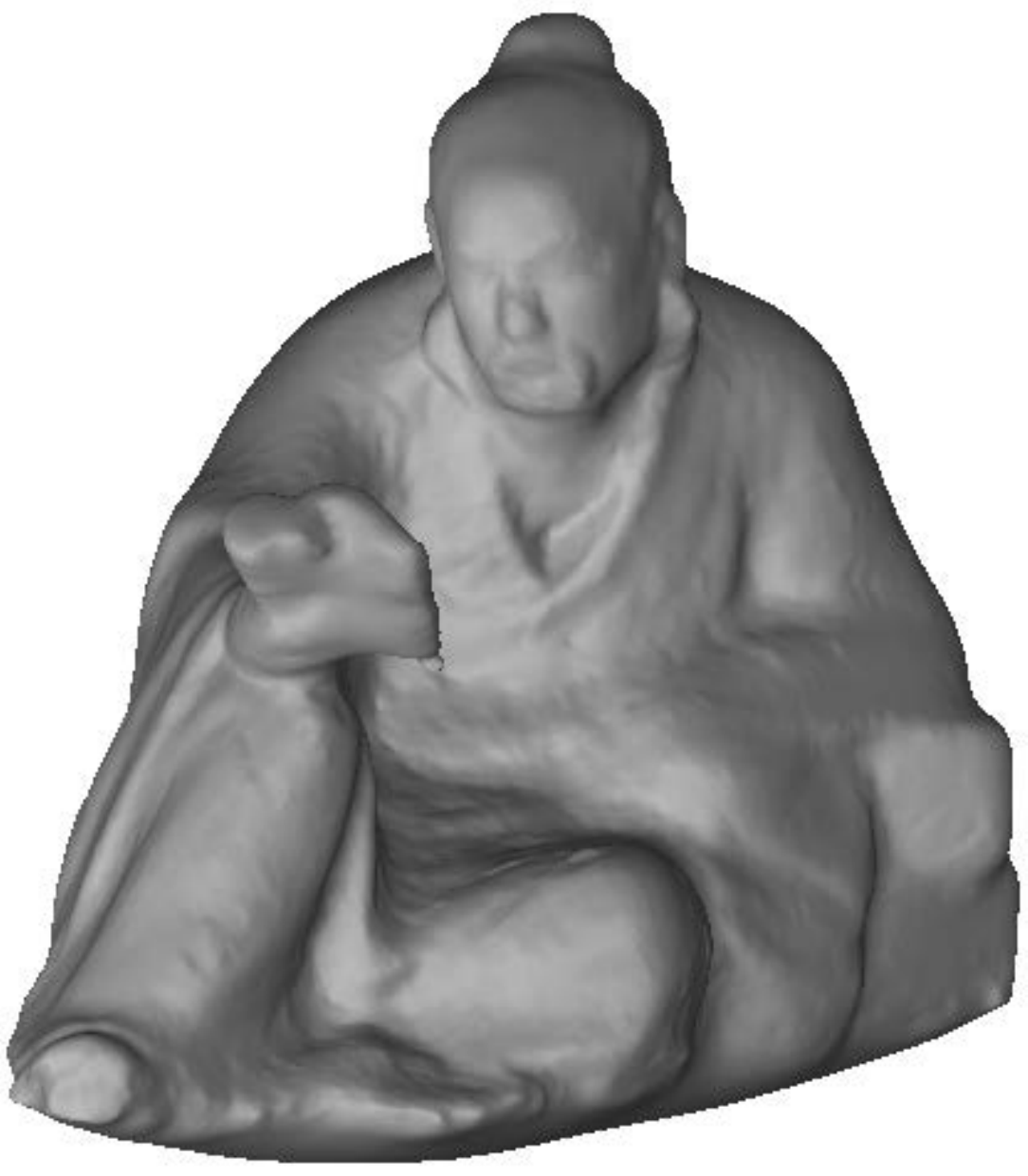}
		\\
		(a) & (b) & (c)
	\end{tabular}
	\caption{Reconstructed shapes by PJ16\cite{park2016robust} (b) and our method (c) given the initial mesh from (a).}
	\label{fig:shape_benchmark}
\end{figure}

We employ GIPUMA\cite{Galliani2015} to generate initial points and only keep points with reprojection error less than 0.5 pixels.
At very large textureless regions, we also manually establish some corresponding points to facilitate further processing.
Note that our method can be initialized by a small set of discrete 3D points, while PJ16 \cite{park2016robust} needs an initial mesh as initialization. Therefore, we apply the Poisson surface reconstruction on the initial 3D points to generate a mesh for PJ16 \cite{park2016robust}, and feed the initial 3D points to our method for initialization.
The initial meshes are shown in \figref{shape_benchmark} (a).
\figref{shape_benchmark} (b) and (c) are the final results by PJ16 \cite{park2016robust} and our method respectively.
The mean and median shape errors are shown in \tabref{benchmark_mesh_error1}. 
Our method reduces the average mean errors across all examples from 1.04 mm to 0.52 mm (a drop of 50\%), and from 0.82 mm to 0.50 mm (a drop of 40\%) for the average median error.
As we can tell from \figref{shape_benchmark},  PJ16\cite{park2016robust} often fails on textureless regions with non-Lambertian reflectances,  such as the ear of COW or the book of READING, where the initial mesh is quite unreliable. 
In comparison our method is much more robust at these regions.

\begin{figure*}[!htb]
\centering
	\begin{tabular}{@{\hspace{1mm}}c@{\hspace{1mm}}c@{\hspace{1mm}}c@{\hspace{1mm}}c@{\hspace{1mm}}c}
		\includegraphics[width = 0.195 \linewidth]{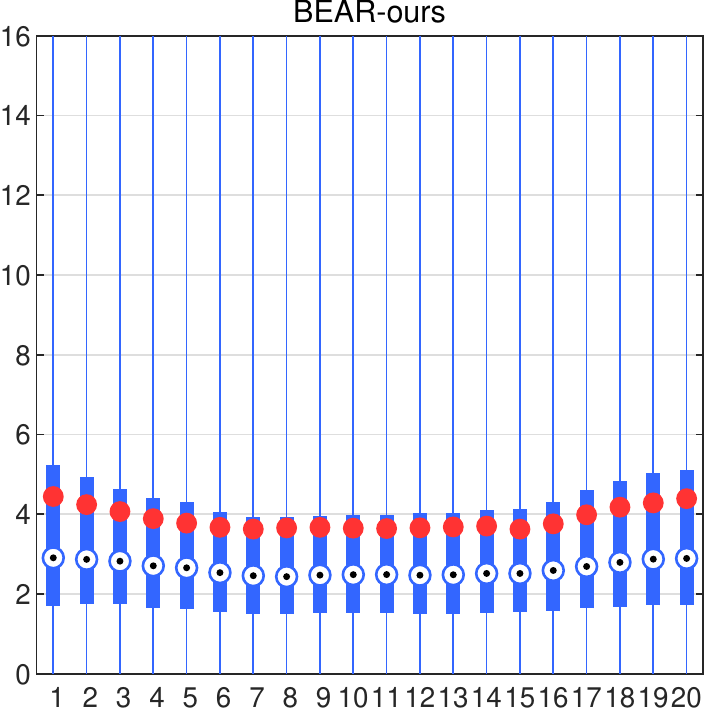}
		&\includegraphics[width = 0.195 \linewidth]{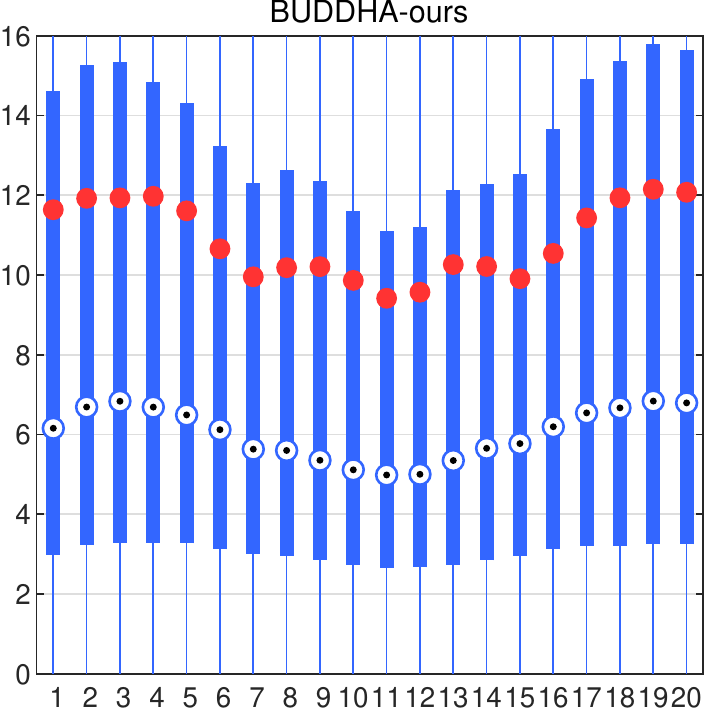}
		&\includegraphics[width = 0.195 \linewidth]{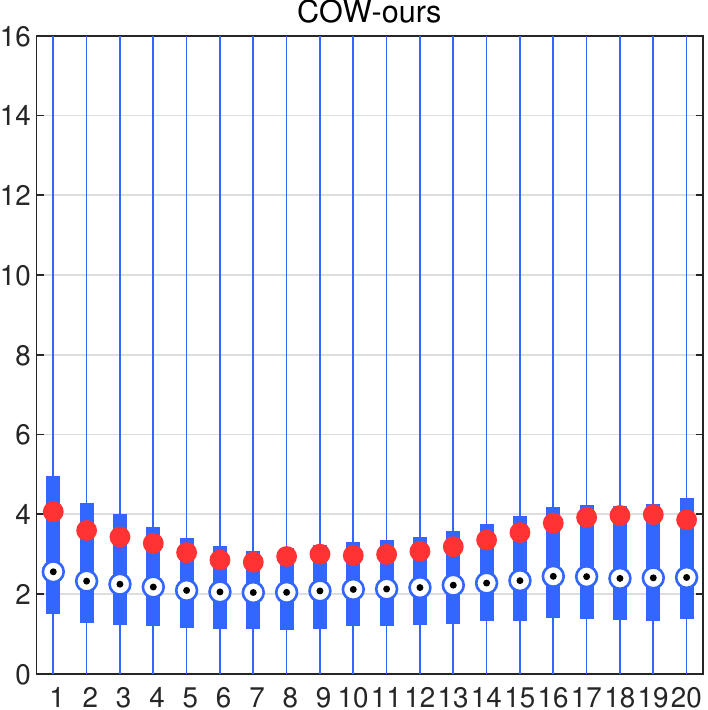}
		&\includegraphics[width = 0.195 \linewidth]{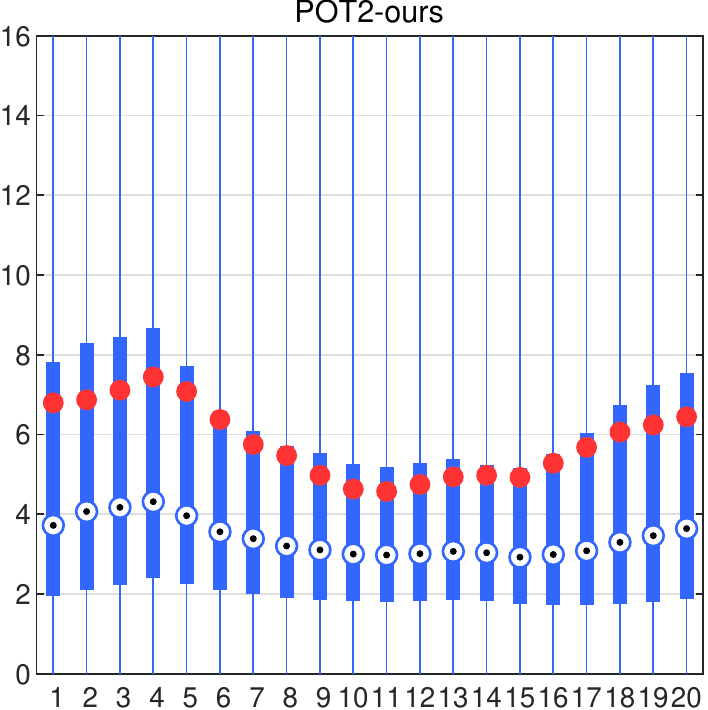}
		&\includegraphics[width = 0.195 \linewidth]{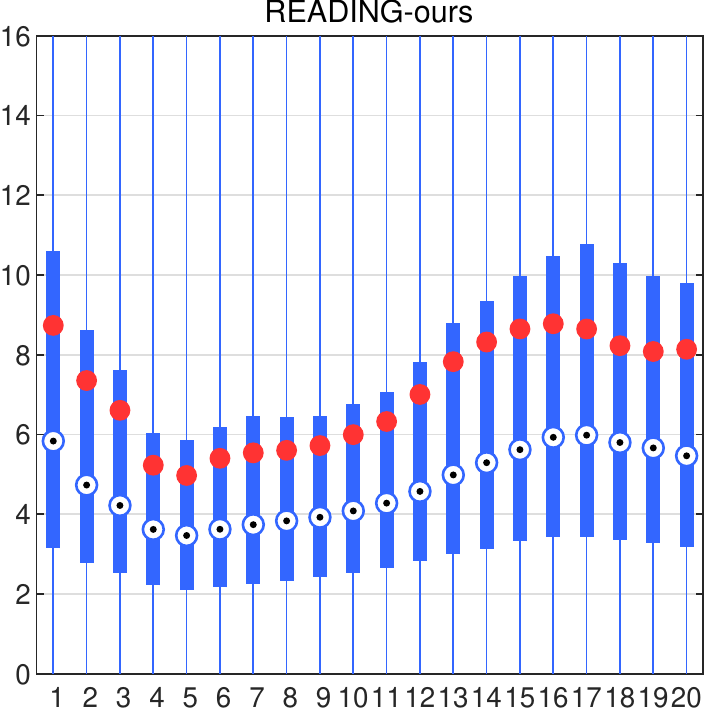}	
		\\
		\includegraphics[width= 0.195 \linewidth]{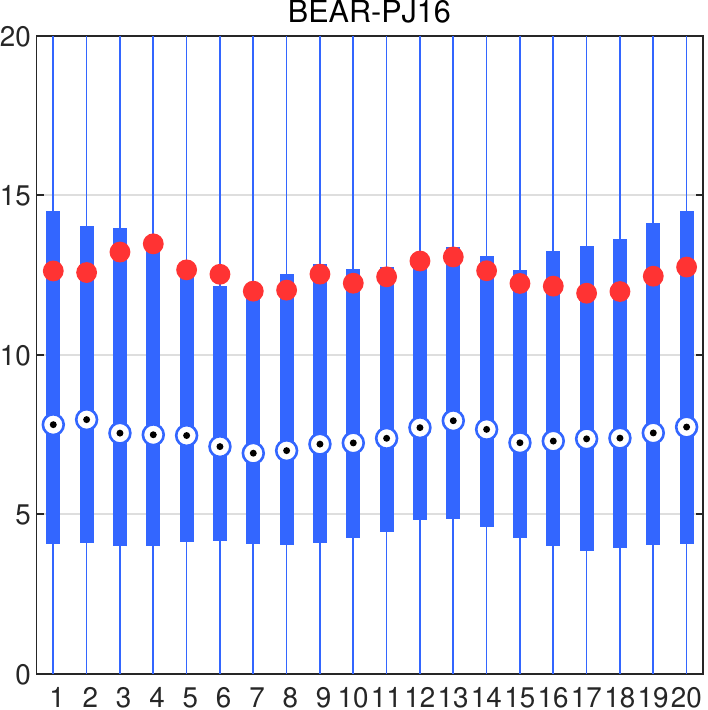}
		&\includegraphics[width= 0.195 \linewidth]{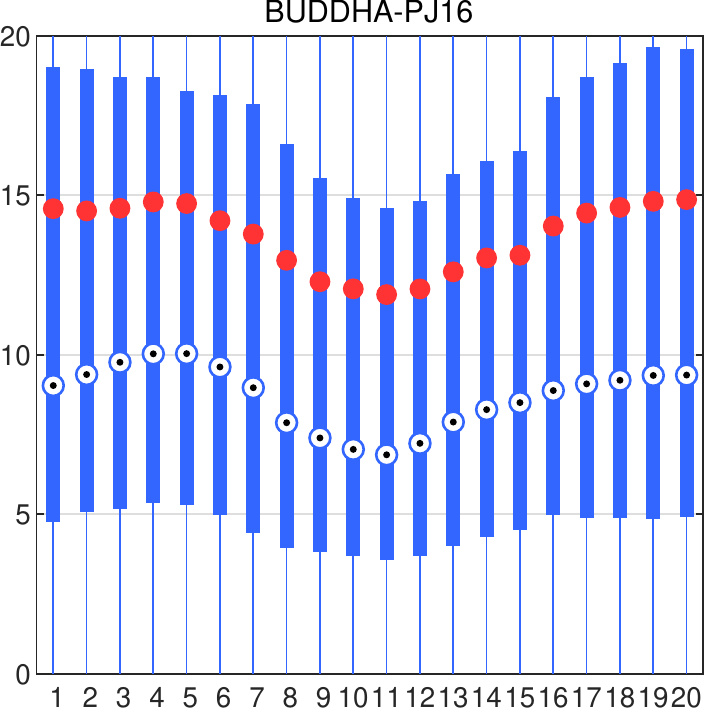}
		&\includegraphics[width= 0.195 \linewidth]{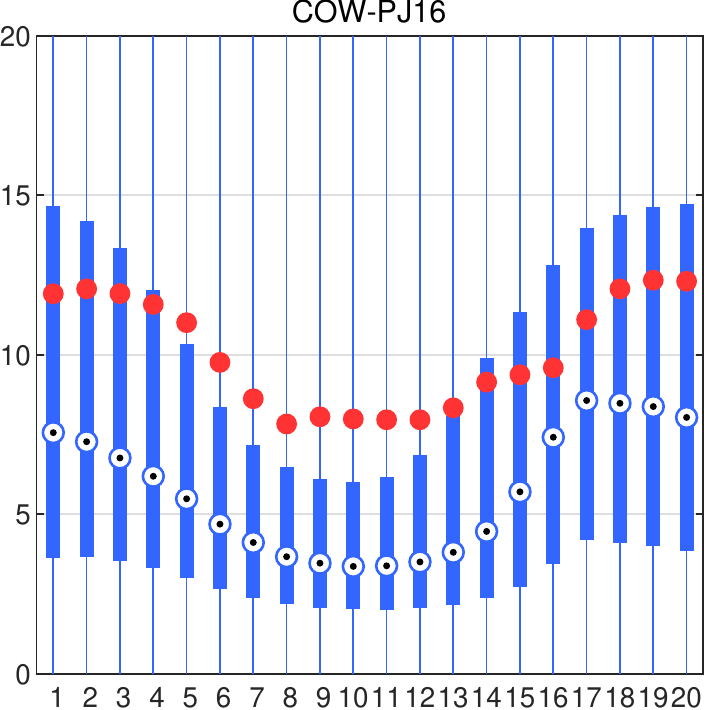}
		&\includegraphics[width= 0.195 \linewidth]{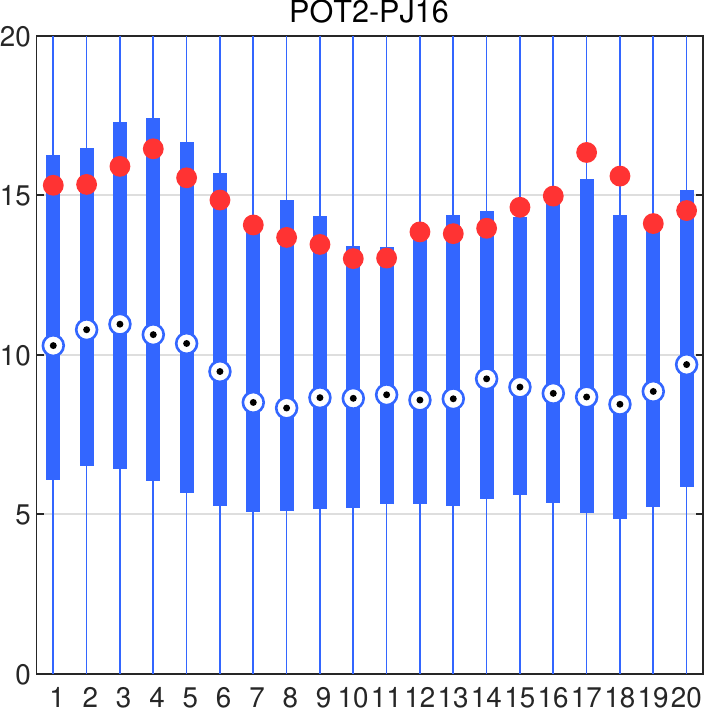}
		&\includegraphics[width= 0.195 \linewidth]{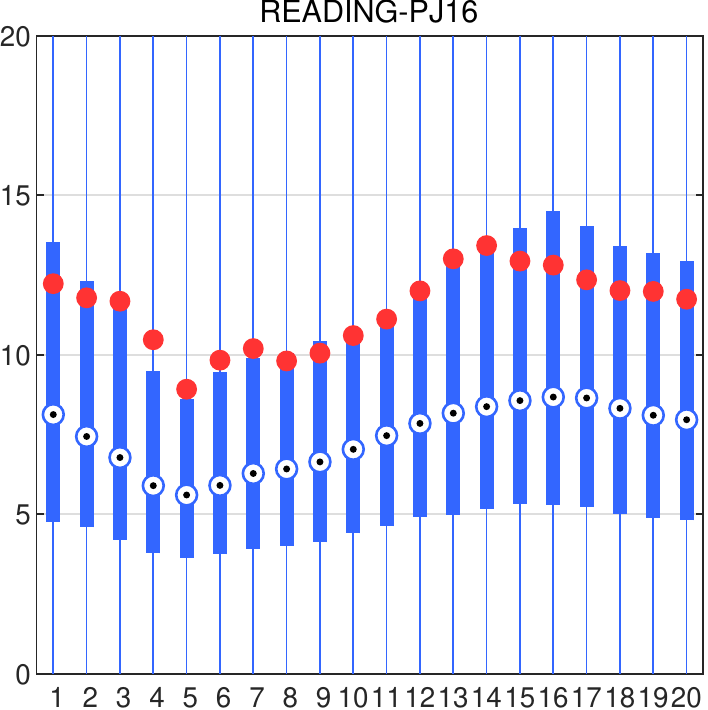}
		\\
	\end{tabular}
	\caption{Normal angular error statistics by using our method (top row) and PJ16\cite{park2016robust} (bottom row) on `DiLiGenT-MV'. Each subplot shows the results by one evaluated method for all views of each data; the X-axis is the ID of view number, and the Y -axis is the angular error (in degree); the statistics of angular errors for all pixels per normal map are displayed using the box-and-whisker plot: The red dot indicates the mean value, the black dot is the median, the top and bottom bounds of the blue box indicate the first and third quartile values, and the top and bottom ends of the vertical blue line indicate the minimum and maximum errors.}
\label{fig:normal_benchmark}
\end{figure*}

For further evaluation, we also evaluate the normal angular errors for each viewpoint of both methods in \figref{normal_benchmark}. As illustrated in \figref{normal_benchmark}, our method outperforms PJ16\cite{park2016robust} overall, since our method is based on isotropic reflectance, which is a more realistic assumption in dealing with real objects. \figref{normal_benchmark} further proves that the initial 3D points improve the final normal enormously (e.g., COW) and large errors appear in occluded and concave areas (e.g., BUDDHA), owing to the missing of image observation and strong inter-reflections. These are consistent with the observation of the mesh estimation.


\begin{table*}
\centering
\caption{Mean and median shape errors (in mm) of \cite{park2016robust} and ours on `DiLiGenT-MV'.}
    \vspace{-0.2cm}
	\begin{tabular}{c|ccccccc}
       \hline
		\hline
		& & BEAR & BUDDHA & COW & POT2 & READING & Avg.
		\\
	    \hline
		\multirow{2}{*}{mean} & PJ16\cite{park2016robust}& 1.08 & 1.05 & 0.57 & 1.54 & 0.96 & 1.04
		\\
		     & ours & \textbf{0.56} & \textbf{0.63} & \textbf{0.38} & \textbf{0.55} & \textbf{0.49} & \textbf{0.52}
		\\
		       \hline \hline
		\multirow{2}{*}{median} & PJ16\cite{park2016robust} & 0.94 & 0.75 & 0.43 & 1.34 & 0.66 & 0.82
		\\
		       & ours & \bf{0.55} & \bf{0.62} & \textbf{0.37} & \bf{0.54} & \bf{0.45} & \bf{0.50}
		\\
		\hline
	\end{tabular}
   \vspace{-0.4cm}
	\label{tab:benchmark_mesh_error1}
\end{table*}


\section{Discussion}
We propose a multi-view photometric stereo method to capture both the shape and reflectance of real objects.
Our method is general and works with spatially varying isotropic BRDFs.
It involves simple hardware setup of a video camera and some LEDs.
The captured 3D shape is accurate up to $0.5$ millimeters and the reflectance has a RMSE as low as 9\%. We also quantitatively evaluate state-of-the-art MVPS using a newly collected benchmark dataset `DiLiGenT-MV', which is publicly available for inspiring future research.

Our method has a few limitations.
First, our method cannot model anisotropic material. It also cannot handle translucent objects and mirror surfaces.
Second, our ring-light capture setup contains only two circles of LEDs.
Hence, we only capture the BRDF of a point with two different $\theta_d$ values.
As a result, during reflectance capturing, we can only discretize $\theta_d$ to two levels, and cannot capture Fresnel effects faithfully.
We could increase the number of circles of LED lights, or fit parametric Fresnel terms \cite{Schlick1994} to solve this problem.
Third, the calibration of our system still requires some skills for amateur users.
An interesting direction is to extend the compressive sensing framework in \cite{Yang2015} to strategically plan the illumination. 
Further enhancing its robustness to noisy 3D points from SfM is another interesting direction for future research. Last, it is still challenging for MVS and MVPS methods to reconstruct high-quality 3D points from the texture-less or highly non-Lambertian surfaces of  `DiGiLenT-MV'.

\section{Acknowledgement}

We thank Todd Zickler, Kyros Kutulakos, and Stephen
Lin for their helpful discussions and suggestions. We also thank Sai-Kit Yeung and Zhipeng Mo for helping on building the `DiLiGenT-MV' dataset. This work is 
partially supported by the National Natural Science Foundation of China under Grant No.61872012 and 61425025, the Key Research and Development Program of Zhejiang Province of China under Grant No.2018C03051, and Principles Research for the Conservation of Heritage Sites Grant S17-176000-016. Changyu Diao is supported by the Key Scientific Research Base for Digital Conservation of Cave Temples in Zhejiang University, State Administration for Cultural Heritage of China. Ping Tan is supported by the Canada NSERC Discovery Grant No.611664.

{\small
	\bibliographystyle{ieee}
	\bibliography{3dscanner}

\begin{thebibliography}{10}\itemsep=-1pt

\bibitem{Aliaga2008}
D.~Aliaga and Y.~Xu.
\newblock Photogeometric structured light: A self-calibrating and
  multi-viewpoint framework for accurate 3d modeling.
\newblock In {\em Proc. of Computer Vision and Pattern Recognition}, 2008.

\bibitem{Alldrin2008}
N.~Alldrin, T.~Zickler, and D.~Kriegman.
\newblock Photometric stereo with non-parametric and spatially-varying
  reflectance.
\newblock In {\em Proc. of Computer Vision and Pattern Recognition}, 2008.

\bibitem{Besl1992}
P.~J. Besl and N.~D. McKay.
\newblock A method for registration of 3-d shapes.
\newblock {\em IEEE Transactions on Pattern Analysis and Machine Intelligence},
  14:239--256, 1992.

\bibitem{Chandraker2011}
M.~Chandraker, J.~Bai, and R.~Ramamoorthi.
\newblock A theory of differential photometric stereo for unknown brdfs.
\newblock In {\em Proc. of Computer Vision and Pattern Recognition}, 2011.

\bibitem{Furukawa2010}
Y.~Furukawa and J.~Ponce.
\newblock Accurate, dense, and robust multiview stereopsis.
\newblock {\em IEEE Transactions on Pattern Analysis and Machine Intelligence},
  32:1362--1376, 2010.

\bibitem{Ghosh2009}
A.~Ghosh, T.~Chen, P.~Peers, C.~A. Wilson, and P.~Debevec.
\newblock Estimating specular roughness and anisotropy from second order
  spherical gradient illumination.
\newblock {\em Computer Graphics Forum}, 28, 2009.

\bibitem{Goldman2005}
D.~B. Goldman, B.~Curless, A.~Hertzmann, and S.~M. Seitz.
\newblock Shape and spatially-varying brdfs from photometric stereo.
\newblock In {\em Proc. of International Conference on Computer Vision}, pages
  341--348, 2005.

\bibitem{han2015photometric}
T.-Q. Han and H.-L. Shen.
\newblock Photometric stereo for general brdfs via reflection sparsity
  modeling.
\newblock {\em IEEE Transactions on Image Processing}, 24(12):4888--4903, 2015.

\bibitem{Hartley2003}
R.~Hartley and A.~Zisserman.
\newblock {\em Multiple View Geometry in Computer Vision}.
\newblock Cambridge University Press, New York, NY, USA, 2 edition, 2003.

\bibitem{Hernandez2008}
C.~Hernandez, G.~Vogiatzis, and R.~Cipolla.
\newblock Multiview photometric stereo.
\newblock {\em IEEE Transactions on Pattern Analysis and Machine Intelligence},
  30:548--554, 2008.

\bibitem{higo2009}
T.~Higo, Y.~Matsushita, N.~Joshi, and K.~Ikeuchi.
\newblock A hand-held photometric stereo camera for 3-d modeling.
\newblock In {\em Proc. of International Conference on Computer Vision}, pages
  1234--1241. IEEE, 2009.

\bibitem{Holroyd2008}
M.~Holroyd, J.~Lawrence, G.~Humphreys, and T.~Zickler.
\newblock A photometric approach for estimating normals and tangents.
\newblock {\em ACM Transactions on Graphics}, 27, 2008.

\bibitem{Holroyd2010}
M.~Holroyd, J.~Lawrence, and T.~Zickler.
\newblock A coaxial optical scanner for synchronous acquisition of 3d geometry
  and surface reflectance.
\newblock {\em ACM Transactions on Graphics}, 2010.

\bibitem{ikehata2014photometric}
S.~Ikehata, D.~Wipf, Y.~Matsushita, and K.~Aizawa.
\newblock Photometric stereo using sparse bayesian regression for general
  diffuse surfaces.
\newblock {\em IEEE Transactions on Pattern Analysis and Machine Intelligence},
  36(9):1816--1831, 2014.

\bibitem{iwahori1990}
Y.~Iwahori, H.~Sugie, and N.~Ishii.
\newblock Reconstructing shape from shading images under point light source
  illumination.
\newblock In {\em Proc. of International Conference on Pattern Recognition},
  volume~1, pages 83--87. IEEE, 1990.

\bibitem{jensen2014large}
R.~Jensen, A.~Dahl, G.~Vogiatzis, E.~Tola, and H.~Aan{\ae}s.
\newblock Large scale multi-view stereopsis evaluation.
\newblock In {\em Proc. of Computer Vision and Pattern Recognition}, pages
  406--413, 2014.

\bibitem{kang2019learning}
K.~Kang, C.~Xie, C.~He, M.~Yi, m.~Gu, Z.~Chen, K.~Zhou, and H.~Wu.
\newblock Learning efficient illumination multiplexing for joint capture of
  reflectance and shape.
\newblock {\em ACM Transactions on Graphics}, 38(6):165, 2019.

\bibitem{Kazhdan2006}
M.~Kazhdan, M.~Bolitho, and H.~Hoppe.
\newblock Poisson surface reconstruction.
\newblock In {\em Proc. of Eurographics Symposium on Geometry Processing},
  pages 61--70, 2006.

\bibitem{knapitsch2017tanks}
A.~Knapitsch, J.~Park, Q.-Y. Zhou, and V.~Koltun.
\newblock Tanks and temples: Benchmarking large-scale scene reconstruction.
\newblock {\em ACM Transactions on Graphics}, 36(4):78, 2017.

\bibitem{Lawrence2006}
J.~Lawrence, A.~Ben-Artzi, C.~DeCoro, W.~Matusik, H.~Pfister, R.~Ramamoorthi,
  and S.~Rusinkiewicz.
\newblock Inverse shade trees for non-parametric material representation and
  editing.
\newblock {\em ACM Transactions on Graphics}, 25:735--745, July 2006.

\bibitem{Lensch2003}
H.~Lensch, J.~Kautz, M.~Goesele, W.~Heidrich, and H.-P. Seidel.
\newblock Image-based reconstruction of spatial appearance and geometric
  detail.
\newblock {\em ACM Transactions on Graphics}, 22:234--257, 2003.

\bibitem{Levoy2000}
M.~Levoy, K.~Pulli, B.~Curless, S.~Rusinkiewicz, D.~Koller, L.~Pereira,
  M.~Ginzton, S.~Anderson, J.~Davis, J.~Ginsberg, J.~Shade, and D.~Fulk.
\newblock The digital michelangelo project: 3d scanning of large statues.
\newblock In {\em Proc. of SIGGRAPH}, pages 131--144, 2000.

\bibitem{Lhuillier2005}
M.~Lhuillier and L.~Quan.
\newblock A quasi-dense approach to surface reconstruction from uncalibrated
  images.
\newblock {\em IEEE Transactions on Pattern Analysis and Machine Intelligence},
  27:418--433, 2005.

\bibitem{Lim2005}
J.~Lim, J.~Ho, M.-H. Yang, and D.~Kriegman.
\newblock Passive photometric stereo from motion.
\newblock In {\em Proc. of International Conference on Computer Vision}, pages
  1635--1642, 2005.

\bibitem{Lu2013}
F.~Lu, Y.~Matsushita, I.~Sato, T.~Okabe, and Y.~Sato.
\newblock Uncalibrated photometric stereo for unknown isotropic reflectances.
\newblock In {\em Proc. of Computer Vision and Pattern Recognition}, 2013.

\bibitem{Ma2007}
W.-C. Ma, T.~Hawkins, P.~Peers, C.-F. Chabert, M.~Weiss, and P.~Debevec.
\newblock Rapid acquisition of specular and diffuse normal maps from polarized
  spherical gradient illumination.
\newblock In {\em Proc. of Eurographics Symposium on Geometry Processing},
  2007.

\bibitem{Nehab2005}
D.~Nehab, S.~Rusinkiewicz, J.~Davis, and R.~Ramamoorthi.
\newblock Efficiently combining positions and normals for precise 3d geometry.
\newblock {\em ACM Transactions on Graphics}, 24:536--543, 2005.

\bibitem{Alldrin2007}
A.~Neil and K.~David.
\newblock Toward reconstructing surfaces with arbitrary isotropic reflectance :
  A stratified photometric stereo approach.
\newblock In {\em Proc. of International Conference on Computer Vision}, 2007.

\bibitem{papadhimitri2013}
T.~Papadhimitri and P.~Favaro.
\newblock A new perspective on uncalibrated photometric stereo.
\newblock In {\em Proc. of Computer Vision and Pattern Recognition}, pages
  1474--1481, 2013.

\bibitem{papadhimitri2014}
T.~Papadhimitri and P.~Favaro.
\newblock Uncalibrated near-light photometric stereo.
\newblock 2014.

\bibitem{Park2013}
J.~Park, S.~N. Sinha, Y.~Matsushita, Y.-W. Tai, and I.~S. Kweon.
\newblock Multiview photometric stereo using planar mesh parameterization.
\newblock In {\em Proc. of International Conference on Computer Vision}, 2013.

\bibitem{park2016robust}
J.~Park, S.~N. Sinha, Y.~Matsushita, Y.-W. Tai, and I.~S. Kweon.
\newblock Robust multiview photometric stereo using planar mesh
  parameterization.
\newblock {\em IEEE Transactions on Pattern Analysis and Machine Intelligence},
  39(8):1591--1604, 2016.

\bibitem{Yvain2018}
Y.~Qu\'{e}au, B.~Durix, T.~Wu, D.~Cremers, F.~Lauze, and J.-D. Durou.
\newblock Led-based photometric stereo: Modeling, calibration and numerical
  solution.
\newblock {\em Journal of Mathematical Imaging and Vision}, 60(3):313--340,
  2018.

\bibitem{Yvain2017}
Y.~Qu\'{e}au, T.~Wu, F.~Lauze, J.-D. Durou, and D.~Cremers.
\newblock A non-convex variational approach to photometric stereo under
  inaccurate lighting.
\newblock In {\em Proc. of Computer Vision and Pattern Recognition}, 2017.

\bibitem{Raskar2004}
R.~Raskar, K.-H. Tan, R.~Feris, J.~Yu, and M.~Turk.
\newblock Non-photorealistic camera: depth edge detection and stylized
  rendering using multi-flash imaging.
\newblock {\em ACM Transactions on Graphics}, 23:679--688, August 2004.

\bibitem{Ren2011}
P.~Ren, J.~Wang, J.~Snyder, X.~Tong, and B.~Guo.
\newblock Pocket reflectometry.
\newblock {\em ACM Transactions on Graphics}, 30(4), 2011.

\bibitem{Romeiro2010}
F.~Romeiro and T.~Zickler.
\newblock Inferring reflectance under real-world illumination.
\newblock {\em Technical Report TR-10-10, Harvard School of Engineering and
  Applied Sciences}, 2010.

\bibitem{Rusinkiewicz2002}
S.~Rusinkiewicz, O.~Hall-Holt, and M.~Levoy.
\newblock Real-time 3d model acquisition.
\newblock {\em ACM Transactions on Graphics}, 21:438--446, 2002.

\bibitem{Sato1997}
Y.~Sato, M.~D. Wheeler, and K.~Ikeuchi.
\newblock Object shape and reflectance modeling from observation.
\newblock In {\em Proc. of SIGGRAPH}, pages 379--387, 1997.

\bibitem{Schlick1994}
C.~Schlick.
\newblock An inexpensive {BRDF} model for physically-based rendering.
\newblock {\em Computer Graphics Forum}, 13(3):233--246, 1994.

\bibitem{schops2017multi}
T.~Schops, J.~L. Schonberger, S.~Galliani, T.~Sattler, K.~Schindler,
  M.~Pollefeys, and A.~Geiger.
\newblock A multi-view stereo benchmark with high-resolution images and
  multi-camera videos.
\newblock In {\em Proc. of Computer Vision and Pattern Recognition}, pages
  3260--3269, 2017.

\bibitem{seitz2006comparison}
S.~M. Seitz, B.~Curless, J.~Diebel, D.~Scharstein, and R.~Szeliski.
\newblock A comparison and evaluation of multi-view stereo reconstruction
  algorithms.
\newblock In {\em Proc. of Computer Vision and Pattern Recognition}, volume~1,
  pages 519--528. IEEE, 2006.

\bibitem{shen2017efficient}
H.-L. Shen, T.-Q. Han, and C.~Li.
\newblock Efficient photometric stereo using kernel regression.
\newblock {\em IEEE Transactions on Image Processing}, 26(1):439--451, 2017.

\bibitem{shi2019benchmark}
B.~Shi, Z.~Mo, Z.~Wu, D.~Duan, S.~Yeung, and P.~Tan.
\newblock A benchmark dataset and evaluation for non-lambertian and
  uncalibrated photometric stereo.
\newblock {\em IEEE Transactions on Pattern Analysis and Machine Intelligence},
  41(2):271--284, 2019.

\bibitem{shi2014bi}
B.~Shi, P.~Tan, Y.~Matsushita, and K.~Ikeuchi.
\newblock Bi-polynomial modeling of low-frequency reflectances.
\newblock {\em IEEE Transactions on Pattern Analysis and Machine Intelligence},
  36(6):1078--1091, 2014.

\bibitem{Galliani2015}
G.~Silvano, L.~Katrin, and S.~Konrad.
\newblock Massively parallel multiview stereopsis by surface normal diffusion.
\newblock In {\em Proc. of International Conference on Computer Vision}, 2015.

\bibitem{Snavely2006}
N.~Snavely, S.~M. Seitz, and R.~Szeliski.
\newblock Photo tourism: exploring photo collections in 3d.
\newblock {\em ACM Transactions on Graphics}, 25(3):835--846, July 2006.

\bibitem{strecha2008benchmarking}
C.~Strecha, W.~Von~Hansen, L.~Van~Gool, P.~Fua, and U.~Thoennessen.
\newblock On benchmarking camera calibration and multi-view stereo for high
  resolution imagery.
\newblock In {\em Proc. of Computer Vision and Pattern Recognition}, pages
  1--8. IEEE, 2008.

\bibitem{Tan2007}
P.~Tan, S.~P. Mallick, L.~Quan, D.~Kriegman, and T.~Zickler.
\newblock Isotropy, reciprocity and the generalized bas-relief ambiguity.
\newblock In {\em Proc. of Computer Vision and Pattern Recognition}, 2007.

\bibitem{Tan2011}
P.~Tan, L.~Quan, and T.~Zickler.
\newblock The geometry of reflectance symmetries.
\newblock {\em IEEE Transactions on Pattern Analysis and Machine Intelligence},
  33:2506--2520, 2011.

\bibitem{Ward1992}
G.~J. Ward.
\newblock Measuring and modeling anisotropic reflection.
\newblock In {\em Proc. of SIGGRAPH}, pages 265--272, 1992.

\bibitem{woodham1979photometric}
R.~J. Woodham.
\newblock Photometric stereo: A reflectance map technique for determining
  surface orientation from image intensity.
\newblock In {\em Image Understanding Systems and Industrial Applications I},
  volume 155, pages 136--144. International Society for Optics and Photonics,
  1979.

\bibitem{xie2015photometric}
W.~Xie, C.~Dai, and C.~C. Wang.
\newblock Photometric stereo with near point lighting: A solution by mesh
  deformation.
\newblock In {\em Proc. of Computer Vision and Pattern Recognition}, pages
  4585--4593, 2015.

\bibitem{Yang2015}
W.~Yang, Y.~Ji, H.~Lin, Y.~Yang, S.~B. Kang, and J.~Yu.
\newblock Ambient occlusion via compressive visibility estimation.
\newblock In {\em Proc. of Computer Vision and Pattern Recognition}, 2015.

\bibitem{Dong2010}
D.~Yue, W.~Jiaping, T.~Xin, S.~John, L.~Yanxiang, B.-E. Moshe, and G.~Baining.
\newblock Manifold bootstrapping for {SVBRDF} capture.
\newblock {\em ACM Transactions on Graphics}, 29(4), 2010.

\bibitem{Zhang2004}
L.~Zhang, N.~Snavely, B.~Curless, and S.~M. Seitz.
\newblock Spacetime faces: high resolution capture for modeling and animation.
\newblock {\em ACM Transactions on Graphics}, 23:548--558, 2004.

\bibitem{zheng2019numerical}
Q.~Zheng, A.~Kumar, B.~Shi, and G.~Pan.
\newblock Numerical reflectance compensation for non-lambertian photometric
  stereo.
\newblock {\em IEEE Transactions on Image Processing}, 2019.

\bibitem{Zhou2013}
Z.~Zhou, Z.~Wu, and P.~Tan.
\newblock Multi-view photometric stereo with spatially varying isotropic
  materials.
\newblock In {\em Proc. of Computer Vision and Pattern Recognition}, 2013.

\end{thebibliography}
}


%



\begin{IEEEbiography}
	[{\includegraphics[width=1in,height=1.25in,clip,keepaspectratio]{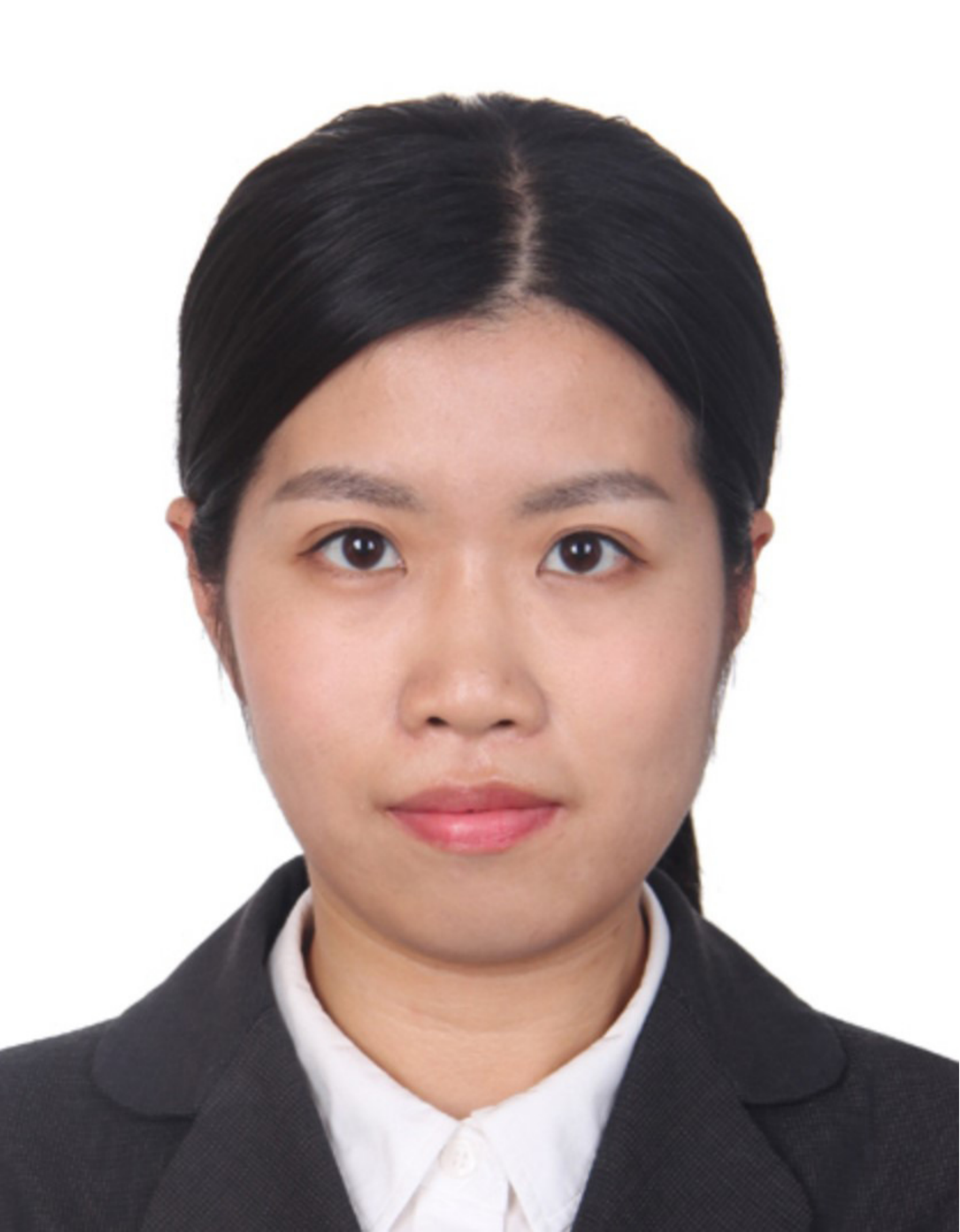}}]
	{Min Li} received the B.E. degree from Northeastern University in China in 2011. She joined the school of computer science and technology at Zhejiang University in 2011, where currently she is a Ph.D. candidate under the supervision of Prof. Duanqing Xu. She was a visiting student at Simon Fraser University supervised by Prof. Ping Tan from 2015 to 2016. Her research interests include photometric methods in computer vision, reflectance and illumination modeling, and 3D reconstruction.	
\end{IEEEbiography}

\begin{IEEEbiography}
	[{\includegraphics[width=1in,height=1.25in,clip,keepaspectratio]{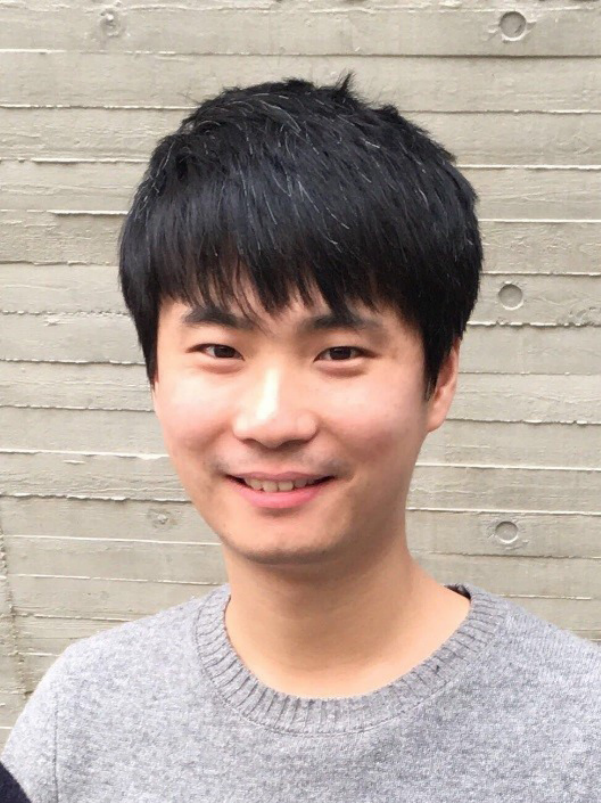}}]
	{Zhenglong Zhou} received his B.S. degree from Shanghai Jiao Tong University, China, in 2009 and Ph.D. degree from National University of Singapore in 2014, supervised by Prof. Ping Tan. He is currently a technical artist at Changyou. Before that, he worked at Giant and 360 as a software engineer in China. He is interested in computer vision and graphics engine development.	
\end{IEEEbiography}

\begin{IEEEbiography}
	[{\includegraphics[width=1in,height=1.25in,clip,keepaspectratio]{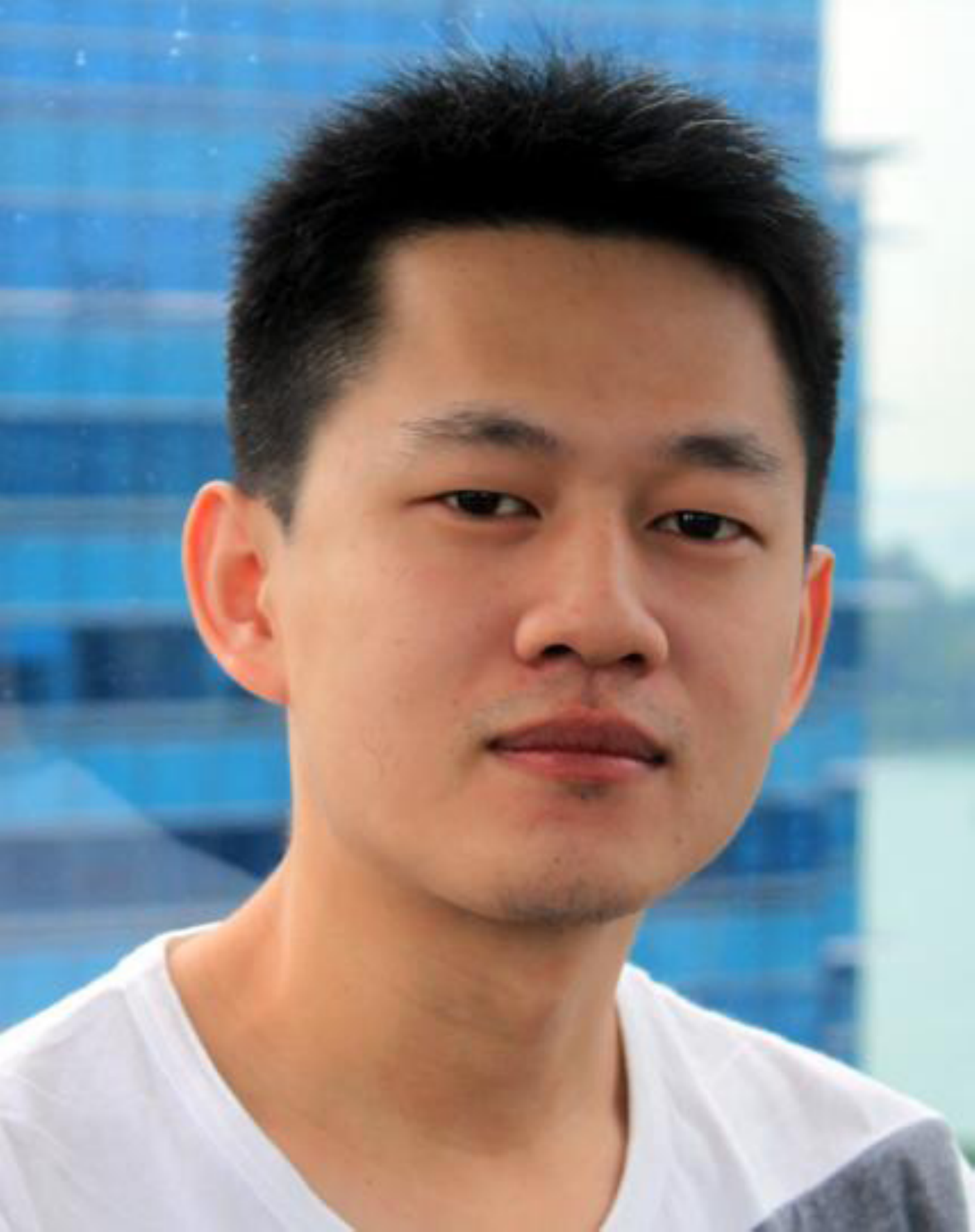}}]
	{Zhe Wu} received the B.E. degree from Tsinghua University in China in 2010 and the Ph.D. degree in computer vision from National University of Singapore in 2015. He is currently a vision engineer at DJI Innovations developing autonomous navigation systems for drones.
\end{IEEEbiography}

\begin{IEEEbiography}
	[{\includegraphics[width=1in,height=1.25in,clip,keepaspectratio]{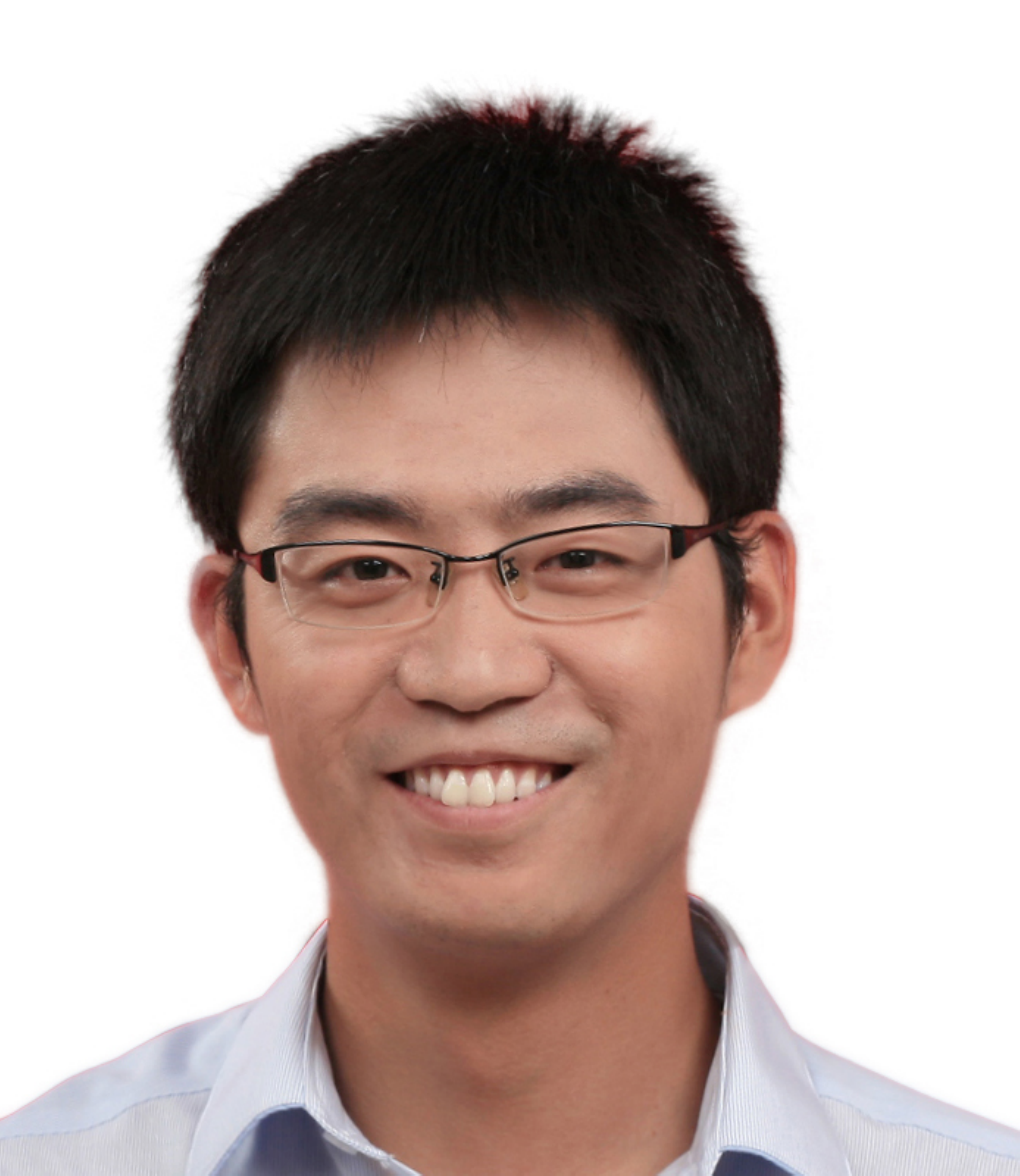}}]
	{Boxin Shi} is currently a Boya Young Fellow Assistant Professor at Peking University, where he leads the Camera Intelligence Group. Before joining PKU, he did postdoctoral research at MIT Media Lab, Singapore University of Technology and Design, Nanyang Technological University from 2013 to 2016, and worked as a Researcher at the National Institute of Advanced Industrial Science and Technology from 2016 to 2017. He received the B.E. degree from Beijing University of Posts and Telecommunications in 2007, M.E. degree from Peking University in 2010, and Ph.D. degree from the University of Tokyo in 2013. He won the Best Paper Runner-up award at International Conference on Computational Photography 2015. He has served as Area Chairs for ACCV 2018, and BMVC 2019.
\end{IEEEbiography}

\begin{IEEEbiography}
	[{\includegraphics[width=1in,height=1.25in,clip,keepaspectratio]{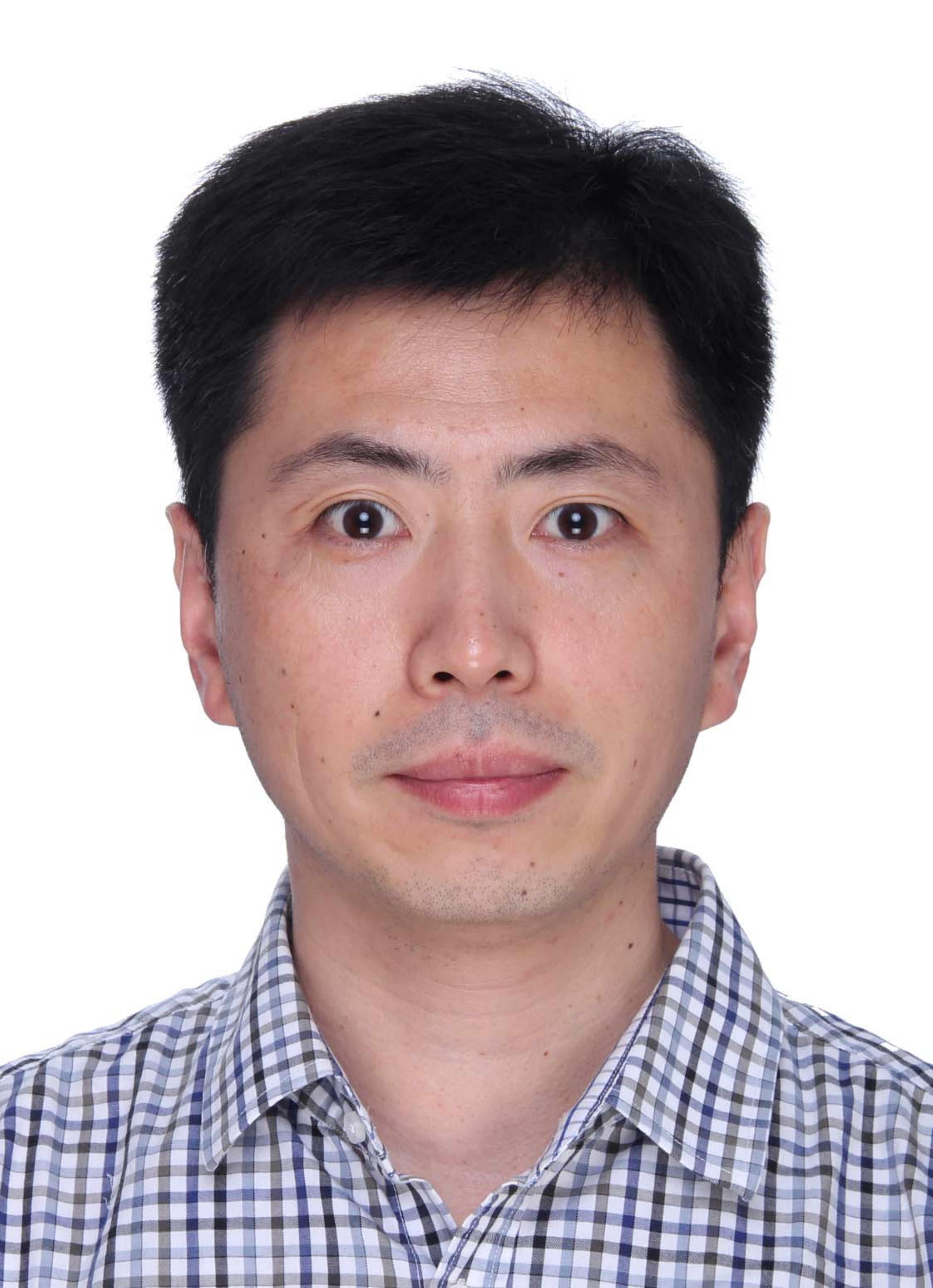}}]
	{Changyu Diao} is currently an associate professor at Zhejiang University, where he leads the Cultural Heritage Digital Research group. Before joinning ZJU, received his B.S., M.S., and Ph.D. degrees in Computer Science and Technology from the Zhejiang University in China in 2000, 2003 and 2008, respectively. He joined the Cultural Heritage Institute of Zhejiang University of China in 2010. He has been engaged in cultural heritage digitization research for over ten years, including 3D digitization, information management, information processing and analysis, virtual exhibition, digital museum, etc.
\end{IEEEbiography}

\begin{IEEEbiography}
	[{\includegraphics[width=1in,height=1.25in,clip,keepaspectratio]{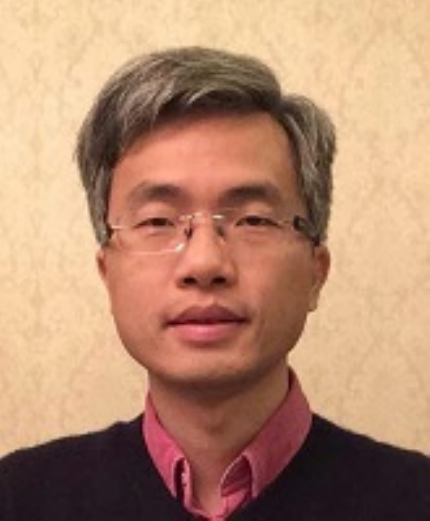}}]
	{Ping Tan} is an associate professor with the School of Computing Science at Simon Fraser University (SFU). Before that, he was an associate professor at National University of Singapore (NUS). He obtained his PhD degree from the Hong Kong University of Science and Technology (HKUST) in 2007, and his Master and Bachelor degrees from Shanghai Jiao Tong University (SJTU), China, in 2003 and 2000 respectively. His research interests include computer vision, computer graphics, and robotics. He has served as an editorial board member of the IEEE Transactions on Pattern Analysis and Machine Intelligence (PAMI), International Journal of Computer Vision (IJCV), Computer Graphics Forum (CGF), and the Machine Vision and Applications (MVA), and served as area chairs for CVPR, SIGGRAPH, SIGGRAPH Asia, and IROS.
\end{IEEEbiography}




\end{document}